\documentclass[a4paper,fleqn]{cas-sc}

\usepackage[authoryear,longnamesfirst]{natbib}

\def\tsc#1{\csdef{#1}{\textsc{\lowercase{#1}}\xspace}}
\tsc{WGM}
\tsc{QE}
\usepackage{booktabs, tabularx}
\usepackage{multirow}
\usepackage[flushleft]{threeparttable}
\usepackage{longtable}
\usepackage{threeparttablex}
\usepackage{placeins}

\usepackage{enumitem}
\setlist[itemize]{noitemsep}
\usepackage{amsmath,amsfonts,amssymb,accents}

\usepackage{algorithmic}
\usepackage{algorithm}
\usepackage{array}
\usepackage{graphicx}
\graphicspath{{./img/}}

\usepackage{rotating}
\usepackage[table]{xcolor}
\usepackage{tablefootnote}
\usepackage{hyperref}
\usepackage{adjustbox}
\usepackage{multicol}
\usepackage{color,soul}
\usepackage{caption}
\usepackage{siunitx}
\usepackage{subcaption}
\usepackage{geometry}
\usepackage{pdflscape}
\usepackage{tcolorbox}
\tcbuselibrary{skins, listings, breakable}

\newtheorem{definition}{Definition}
\makeatletter

\begin{document}
\let\WriteBookmarks\relax
\def\floatpagepagefraction{1}
\def\textpagefraction{.001}

\shorttitle{GGATN}    

\shortauthors{Wang and Damiani}  

\title [mode = title]{Graph Grounded Cross Attention Transformer Neural Network for Structurally Constrained Full Event Sequence Generation in Predictive Process Monitoring}  

\tnotemark[1] 

\tnotetext[1]{The source code associated with this paper is available at \url{https://github.com/skyocean/GGATN}.} 

\author[1]{Fang Wang}[orcid=0000-0002-7349-8888]

\cormark[1]

\ead{florencewong@protonmail.com}

\credit{Conceptualization, Methodology, Software, Validation, Formal analysis, Investigation, Data curation, Writing - original draft, Writing - review and editing, Visualization}

\affiliation[1]{organization={Department of Computer Science, University of Milan},
            addressline={Via Giovanni Celoria, 18}, 
            city={Milano},
            citysep={}, %
            postcode={20133}, 
            state={MI},
            country={Italy}}

\author[1]{Ernesto Damiani}[orcid=0000-0002-9557-6496]

% Email id of the second author
\ead{ernesto.damiani@unimi.it}

\credit{Supervision, Conceptualization, Methodology, Writing - review and editing, Project administration}

\cortext[1]{Corresponding author}



\begin{abstract}
Structurally constrained event sequence generation remains challenging because generated paths must preserve transition feasibility, temporal order, termination, and attribute consistency. In predictive process monitoring (PPM), this challenge appears as full event sequence generation, whereas existing work mainly addresses component tasks such as next activity, remaining time, outcome, and attribute prediction. This paper proposes the Graph Grounded Cross Attention Transformer Neural Network (GGATN) for this unified PPM task. GGATN uses a global process graph as structured activity memory, contextualizes sequence positions through Transformer self attention, and injects process topology through graph grounded cross attention. Unlike autoregressive decoding, GGATN generates activities, timestamps, length, and event level and sequence level attributes in a single pass, followed by Viterbi style graph constrained decoding for feasible paths and explicit termination. Experiments on six benchmark event logs show more reliable generation quality than local instruction prompted LLM baselines. GGATN achieves strong performance on sequence similarity, Damerau Levenshtein similarity, bigram based control flow similarity, and duration distribution, while maintaining zero hallucinated activities and zero sequence level attribute inconsistency. Ablation analyses confirm the global graph encoder as a stable structural prior. Interpretability analyses show how graph structure, sequence context, feedback refinement, and constrained decoding shape generation.
\end{abstract}


%\item Formulates full event sequence generation as a unified PPM task.
%\item Proposes GGATN with graph grounded cross attention.
%\item Global process graph memory beyond Transformer self attention.
%\item Non autoregressive generation with Viterbi graph constrained decoding.
%\item Outperforms local LLM baselines in quality and computation cost.

% Keywords

\maketitle

\section{Introduction}
\label{sec:Intro}
Event sequence generation becomes structurally constrained when generated paths must satisfy temporal ordering, admissible transitions, explicit termination, and attribute consistency. Predictive process monitoring (PPM) provides a representative setting for this problem because future process executions are recorded as timestamped event sequences governed by process structure and operational constraints. PPM aims to anticipate future process behavior from historical event logs \citep{marquez2017predictive}, enabling organizations to intervene before process deviations become costly \citep{pasquadibisceglie2021multi}. Existing PPM research has primarily focused on task specific prediction settings, including next activity prediction \citep{taymouri2020predictive,weinzierl2020prescriptive,tax2017predictive}, remaining time prediction \citep{verenich2019survey,tax2017predictive,navarin2017lstm,guo2024explainable}, outcome prediction \citep{teinemaa2019outcome,wang2025comprehensive}, and process attribute prediction \citep{rivera2022multi}. Although these approaches provide useful local forecasts, practical decision making often depends on understanding the complete future execution rather than individual prediction components. For example, in clinical pathways, decision makers may need to evaluate alternative patient trajectories together with their temporal evolution and associated attributes. Full event sequence generation addresses this limitation by modeling the entire process execution as a structured generation problem conditioned only on high level sequence information, without requiring prefix based observations. Despite its practical importance, this setting remains largely unexplored in PPM.

Sequence modeling has evolved from recurrent neural networks (RNNs) \citep{elman1990finding,medsker2001recurrent,sutskever2014sequence} and encoder decoder architectures \citep{cho2014learning} to Transformer based models \citep{vaswani2017attention}. By replacing recurrence with self-attention, Transformers capture long range dependencies \citep{zhou2021informer} and support flexible sequence to sequence (seq2seq) generation. Large language models (LLMs) further extend this paradigm through large scale pretraining and prompt conditioned generation. In PPM, this architectural progression has also shaped recent modeling efforts, with Transformer based and LLM based approaches increasingly adapted to discriminative or component wise prediction \citep{nguyen2020time,chen2022multi}. However, when applied to full sequence generation, purely Transformer based architectures face two limitations. First, self-attention models positional dependencies but does not explicitly encode process topology or transition admissibility. Second, autoregressive generation is vulnerable to exposure bias and error accumulation, making it difficult to preserve globally coherent process behavior during long horizon generation without observed prefix guidance \citep{schmidt2019generalization}. Consequently, step wise generation may fail to maintain globally coherent process behavior across a complete sequence.

Graph neural networks (GNNs) \citep{scarselli2008graph} provide advantages for structure aware sequence modeling by propagating information through relational dependencies rather than relying only on sequential token interactions. Among GNN variants, graph attention networks (GATs) further learn the relative importance of neighboring nodes through adaptive attention mechanisms \citep{velivckovic2017graph}. These properties are particularly relevant for PPM, where process executions are shaped by activity transitions, temporal dependencies \citep{wang2025time}, and global process topology \citep{yin2025stgnn}. However, graph propagation alone does not naturally preserve position specific sequence dynamics. Full sequence generation in PPM therefore requires a unified framework that combines graph grounded process representations with position aware sequence modeling. Despite the complementary strengths of graph attention and Transformer self-attention, their integration for full sequence generation in PPM remains largely unexplored.
 
To address these limitations, we propose the Graph Grounded Cross Attention Transformer Neural Network (GGATN) for structurally constrained full event sequence generation in PPM. The novelty of this work is reflected in the following contributions:

\begin{itemize}
[leftmargin=*]

\item \textbf{Hybrid graph sequence architecture:} GGATN departs from sequence only and graph only PPM architectures by organizing full sequence generation into three modeling stages: global process graph learning, sequence level contextual modeling, and graph grounded cross attention interaction. This design preserves global process regularities while modeling position specific dependencies. By coupling process topology and sequence dynamics through an explicit cross attention stage, GGATN enables controlled and interpretable structural injection.

\item \textbf{Global graph structural prior:} GGATN uses the global process graph as a learned structural backbone rather than a descriptive process map. Through graph attention network (GAT) based adaptive neighbor weighting, the graph encoder embeds activity transition structure and temporal gap information into graph representations that complement Transformer self-attention. The ablation study compares frozen, unfrozen, and staged graph encoder training regimes, showing that this graph backbone remains stable across training strategies and provides an effective structural prior for full sequence generation. 

\item \textbf{Graph grounded cross attention:} GGATN introduces graph grounded cross attention as the central interaction mechanism between global process structure and sequence level generation. At each position, the sequence representation queries graph learned activity embeddings as a structured process memory. This allows transition semantics, relational dependencies, and temporal regularities to be selectively injected into position specific sequence states, grounding Transformer based generation in the global activity graph rather than relying only on positional dependency modeling.

\item \textbf{Non autoregressive graph constrained decoding:} Unlike traditional Transformer autoregressive decoding, GGATN formulates full event sequence generation as single pass prediction followed by graph constrained structured decoding. The decoder recovers complete activity paths by combining graph aligned activity scores with transition bias, optional bigram priors, adjacency constraints, feedback refinement, and Viterbi style path optimization. This design enforces strict structural validity and explicit termination directly within the graph space, optimizing sequence level transition feasibility globally rather than locally.
     
\item \textbf{Full event sequence generation benchmark:} This study establishes the first full event sequence generation benchmark for PPM, covering activities, timestamps, event level and sequence level attributes, and explicit termination. We demonstrate that GGATN outperforms locally deployed, instruction-prompted LLM baselines in overall generation quality while substantially reducing computational overhead. The results further support a transferable design principle for structured full sequence generation: process graph learning and Transformer based sequence modeling can be coupled through graph grounded interaction and constrained decoding.

\item \textbf{Multi level interpretability analysis:} GGATN includes a multi level interpretability protocol designed to make graph grounded generation inspectable by domain experts and model developers. Through dedicated visualizations, the analysis examines structural learning, graph grounded attention, feedback refinement, and constrained decoding. It shows how graph based information is injected into sequence states, how activity level attention patterns vary across positions, how refinement reshapes activity distributions, and how graph constrained decoding recovers the final activity path.

\end{itemize}

The remainder of this paper is organized as follows. Section~\ref{sec:related} reviews related work. Section~\ref{sec:PD} presents the preliminaries and task definitions, and Section~\ref{sec:method} introduces the proposed GGATN methodology. Section~\ref{sec:llm-base} describes the LLM baseline, followed by the datasets in Section~\ref{sec:data} and the experimental setup and evaluation metrics in Section~\ref{sec:experiments}. Section~\ref{sec:Res} reports and discusses the experimental results. Section~\ref{sec:ablation} presents the ablation analysis, and Section~\ref{sec:interpre} provides the interpretability analysis. Finally, Section~\ref{sec:conclusion} concludes the paper.

\section{Related Work}
\label{sec:related}

This section reviews four research streams relevant to this study: task specific PPM, full sequence generation, graph based structure modeling, and LLM based generation. Together, these streams provide the methodological context for full event sequence generation and motivate the need for a graph grounded model that combines position aware sequence modeling with structured decoding.

\subsection{Predictive Process Monitoring}
\label{subsec:related_pbpm}

PPM research is motivated by the need to support decisions before an ongoing process reaches completion. This objective has led to well established task lines, including next activity or event prediction \citep{dissegna2024graph,tax2017predictive,evermann2016deep}, remaining time prediction \citep{khan2021deepprocess}, outcome prediction \citep{teinemaa2019outcome,wang2019outcome, wang2025auto}, and process attribute prediction \citep{lin2019mm,rivera2022multi,duong2023remaining}. Although these tasks focus on different prediction targets, they share the broader aim of estimating future process behavior from historical execution data.

Next activity and next event prediction form a central task line in PPM. Early approaches adopted recurrent and convolutional sequence models, including Long Short Term Memory networks (LSTMs), Convolutional Neural Networks (CNNs), Memory Augmented Neural Networks (MANNs), and Generative Adversarial Networks (GANs), to model event dependencies and predict upcoming activities or timestamps \citep{tax2017predictive,camargo2019learning,taymouri2020predictive,khan2021deepprocess,pasquadibisceglie2019using}. Later studies introduced graph based models, including GGNN and GAT based architectures, to represent process structure and event relations more explicitly \citep{weinzierl2021exploring,pasquadibisceglie2024sf,rama2023embedding}. More recent Transformer based models further extended this line by using self-attention to capture long range dependencies in event sequences \citep{bukhsh2021processtransformer,wuyts2024sutran}.

Remaining time prediction and outcome oriented PPM have followed similar methodological trajectories. Remaining time prediction has been studied across diverse datasets and machine learning model families \citep{verenich2019survey,kratsch2021machine,harane2020comprehensive}, with later work exploring neural and explainable temporal models \citep{navarin2017lstm,guo2024explainable}. Outcome oriented PPM has progressed through decision tree based, clustering based, regression based, and deep learning models \citep{teinemaa2019outcome}. Recent AutoML and HyperModel frameworks have further extended this line within GNN and LSTM architectures \citep{wang2025comprehensive,wang2025hgcn, wang2025leveraging}. These tasks are important for operational intervention, but their focus on future process behavior is limited to either a temporal scalar or a final outcome label.

Attribute prediction and multi task PPM extend the predictive scope by considering event level or process level attributes in addition to activities, time, or outcomes. \citet{chen2022multi} combine BERT and transfer learning to jointly predict next events and outcomes, while \citet{lin2019mm} introduce an LSTM based component modulation mechanism for multi attribute prediction. These models broaden the output space, but they still operate through predefined target components. Overall, existing PPM research has advanced individual prediction tasks, whereas full sequence generation remains underexplored as a unified setting that jointly produces activities, timestamps, attributes, and termination.

\subsection{Sequence Modeling and Transformer Based Approaches}
\label{subsec:SM}

Sequence modeling is fundamental to full sequence generation across multiple domains, where the objective is to learn conditional dependencies over ordered outputs \citep{hochreiter1997lstm,sutskever2014sequence}. Early sequence modeling approaches relied on Recurrent Neural Networks (RNNs), but their ability to capture long range dependencies was limited by the vanishing gradient problem. Long Short Term Memory (LSTM) networks addressed this limitation through memory cells and gating mechanisms that regulate information flow across time steps \citep{hochreiter1997lstm}. Gated Recurrent Units (GRUs) later simplified this design with merged gates while maintaining competitive performance across sequence learning tasks \citep{chung2014empirical}.

Encoder decoder architectures further extended sequence modeling by separating context representation from autoregressive generation, enabling learning over variable length input and output pairs \citep{cho2014learning}. Attention mechanisms subsequently alleviated the fixed context bottleneck of early seq2seq models by allowing the decoder to dynamically access relevant encoder states during generation \citep{bahdanau2014neural}. Transformers later replaced recurrence with self-attention, where each sequence position dynamically attends to other positions within the sequence to model contextual dependencies. This improved training parallelism and strengthened long range dependency modeling, establishing Transformer architectures as the center backbone for modern sequence generation \citep{vaswani2017attention,radford2021learning,brown2020language,raffel2020exploring,lewis2020bart}.

However, standard Transformer architectures primarily model structure through token order, positional embeddings, and self-attention over sequence positions, while relational dependencies, hierarchical constraints, and graph structured interactions are flattened into sequential representations \citep{zhu2019modeling,zhou2021structure}. Self-attention is effective for modeling position aware contextual dependencies, but it does not explicitly represent process topology or transition admissibility. In addition, autoregressive decoding remains susceptible to exposure bias, where local prediction errors accumulate across long horizons and gradually degrade structural consistency \citep{schmidt2019generalization}. These limitations are particularly problematic for full sequence generation tasks that require strict transition validity, global structural coherence, and explicit termination control, motivating architectures that combine position aware Transformer self-attention with explicit structural grounding.

\subsection{Graph Neural Networks for Structure Aware Sequence Modeling} 
\label{subsec:related_gnn}

Graph neural networks (GNNs) provide a natural framework for modeling structured dependencies through message passing over explicit relational graphs \citep{gilmer2017neural,bronstein2021geometric}. Foundational variants introduced different relational inductive biases, including graph convolutional networks (GCNs) for localized graph propagation \citep{kipf2016semi}, GraphSAGE for inductive neighborhood aggregation \citep{hamilton2017inductive}, and gated graph neural networks (GGNNs) for recurrent graph propagation \citep{li2015gated}. Among these variants, graph attention networks (GATs) are particularly relevant because they learn adaptive neighbor weights through attention driven relational importance modeling \citep{velivckovic2017graph}. This attention mechanism makes GATs well suited for modeling dependency, interaction, and transition structures in graph based learning tasks.

Later sequence generation systems extended GNNs beyond representation learning by combining graph encoders with sequential decoders. Graph2Seq introduced graph encoding with attention based LSTM decoding for graph conditioned sequence generation \citep{xu2018graph2seq}. \citet{beck2018graph} combined GGNN encoding with sequence generation for graph conditioned natural language generation, while GraphWriter incorporated graph encoding into encoder decoder generation architectures \citep{koncel2019text}. However, graph propagation itself, even with attention based neighbor weighting, remains primarily local and representation oriented, making it difficult to preserve ordered generation states, long horizon dependencies, and variable length termination behavior in full sequence generation settings \citep{gilmer2017neural,alon2020bottleneck}.

More recent graph Transformer architectures explored how graph structure can be integrated with self-attention mechanisms \citep{dwivedi2020generalization}. For example, Graphormer incorporates structural encodings into standard Transformer architectures \citep{ying2021transformers}, while GraphGPS explicitly separates local edge aggregation from global attention \citep{rampavsek2022recipe}. This progression suggests that graph attention for structural representation learning and Transformer self-attention for position aware sequence modeling should be treated as complementary components for full sequence generation.

\subsection{LLM Based Generation for Structured Sequences}
\label{subsec:LLM-review}

Transformer based large language models (LLMs) are powerful general purpose sequence generators, with strong zero shot and few shot performance across diverse language tasks \citep{vaswani2017attention,brown2020language,touvron2023llama}. Through prompt based linearization and serialization, structured objects can be reformulated as text generation problems. This allows graphs, event sequences, activities, timestamps, attributes, and termination markers to be represented as token sequences and generated autoregressively \citep{fan2019using,cao2025graphinsight,kasner2024beyond}. 

For PPM full sequence generation, however, this formulation introduces several limitations. It requires nontrivial serialization and post hoc parsing, provides limited guarantees on transition validity or structural constraints without additional constrained decoding, remains brittle for temporal consistency and numerical precision, and can incur substantial inference cost and latency \citep{geng2025generating,wang2024tram,li2025exposing,kandpal2023large}. Accordingly, LLMs provide flexible and practically relevant baselines for this study, while their limitations motivate graph grounded architectures that better align generation with the structural validity and efficiency requirements of PPM full sequence generation \citep{kasner2024beyond,geng2025generating}.

In summary, existing studies provide important foundations for modeling future process behavior, sequence dependencies, relational structure, and prompt based generation. However, they remain fragmented across component prediction, sequence only generation, graph based representation learning, and text based LLM prompting. This gap motivates GGATN, which integrates GAT based process graph learning, Transformer self-attention for position aware sequence modeling, graph grounded cross attention, and graph constrained decoding for full sequence generation.

\section{Preliminaries and Definitions}
\label{sec:PD}
This section defines the notation and task formulation used throughout the paper.

\begin{definition}[Sequence and Event]
\normalfont A sequence $\sigma$ is an ordered list of events denoted by $\sigma = \langle e_1, e_2, \dots, e_T \rangle$, where $e_t$ is the event at position $t$ and $T$ is the sequence length.

Each event $e_t$ is associated with an activity label $a_t \in \mathcal{A}$, a timestamp $\tau_t \in \mathbb{R}$, and event level attributes $\mathbf{x}_t = (\mathbf{x}_t^{(cat)}, \mathbf{x}_t^{(num)})$. Each sequence $\sigma$ is further associated with sequence level attributes $\mathbf{s} = (\mathbf{s}^{(cat)}, \mathbf{s}^{(num)})$. Boolean attributes are treated as categorical variables.
\end{definition}

\begin{definition}[Special Tokens]
\normalfont To support sequence generation and alignment, the activity space is extended with four special tokens: $\texttt{PAD} = 0$, $\texttt{UNK} = 1$, $\texttt{EOS} = 2$, and $\texttt{SOS} = 3$. The extended activity space is denoted by $\mathcal{A}'$.
\end{definition}

\begin{definition}[Full Event Sequence Generation]
\normalfont Given a start time $\tau_1 \in \mathbb{R}$ and a target sequence length $T$, the full sequence generation task aims to generate a complete sequence $\hat{\sigma} = \langle \hat{e}_1, \hat{e}_2, \dots, \hat{e}_T \rangle$.

Each generated event $\hat{e}_t$ consists of an activity label $\hat{a}_t \in \mathcal{A}$, a timestamp $\hat{\tau}_t \in \mathbb{R}$, and event level attributes $\hat{\mathbf{x}}_t = (\hat{\mathbf{x}}_t^{(cat)}, \hat{\mathbf{x}}_t^{(num)})$. The generated sequence $\hat{\sigma}$ is further associated with sequence level attributes $\hat{\mathbf{s}} = (\hat{\mathbf{s}}^{(cat)}, \hat{\mathbf{s}}^{(num)})$. The generated timestamps satisfy $\hat{\tau}_1 = \tau_1$ and $\hat{\tau}_t \geq \hat{\tau}_{t-1}$ for all $t > 1$. The objective is to jointly generate the activity sequence, timestamps, event level attributes, and sequence level attributes conditioned on $(\tau_1,T)$.
\end{definition}

\section{Methodology}
\label{sec:method}

\begin{figure}[pos=htbp]
    \centering
    \includegraphics[width=\textwidth]{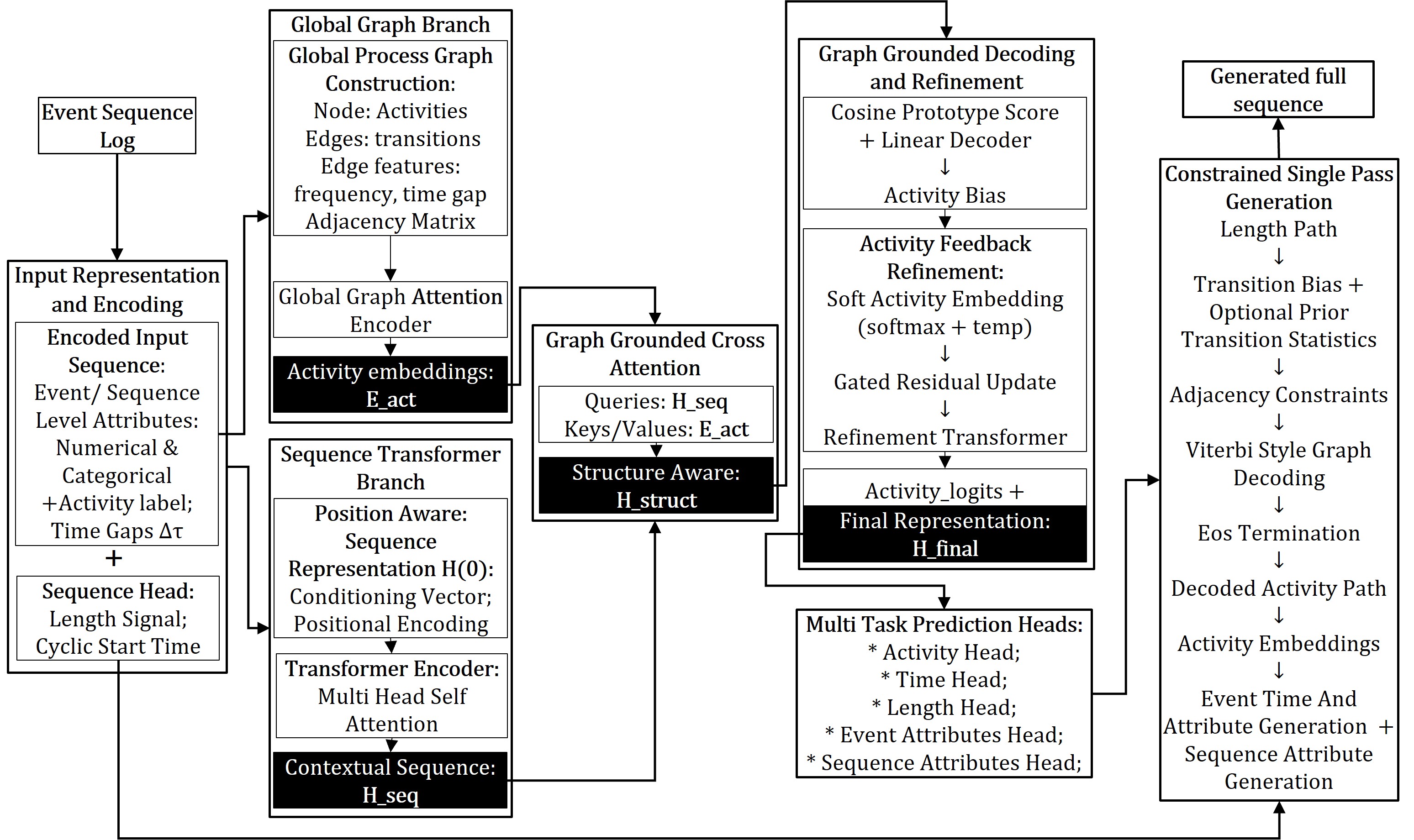}
    \caption{Overview of GGATN for Structurally Constrained Full Event Sequence Generation}
    \label{fig:model}
\end{figure}

The proposed Graph Grounded Cross Attention Transformer Neural Network (GGATN) addresses full sequence generation by coupling structured process knowledge with position aware sequence modeling. As illustrated in Figure~\ref{fig:model}, GGATN consists of five components: input representation and sequence encoding, global process graph construction with graph attention network (GAT) based activity embeddings, Transformer based sequence contextualization, graph grounded cross attention for structural injection, and joint decoding of activities, time, length, and attributes. The training objective and graph constrained generation procedure are presented at the end of this section. 

\subsection{Input Representation and Encoding}

Given an event log $\mathcal{L}$, each sequence $\sigma$ is transformed into a fixed length representation of size $T_{\max}$. The input encoding contains the following components:
\begin{itemize}[leftmargin=*]
    \item \textbf{Categorical attributes:} Activity labels, event level categorical attributes, and sequence level categorical attributes are mapped to integer indices through vocabulary mappings with special tokens $\{\texttt{PAD}, \texttt{UNK}, \texttt{EOS}, \texttt{SOS}\}$.
    
    \item \textbf{Numerical attributes:} Event level and sequence level numerical attributes are standardized using training set statistics. For strictly positive attributes, a $\log(1+x)$ transformation is applied before standardization.
    
    \item \textbf{Temporal attributes:} Inter event time differences are defined as $\Delta \tau_1 = 0$ and $\Delta \tau_t = \tau_t - \tau_{t-1}$ for $t > 1$. These values are transformed using $\log(1+\Delta \tau_t)$ and then standardized.
    
    \item \textbf{Initial timestamp:} The initial timestamp $\tau_1$ is encoded using four cyclic features derived from the hour of day and day of week, namely $\sin(2\pi h/24)$, $\cos(2\pi h/24)$, $\sin(2\pi w/7)$, and $\cos(2\pi w/7)$, where $h$ denotes the fractional hour of day and $w$ denotes the day of week.
    
    \item \textbf{Sequence head:} Each sequence is associated with a compact conditioning vector consisting of the normalized sequence length $T / T_{\max}$ and the encoded initial timestamp.
\end{itemize}

The encoded input sequence is finally represented as a matrix $\mathbf{H} \in \mathbb{R}^{T_{\max} \times d}$, where each position corresponds to the concatenation of embedded categorical features, standardized numerical features, and temporal features.

\subsection{Global Graph Attention Encoder}
\label{sec:global_gat}

To capture global transition structure and temporal dynamics across the event log, we construct a global process graph and encode activity nodes into structure aware latent representations using an edge aware GAT encoder.

\subsubsection{Global Process Graph}
\label{subsec:graph_pro}

Given an event log $\mathcal{L}$, we construct a global process graph
\[
\mathcal{G} = (\mathcal{V}, \mathcal{E}, \mathbf{X}^e, A),
\]
where
\begin{itemize}
    \item $\mathcal{V}$ is the set of activity nodes,
    \item $\mathcal{E}$ is the set of directed transitions between activities,
    \item $\mathbf{X}^e$ is the set of edge attributes,
    \item $A \in \{0,1\}^{|\mathcal{V}| \times |\mathcal{V}|}$ is the adjacency matrix.
\end{itemize}

Each node $v \in \mathcal{V}$ corresponds to an activity in the extended activity space $\mathcal{A}'$, and each directed edge $(u,v) \in \mathcal{E}$ indicates that activity $v$ directly follows activity $u$ in at least one observed sequence.

For each transition $(u,v)$, the associated edge attribute vector is defined as $\mathbf{x}^{e}_{uv} = (f_{uv}, d_{uv})$:
\[
f_{uv} = \log(1 + c_{uv}),
\qquad
d_{uv} = \frac{1}{c_{uv}} \sum_{k=1}^{c_{uv}} \log(1 + \Delta \tau_{uv}^{(k)}),
\]
where $c_{uv}$ is the number of observed occurrences of transition $(u,v)$ in the training log, and $\Delta \tau_{uv}^{(k)}$ is the inter event time gap for its $k$th occurrence. To incorporate boundary structure, transitions from $\texttt{SOS}$ to the first activity of each sequence and from the last activity of each sequence to $\texttt{EOS}$ are included. For transitions into $\texttt{EOS}$, the temporal gap is set to $0$. The attribute $f_{uv}$ captures the global strength of transition $(u,v)$, whereas $d_{uv}$ captures its typical temporal separation, allowing the encoder to model both structural and temporal transition patterns.

\subsubsection{Graph Attention Encoding}
\label{subsec:global_att}

Let $\mathbf{E}^{(0)} \in \mathbb{R}^{|\mathcal{V}| \times d}$ denote the learnable initial node embedding matrix, where $d$ is the embedding dimension. The global graph encoder applies $L$ stacked edge aware multi-head graph attention network (GAT) layers over $\mathcal{G}$ to produce structure aware activity representations. For layer $l \in \{0, \dots, L-1\}$, the node representations are updated as
\[
\widetilde{\mathbf{E}}^{(l+1)} = \mathrm{GAT}^{(l)}\big(\mathbf{E}^{(l)}, \mathcal{E}, \mathbf{X}^e\big),
\]
where the attention mechanism is conditioned on both neighboring node features and edge attributes.

To stabilize training, each layer is followed by a residual connection:
\[
\mathbf{E}^{(l+1)} = \widetilde{\mathbf{E}}^{(l+1)} + \mathbf{E}^{(l)}.
\]
Layer normalization and dropout are then applied to $\mathbf{E}^{(l+1)}$. After $L$ layers, the encoder outputs the final activity embedding matrix
\[
\mathbf{E}_{act} = \mathbf{E}^{(L)} \in \mathbb{R}^{|\mathcal{V}| \times d}.
\]

In our implementation, the encoder uses multi-head attention with concatenated heads such that the output dimensionality remains $d$ at every layer. The attention weights from the final graph attention layer are retained for interpretability analysis.

The resulting $\mathbf{E}_{act}$ serves as the global structural memory that is later used as the key and value space in graph grounded cross attention.

\subsection{Sequence Transformer Encoder}
\label{sec:seq_transformer}

In parallel, each input sequence is encoded using a Transformer encoder with sinusoidal positional encoding, yielding
\[
\mathbf{H}^{(0)} = \mathbf{H} + \mathbf{P},
\]
where $\mathbf{H} \in \mathbb{R}^{B \times L \times d}$ denotes the batch of input sequence representations and $\mathbf{P} \in \mathbb{R}^{L \times d}$ denotes the positional encoding matrix.

A stack of $M$ pre-layer normalization Transformer encoder layers is applied to $\mathbf{H}^{(0)}$. For layer $m \in \{0, \dots, M-1\}$, the hidden representation is updated by multi-head self-attention ($\mathrm{MHA}(\cdot)$), followed by a position-wise feed-forward network ($\mathrm{FFN}(\cdot)$), each with residual connections:
\[
\widetilde{\mathbf{H}}^{(m+1)} = \mathbf{H}^{(m)} + \mathrm{MHA}\big(\mathrm{LN}(\mathbf{H}^{(m)})\big)
\qquad \text{and} \qquad
\mathbf{H}^{(m+1)} = \widetilde{\mathbf{H}}^{(m+1)} + \mathrm{FFN}\big(\mathrm{LN}(\widetilde{\mathbf{H}}^{(m+1)})\big).
\]

After the final layer, layer normalization is applied to obtain the contextualized sequence representation
\[
\mathbf{H}_{seq} = \mathrm{LN}(\mathbf{H}^{(M)}).
\]

The resulting $\mathbf{H}_{seq}$ provides position aware sequence states that are later used as queries in graph grounded cross attention.

Padding positions are masked during self-attention using a key padding mask. The attention weights from all Transformer layer are retained for interpretability analysis.

\subsection{Graph Grounded Cross Attention}
\label{sec:cross_att}

To inject global process structure into the sequence representation, we design a graph grounded cross attention module in which each sequence position attends to the global activity embeddings produced by the GAT encoder.

Let $\mathbf{H}_{seq} \in \mathbb{R}^{B \times L \times d}$ denote the contextualized sequence representation from the transformer encoder, and let $\mathbf{E}_{act} \in \mathbb{R}^{|\mathcal{V}| \times d}$ denote the global activity embedding matrix produced by the graph encoder. For each batch, $\mathbf{E}_{act}$ is broadcast across the batch dimension and used as the key and value memory in a multi-head cross attention operation, while $\mathbf{H}_{seq}$ serves as the query representation.

The cross attention output is computed as
\[
\mathbf{C} = \mathrm{MHA}\big(\mathrm{LN}(\mathbf{H}_{seq}), \mathbf{E}_{act}, \mathbf{E}_{act}\big),
\]
where layer normalization is applied to the query representation prior to attention. The resulting graph informed signal is injected into the sequence representation through a gated residual update followed by layer normalization:
\[
\mathbf{H}_{tmp} = \mathrm{LN}\big(\mathbf{H}_{seq} + g \cdot \mathbf{C}\big),
\qquad
g = \tanh(\gamma),
\]
where $\gamma$ is a learnable scalar parameter updated during training, and $g \in (-1,1)$ controls the strength of the cross attention signal injected from the global graph into each sequence position.

To refine the representation, a position wise feed forward transformation is applied:
\[
\mathbf{H}_{cross} = \mathbf{H}_{tmp} + \mathrm{FFN}\big(\mathrm{LN}(\mathbf{H}_{tmp})\big).
\]

The resulting representation $\mathbf{H}_{cross} \in \mathbb{R}^{B \times L \times d}$ captures both sequence level position aware contextual dependencies and graph grounded activity structure. During analysis, the cross attention weights are retained to quantify how each sequence position attends to the global activity nodes.

\subsection{Graph Grounded Cross Attention Transformer Neural Network (GGATN) Architecture}
\label{sec:ggatn}

The proposed Graph Grounded Cross Attention Transformer Neural Network (GGATN) integrates the global GAT encoder, sequence Transformer encoder, and graph grounded cross attention module into a unified architecture for full sequence generation. In contrast to autoregressive decoders, GGATN generates the entire sequence in a single pass with graph constrained structure decoding, avoiding step wise decoding while preserving structural and temporal coherence through graph informed attention and constrained decoding. 

Given a sequence head conditioning vector and a sequence mask $\mathbf{M}$ indicating valid and padded positions, the model first constructs an initial position aware sequence representation $\mathbf{H}^{(0)}$. The sequence Transformer encoder produces contextualized sequence states $\mathbf{H}_{seq}$, while the global GAT encoder produces the activity embedding matrix $\mathbf{E}_{act}$. The graph grounded cross attention module then conditions $\mathbf{H}_{seq}$ on $\mathbf{E}_{act}$ to obtain a structure aware sequence representation $\mathbf{H}_{struct}$, which serves as the shared latent representation for activity, time, and attribute prediction.

Although GGATN performs generation in a single pass, inter position dependencies are captured through three complementary mechanisms: Transformer self-attention for sequence contextualization, graph structural priors encoded through $\mathbf{E}_{act}$, and graph constrained decoding for transition consistent sequence recovery. To avoid leakage of target length information into the sequence generation pathway, the normalized length component can be removed from the conditioning input, while sequence length is learned separately through a dedicated prediction head.

\subsubsection{Graph Grounded Decoding and Refinement}
\label{subsec:graph_decoding}
Given the graph informed sequence representation $\mathbf{H}_{struct}$ and the global activity embedding matrix $\mathbf{E}_{act}$, GGATN performs graph grounded activity decoding by scoring each sequence position against graph learned activity embeddings, which serve as activity prototypes. Let $\widehat{\mathbf{H}}_{struct}$ and $\widehat{\mathbf{E}}_{act}$ denote the $\ell_2$ normalized hidden states and activity embeddings, respectively. The prototype based activity scores are computed as scaled cosine similarities:
\[
\mathbf{Y}_{act}^{cos} = s \, \widehat{\mathbf{H}}_{struct}\widehat{\mathbf{E}}_{act}^{\top},
\]
where $s$ is a learnable logit scale. To increase flexibility, these scores are combined with a learnable linear decoder,
\[
\mathbf{Y}_{act}^{(1)} = \mathbf{Y}_{act}^{cos} + \lambda \, \mathbf{Y}_{act}^{lin} + \mathbf{b}_{act},
\]
where $\mathbf{Y}_{act}^{lin}$ is obtained from a linear projection, $\lambda$ is a learnable mixing weight, and $\mathbf{b}_{act}$ is an optional activity bias term. Invalid symbols such as $\texttt{PAD}$, $\texttt{UNK}$, and $\texttt{SOS}$ are masked before decoding. The terminal symbol $\texttt{EOS}$ is forbidden at real event positions and is introduced only through the terminal transition after the graph constrained activity path is recovered.

During training, GGATN can incorporate transition aware supervision by adding a learned transition bias indexed by the previous ground truth activity under teacher forcing. This introduces transition level supervision without altering the parallel sequence prediction formulation.

To further improve activity prediction, GGATN optionally applies an activity feedback refinement stage. Unlike conventional single pass activity decoders that directly produce final activity logits from hidden representations, GGATN first generates provisional activity logits and feeds the induced activity distribution back into the model before final decoding. This introduces an explicit distribution reshaping stage before structured decoding, allowing the model to recalibrate position level activity distributions rather than relying only on the first decoded logits.

Let $\mathbf{P}_{act}^{(1)} = \mathrm{softmax}(\mathbf{Y}_{act}^{(1)}/\tau)$ denote the predicted activity distribution induced by the provisional logits, where $\tau$ is a temperature parameter. A soft activity embedding is then computed as
\[
\mathbf{E}_{soft} = \mathbf{P}_{act}^{(1)} \mathbf{E}_{act}, \]
and injected back into the hidden representation through a gated residual update,
\[
\mathbf{H}_{ref} = \mathrm{LN}\big(\mathbf{H}_{struct} + g_{fb}\,\phi(\mathbf{E}_{soft})\big),
\qquad
g_{fb} = \tanh(\gamma_{fb}),\]
where $\phi(\cdot)$ is a learnable projection and $\gamma_{fb}$ is a learnable scalar gate. The refined representation is then processed by an additional transformer refinement block to obtain the final sequence representation $\mathbf{H}_{final}$, from which the final activity logits $\mathbf{Y}_{act}$ are produced.

At generation time, activity generation is formulated as graph constrained path optimization. Unlike autoregressive decoding, all position level activity scores are produced simultaneously in a single forward pass before Viterbi style sequence recovery. Unary scores are obtained from the predicted activity logits, while pairwise transition scores are derived from the learned transition bias and, when available, an empirical bigram prior. The decoding horizon is determined either by the predicted length head or by an externally supplied target length. Hard constraints are enforced by masking invalid symbols and transitions according to the process graph adjacency structure.

The final activity sequence is recovered by a Viterbi style dynamic programming decoder that maximizes a joint path score over the valid activity graph by combining position level unary scores with pairwise transition scores between adjacent activities. Invalid symbols are assigned prohibitive scores, and sequence termination is handled by explicitly optimizing the transition from the final real activity to the $\texttt{EOS}$ state. Given the predicted real event length $T$, the decoded activity sequence is defined as
\[
\hat{y}_{1:T}
=
\arg\max_{y_1,\ldots,y_T}
\left[
\sum_{t=1}^{T} w_{\mathrm{unary}} s_t(y_t)
+
s_{\mathrm{start}}(y_1)
+
\sum_{t=2}^{T} s_{\mathrm{pair}}(y_{t-1}, y_t)
+
s_{\mathrm{pair}}(y_T, \texttt{EOS})
\right],
\]
where $s_t(y_t)$ denotes the neural activity score for assigning activity $y_t$ to position $t$, $s_{\mathrm{pair}}(y_{t-1}, y_t)$ denotes the graph based transition score between adjacent activities, $s_{\mathrm{start}}(y_1)$ denotes the start score for the first activity, and $w_{\mathrm{unary}}$ controls the contribution of the neural unary scores. The final term explicitly scores the transition from the last generated activity to $\texttt{EOS}$, thereby enforcing terminal feasibility within the structured decoding process.

\subsubsection{Multi Task Prediction Heads}
\label{subsec:prediction_heads}

All prediction heads operate on shared latent representations, enabling joint learning of activity, temporal, and attribute dependencies.

From the final hidden representation $\mathbf{H}_{final}$, GGATN predicts sequence length, activities, temporal values, event level attributes, and sequence level attributes. Sequence length is predicted from a masked mean pooled sequence embedding over valid positions,
\[
\mathbf{h}_{seq} = \mathrm{Pool}(\mathbf{H}_{final}),
\qquad
\mathbf{Y}_{len} = \mathrm{MLP}_{len}(\mathbf{h}_{seq}),
\]
where pooling excludes padded positions.

For event level prediction, the model concatenates the hidden state at each position with the corresponding selected activity embedding,
\[
\mathbf{Z}_{evt} = [\mathbf{H}_{final}; \mathbf{E}_{act}^{sel}],
\]
where $\mathbf{E}_{act}^{sel}$ is obtained from ground truth activity labels during teacher forced training, and from model predicted activity assignments or decoded activity sequences during evaluation and generation. The resulting event level representation $\mathbf{Z}_{evt}$ is used to predict transformed inter event time (i.e., $\log(1+\Delta \tau_t)$), as well as event level categorical and numerical attributes.

For sequence level prediction, GGATN pools both the hidden representation and the selected activity embeddings over valid non $\texttt{EOS}$ positions,
\[
\mathbf{z}_{seq} =
\left[
\mathrm{Pool}\!\left(\mathbf{H}_{final}\right);
\mathrm{Pool}\!\left(\mathbf{E}_{act}^{sel}\right)
\right].
\]
The resulting sequence representation is then used to predict sequence level categorical and numerical attributes.

\subsection{Training Objective}
\label{sec:training_objective}

GGATN is trained using a weighted multi task objective:
\[
\mathcal{L}
=
\lambda_{act}\mathcal{L}_{act}
+
\lambda_{time}\mathcal{L}_{time}
+
\lambda_{len}\mathcal{L}_{len}
+
\lambda_{evt}^{cat}\mathcal{L}_{evt}^{cat}
+
\lambda_{evt}^{num}\mathcal{L}_{evt}^{num}
+
\lambda_{seq}^{cat}\mathcal{L}_{seq}^{cat}
+
\lambda_{seq}^{num}\mathcal{L}_{seq}^{num}
+
\lambda_{trans}\mathcal{L}_{trans}.
\]

The activity loss is computed over all valid sequence positions using cross entropy, with additional weighting applied to the final and preterminal positions to improve sequence termination behavior. The length loss is formulated as cross entropy over clipped length indices. The time loss is computed in the transformed log domain using a Smooth L1 objective and is applied only to valid non $\texttt{EOS}$ positions.

Event level categorical losses are computed using cross entropy with label smoothing over valid non $\texttt{EOS}$ positions, whereas event level numerical losses are computed using mean squared error over the same positions. Sequence level categorical and numerical losses are computed from sequence level targets.

When enabled, an optional transition consistency penalty is introduced by penalizing predicted probability weight assigned to invalid adjacent activity pairs according to the process graph adjacency constraints.

\subsection{Single Pass Generation with Graph Constrained Structured Decoding}
\label{sec:generation}

In GGATN, single pass generation with graph constrained structured decoding refers to a generation procedure in which the model predicts position wise representations and activity scores for the full sequence, optionally refines these representations through activity feedback, and then recovers the final activity path using Viterbi style constrained dynamic programming over the global process graph. Here, single pass means that events are not generated sequentially through event by event autoregressive decoding.

At generation time, GGATN first obtains latent representations, activity logits, and length logits for all sequence positions. If the target length is not provided, the number of real events is determined by the length prediction head; otherwise, an externally supplied length defines the decoding horizon. The final activity sequence is recovered through the graph grounded constrained decoding procedure described in Section~\ref{subsec:graph_decoding}, yielding a valid activity path with an explicit terminal $\texttt{EOS}$ state.

Once the activity path is fixed, the corresponding decoded activity embeddings are used to construct the event level and sequence level representations defined in Section~\ref{subsec:prediction_heads}. Event level time and attribute predictions are then recomputed from the activity conditioned representation, while sequence level attributes are predicted from pooled hidden states and pooled decoded activity embeddings. This design preserves full sequence latent prediction while ensuring that temporal and attribute predictions remain consistent with the final decoded activity sequence.

\section{Large Language Model (LLM) Baseline for Sequence Generation}
\label{sec:llm-base}

Large language models are included as baselines by formulating full sequence generation as a structured sequence to sequence task. Each sequence is serialized into textual input and output representations. Generated outputs are parsed back into structured event log records for evaluation.

\subsection{LLM Input and Output Formulation}
\label{subsec:llm_io}

To cast full sequence generation as text generation, each sequence is transformed into a sequence to sequence input and output pair. The input is a compact sequence level prompt specifying the sequence identifier, initial timestamp, start weekday, start hour, and sequence length, with an instruction to generate the complete event sequence. The output is represented as an ordered textual sequence in which each line corresponds to one event position and contains the activity label, event level attributes, sequence level attributes, and timestamp. Sequence level attributes are repeated at each event position for output consistency with a final \texttt{END} marker explicitly indicating sequence termination. An example of the LLM input and output format is shown below.

\begin{tcolorbox}[
    title={Example of LLM input and output format},
    colback=gray!5,
    colframe=black,
    fonttitle=\bfseries,
    width=\textwidth,
    sharp corners,
    boxrule=1pt,
    breakable
]
\ttfamily\small
<INPUT>

Case attributes: \\
\hspace*{1em}case\_id = Case1 \\
\hspace*{1em}start\_time = 2012-10-09 14:50:17 \\
\hspace*{1em}start\_weekday = Tuesday \\
\hspace*{1em}start\_hour = 14 \\
\hspace*{1em}case\_length = 5 \\

Task: generate the full event trace until completion.

<OUTPUT>

1. Assign seriousness | org\_resource:Value 1 | case\_variant:Variant 12 | time\_timestamp:2012-10-09 14:50:17 \\
2. Take in charge ticket | org\_resource:Value 1 | case\_variant:Variant 12 | time\_timestamp:2012-10-09 14:51:01 \\
3. Take in charge ticket | org\_resource:Value 2 | case\_variant:Variant 12 | time\_timestamp:2012-10-12 15:02:56 \\
4. Resolve ticket | org\_resource:Value 1 | case\_variant:Variant 12 | time\_timestamp:2012-10-25 11:54:26 \\
5. Closed | org\_resource:Value 3 | case\_variant:Variant 12 | time\_timestamp:2012-11-09 12:54:39 \\
6. END
\end{tcolorbox} 

\subsection{LLM Prompting Strategy}
\label{subsec:llm_prompt}

The LLM baseline is implemented through instruction based in context prompting with few shot examples. Each prompt consists of three parts: (1) a task instruction that specifies the required sequence generation format, (2) a set of few shot input and output examples, and (3) a new sequence input for which the full sequence must be generated.

The instruction constrains the model to generate only the event sequence, preserve the required line based format, maintain chronological timestamps, generate exactly the specified number of events, and append a final \texttt{END} marker. Each few shot example follows the same \texttt{<INPUT>} and \texttt{<OUTPUT>} structure as the target sequence. The full prompt template used for the LLM baselines is shown below.

\begin{tcolorbox}[
    title={LLM prompt},
    colback=gray!5,
    colframe=black,
    fonttitle=\bfseries,
    width=\textwidth,
    sharp corners,
    boxrule=1pt,
    breakable
]
\ttfamily\small
You are an expert Predictive Process Monitoring event-log generator.

You MUST output ONLY the content between \textless TRACE\textgreater and \textless /TRACE\textgreater. \\
Anything outside these tags is strictly forbidden.

Follow these rules strictly:\\
1. Your answer MUST ONLY be the event trace, nothing else.\\
2. Output the full event trace in EXACTLY the same style as the examples.\\
3. Each line MUST start with an integer index followed by a dot, e.g. '1.' '2.' '3.'. \\
4. Each line MUST preserve attribute names and MUST follow the format:
   \textless index\textgreater. \textless activity\textgreater | \textless attr1\textgreater:\textless value1\textgreater | \textless attr2\textgreater:\textless value2\textgreater | ... | \textless timestamp\textgreater \\
5. You MUST maintain chronological timestamps.\\
6. Let case\_length be the number given in the input. You MUST produce exactly case\_length events. \\
7. After those events, you MUST output exactly one final line: (case\_length+1). END \\
8. Do NOT output any explanations, apologies, comments, or meta-text.\\
9. If information is missing or uncertain, make a reasonable guess and STILL output a full trace.\\
10. NEVER mention that information is missing. Just output the best possible trace.\\
\end{tcolorbox}

To improve comparability across sequences of different lengths, few shot examples are selected by length bucket. For each bucket, a fixed set of examples is sampled from training sequences within the same length range and reused for all holdout sequences in that bucket. The number of examples is adjusted according to bucket size, subject to predefined minimum and maximum limits.

\subsection{LLM Output Parsing and Event Log Reconstruction}
\label{subsec:llm_parse}

The generated LLM sequences are post processed into structured event logs for evaluation. Each generated response is parsed line by line, and only lines matching the expected indexed event format are retained. The parsing procedure extracts the activity label, timestamp, and all required attributes from each event line, while the final \texttt{END} marker is used to terminate the sequence.

Sequence identifiers are recovered from the corresponding input block, and event positions are reconstructed sequentially after parsing. The resulting outputs are converted into a tabular event log representation, where each row corresponds to one generated event and preserves the generated sequence identifier, event order, activity, timestamp, and available attributes.

\section{Datasets}
\label{sec:data}
We evaluate our models on six benchmark event logs: \textbf{Helpdesk} \citep{Polato2017HelpDesk}, \textbf{Sepsis} \citep{Sepsis}, \textbf{BPI13I} \citep{Steeman2013BPI}, \textbf{BPI13C} \citep{Steeman2013BPIclosed}, \textbf{BPI20} \citep{vanDongen2020BPI}, and \textbf{BPI17} \citep{BPI17}. These datasets originate from real world business processes across multiple domains. \textbf{Helpdesk} represents ticket handling in an Italian software company. \textbf{Sepsis} records hospital pathways of patients with sepsis and contains a rich attributes space. \textbf{BPI13I} (incident) and \textbf{BPI13C} (closed problem) are derived from Volvo IT’s service management system. \textbf{BPI17} captures a loan application process from a Dutch financial institution. \textbf{BPI20} (prepaid travel cost) records prepaid travel cost reimbursement requests from a Dutch university, including both domestic and international trips with multi level approvals. Together, these datasets cover diverse domains, temporal scales, sequence lengths, process complexity, and attribute richness, providing a comprehensive testbed for evaluating full sequence generation (Table~\ref{tab:data}).

\begin{table}[pos=htbp]
\centering
\setlength{\tabcolsep}{1.5pt}
\caption{Summary Statistics of Datasets}
\label{tab:data}
\begin{threeparttable}
\begin{tabularx}{\linewidth}{Xcccccccccccc}
\toprule
\textbf{Dataset} & \textbf{\#Seq.} & \textbf{\#Act.} & \textbf{Med.($l$)} & \textbf{Range($l$)} & \textbf{Med.($d$)} & \textbf{Med.($\Delta t$)} & \textbf{\#Attr.} & \textbf{\#S} & \textbf{\#M} & \textbf{\#L} & \textbf{\#XL} & \textbf{B} \\ 
\midrule
        Helpdesk & 4580 & 14 & 4.63 & 2\textasciitilde 15(13) & 3530166 & 971586 & E(1,0) $|$ S(1,0) & 2767 & 1504 & 309 & - & 5 $|$ 7 \\ 
        Sepsis & 1049 & 16 & 14.48 & 3\textasciitilde 185(52) & 2461064 & 182604 & E(1,3) $|$ S(23,1) & 156 & 681 & 212 & - & 8 $|$ 18 \\ 
        BPI13I & 7554 & 13 & 8.68 & 1\textasciitilde 123(62) & 1043768 & 135991 & E(5,0) $|$ S(3,0) & 3070 & 3558 & 926 & - & 6 $|$ 16 \\ 
        BPI20 & 2099 & 29 & 8.69 & 1\textasciitilde 21(16) & 3179104 & 413262 & E(1,0) $|$ S(2,2) & 399 & 1552 & 148 & - & 8 $|$ 12 \\ 
        BPI13C & 1487 & 7 & 4.48 & 1\textasciitilde 35(18) & 15455380 & 4442712 & E(5,0) $|$ S(3,0) & 750 & 569 & 168 & - & 4 $|$ 8 \\ 
        BPI17 & 31509 & 66 & 38.16 & 10\textasciitilde 180(145) & 1892126 & 50923 & E(5,5) $|$ S(2,1) & 2296 & 17691 & 8473 & 3049 & 21 $|$ 42 $|$ 61 \\ 
\bottomrule
\end{tabularx}
\begin{tablenotes}
\item[*] \#Seq.: number of sequences. \#Act.: number of unique activity labels. Med.($l$): median sequence length. Range($l$): minimum and maximum sequence length. Med.($d$): median sequence duration in seconds, where $d$ is defined as the elapsed time from the first event to the last event in a sequence. Med.($\Delta t$): median inter event time in seconds, where $\Delta t$ is defined as the elapsed time between two consecutive events. \#Attr.: number of attributes, reported as E(c,n)$|$S(c,n), where E and S denote event level and sequence level, and c and n denote categorical and numerical attributes. \#S, \#M, \#L, and \#XL: number of sequences in the Small, Medium, Large, and Extra Large length buckets. B: exclusive bucket upper boundaries used to define the length groups.
\item[*] In Range($l$), if only one sequence exists at the maximum length, the value in parentheses reports the second largest observed length, which corresponds to the maximum length appearing in the holdout set.
\end{tablenotes}

\end{threeparttable}
\end{table}

Table~\ref{tab:data} shows substantial heterogeneity across the six datasets. \textbf{BPI17} is the largest and most structurally complex log, with the highest number of sequences, the largest activity space, and the longest average sequence length. By contrast, \textbf{Helpdesk} and \textbf{BPI13C} contain comparatively short sequences, while \textbf{Sepsis} and \textbf{BPI13I} exhibit wider length variation, indicating more heterogeneous execution patterns. The datasets also differ markedly in temporal behavior and attribute richness. \textbf{BPI13C} has the largest average sequence duration and mean inter event gap, indicating sparse and temporally stretched executions, whereas \textbf{BPI17} combines long sequences with relatively small average inter event gaps, reflecting denser event progression. In terms of attributes, \textbf{Sepsis} is particularly rich in sequence level attributes, while \textbf{BPI17} contains the largest combined activity and event level attribute space. The bucket distributions further show that sequence lengths are unevenly distributed across logs, with some datasets dominated by short and medium sequences and \textbf{BPI17} containing substantial numbers of large and extra large sequences. This diversity creates a challenging evaluation setting for full sequence generation.

In our formulation, the event label space is defined by combining the original activity label with its corresponding \textit{lifecycle:transition} value. This choice is important because, in several real world logs, the base activity vocabulary alone does not adequately represent the effective process state. For example, in \textbf{BPI13I} and \textbf{BPI13C}, the original activity space contains only $4$ base activity types, whereas \textit{lifecycle:transition} categories substantially expand the effective label space. A similar pattern appears in \textbf{BPI17}, where transition states such as scheduled, withdrawn, and completed provide process semantics beyond the base activity label. Therefore, we adopt a joint activity and \textit{lifecycle:transition} prediction setting, consistent with recent graph based PPM models \citep{rama2021deep,rama2023embedding}. This design better reflects the multi dimensional nature of process states and supports more fine grained prediction.

\section{Experiments}
\label{sec:experiments}

All experiments were conducted on a workstation equipped with an AMD Ryzen 9 7950X CPU (16 cores, 32 threads), 30\,GB system memory, and an NVIDIA GeForce RTX 3060 GPU with 12\,GB VRAM. Each dataset was split into 90\% for model development and 10\% as a holdout test set. Within the 90\% development portion, 80\% was used for training and 20\% for validation.

\subsection{Experimental Pipeline}
\label{subsec:pipeline}

The experimental pipeline is designed to evaluate full sequence generation under consistent settings, including baseline comparison, controlled ablation analysis, and interpretability assessment. The pipeline consists of four stages. First, LLM baselines are constructed using instruction guided few shot prompting across different model and context window settings, with each model generating full event sequences from the same sequence level input information used by GGATN. Second, the proposed GGATN model is trained and used to generate full event sequences through single pass sequence generation with graph constrained structured decoding. The generated sequences from both approaches are evaluated using the same metrics to ensure comparability. Third, an ablation study is conducted to assess the stability and role of the global graph attention encoder. Specifically, we compare a frozen graph encoder, a fully trainable graph encoder, and staged unfreezing strategies in which the graph encoder is initially frozen and later updated. This analysis examines whether the graph representation acts as a stable structural prior and whether graph encoder training should be dataset dependent. Finally, interpretability analysis is performed across the main decision stages of GGATN. We analyze dual stage attention, graph grounded cross attention decomposition, refinement based activity distribution reshaping, and structured decoding based transition correction. This provides a stage wise view of how sequence context, graph conditioning, feedback refinement, and graph constrained transition selection contribute to process aware sequence generation.

\subsection{Experiment Setup}
\label{subset:setup}
This section summarizes the experimental configurations for both the LLM baselines and the proposed GGATN.

\subsubsection{LLM Model Configuration}
\label{subsec:llm_config}

Because single pass full sequence generation with graph constrained structured decoding is a new PPM setting without established baselines, we include LLM baselines to compare GGATN with locally deployable instruction prompted generative models. The comparison evaluates whether GGATN achieves performance and computational advantages over LLM based generation, and whether these advantages remain under expanded LLM settings, including larger context windows. The goal is not to exhaustively optimize LLM prompting, identify the best possible LLM configuration, or estimate the upper bound of frontier scale proprietary LLMs. Instead, the LLM experiments provide a practical and reproducible local generative baseline for assessing whether instruction prompted sequence generation can compete with graph grounded structured generation under realistic computational constraints.

All LLM experiments are conducted through a local Ollama server. In the initial stage, we evaluate two quantized instruction prompted models, \texttt{llama3.1:8b-instruct-q5\_K\_M} and \texttt{mistral:7b-instruct-q5\_K\_M}, both using a $4096$ token context window. For each target sequence, we apply a bucket conditioned $10$ shot prompting strategy, where demonstrations are selected from the same sequence length bucket as the target sequence. This design improves prompt relevance and reduces mismatch between short and long sequences.

Preliminary experiments showed that \texttt{mistral:7b-instruct-q5\_K\_M} produced substantially lower generation quality and required considerably longer runtime, with BPI17 exceeding $96$ hours. In addition, the $4096$ token context window remained restrictive for long and complex sequences. Therefore, we conducted a second stage evaluation using \texttt{llama3.1:8b-instruct-q5\_K\_M} with an extended $32,768$ token context window. Although this substantially increased computational burden, it provided a stronger LLM baseline by reducing context truncation in long horizon full sequence generation.

\subsubsection{GGATN Model Configuration}
\label{subsec:ggatn_config}

GGATN uses a shared hidden dimension of $256$. The global GAT encoder consists of $2$ edge aware GAT layers with $4$ heads and $2$ dimensional edge attributes. This $2$ layer setup allows activity embeddings to capture both direct transitions and indirect neighbors in the process graph. Each layer uses residual connections, layer normalization, and dropout $0.05$. Initial activity node embeddings are sampled from a normal distribution with scale $0.01$. The sequence Transformer encoder contains $4$ pre-layer normalization Transformer layers with $4$ heads, feed forward expansion factor $4$, and dropout $0.20$. The graph grounded cross attention module uses a single cross attention layer with the same hidden dimension and head configuration. Sequence states are layer normalized before querying the graph learned activity embeddings. The module applies gated residual injection initialized at $0.05$ and uses a feed forward expansion factor of $4$. The activity decoder uses this hybrid cosine linear formulation with a learnable logit scale, a learnable linear mixing weight initialized at $0.20$, and an enabled activity bias. Activity feedback refinement is enabled with $1$ refinement Transformer layer, feedback gate initialization $0.30$, and feedback temperature $1.0$. Transition bias is enabled with weight $0.50$. The reserved activity ids are fixed as $\texttt{PAD}=0$, $\texttt{UNK}=1$, $\texttt{EOS}=2$, and $\texttt{SOS}=3$.

During training, the activity loss assigns additional weights of $5.0$ and $3.0$ to the final and preterminal positions, respectively. The main loss weights are fixed as $\lambda_{act}=1.0$, $\lambda_{time}=0.1$, and $\lambda_{len}=0.25$. The optional latent noise branch is disabled by setting the latent dimension to $0$. A transition consistency penalty can be applied by penalizing probability mass assigned to invalid adjacent activity pairs under the process graph adjacency constraints.

Training uses AdamW with learning rate $2\times10^{-4}$, weight decay $3\times10^{-4}$, $\beta_1=0.9$, $\beta_2=0.98$, and $\epsilon=10^{-8}$. Bias and normalization parameters are excluded from weight decay. The learning rate follows a warmup cosine schedule over $300$ epochs, with warmup steps set to $5\%$ of the total training steps and a minimum learning rate multiplier of $0.1$. For the main GGATN setting, the graph encoder is frozen and only the sequence generation model is optimized. In ablation settings where the graph encoder is trainable, separate AdamW optimizers are used for the sequence model and graph encoder, with learning rates $2\times10^{-4}$ and $3\times10^{-5}$, respectively, under the same warmup cosine scheduling strategy.

\begin{table}[pos = htbp]
\centering
\caption{Dataset Dependent GGATN Configuration}
\label{tab:ggatn_config}
\begin{threeparttable}
\begin{tabular}{lccccccccccc}
\hline
\textbf{Dataset} & $\lambda_{evt}^{cat}$ & $\lambda_{evt}^{num}$ & $\lambda_{seq}^{cat}$ & $\lambda_{seq}^{num}$ & $\omega_{unary}$ & \textbf{Temp.} & \textbf{Det.} & $T_{\sigma}$ & \textbf{Bigram} & \textbf{Adj} & \textbf{Top-$k$}\\
\hline
Helpdesk & 0.80 & 0.00 & 0.50 & 0.00 & 0.55 & 1.5 & False & 0.02 & True & True & None \\
Sepsis & 0.15 & 0.10 & 0.05 & 0.05 & 0.55  & 1.5 & False & 0.02 & False & True & None \\
BPI13I & 0.15 & 0.00 & 0.05 & 0.00 & 3.00  & 1.0 & True & 0 & False & True & None\\
BPI13C & 0.15 & 0.00 & 0.05 & 0.00 & 0.55  & 1.5 & False & 0.02 & True & True & None\\
BPI20 & 0.15 & 0.00 & 0.05 & 0.05 & 0.55  & 1.5 & False & 0.02 & False & True & None\\
BPI17 & 0.15 & 0.10 & 0.05 & 0.05 & 3.00  & 1.0 & True & 0 & True & True & None \\
\hline
\end{tabular}
\begin{tablenotes}[flushleft]
\footnotesize
\item \textit{Note.} The first four columns report dataset dependent auxiliary loss weights for event and sequence level categorical and numerical attribute prediction. $\omega_{unary}$ denotes the dataset specific weight assigned to neural activity logits in Viterbi decoding.  \textit{Temp.} denotes the temperature used to rescale unary activity logits before Viterbi decoding when deterministic decoding is disabled. \textit{Det.} indicates deterministic decoding, where stochastic controls are disabled. $T_{\sigma}$ denotes the standard deviation of Gaussian noise added to transformed time predictions during non deterministic generation. \textit{Bigram} indicates whether an external bigram transition prior is used.  \textit{Adj} indicates whether adjacency masking from the global process graph is applied. Top-$k$ indicates optional filtering of activity logits before Viterbi decoding.
\end{tablenotes}
\end{threeparttable}
\end{table}

The model architecture and main training configuration are kept global across datasets to ensure comparability and avoid dataset specific training optimization, whereas generation uses dataset specific decoding controls to account for differences in sequence length, activity space, temporal scale, and transition density. These controls include deterministic or temperature based stochastic decoding, optional Top-$k$ filtering, optional bigram transition priors, and optional adjacency masking from the global process graph. The Viterbi objective combines unary activity logits, weighted by a dataset specific unary coefficient, with learned transition bias weighted by $0.50$. Time outputs are clipped to dataset specific bounds in the standardized log time space, with optional Gaussian noise during non deterministic generation. Deterministic decoding means that stochastic controls are disabled; it does not refer to autoregressive next event generation. In all settings, the final activity sequence is recovered through constrained Viterbi decoding over the activity graph. Dataset dependent weights are used only for auxiliary event and sequence attribute losses and the unary decoding weight, as summarized in Table~\ref{tab:ggatn_config}.

\begin{table}[pos = htbp]
\centering
\caption{Dataset Specific Validation Score Weights}
\label{tab:val_score_weights}
\begin{threeparttable}
\begin{tabular}{lcccccc}
\hline
\textbf{Dataset} & ActNoEOS & SeqCat & EvtCat & SeqNum & EvtNum & Time \\
\hline
Helpdesk & 0.75 & 0.03 & 0.20 & 0.00 & 0.00 & 0.02 \\
Sepsis & 0.93 & 0.01 & 0.02 & 0.01 & 0.02 & 0.01 \\
BPI13I & 0.80 & 0.03 & 0.15 & 0.00 & 0.00 & 0.02 \\
BPI13C & 0.80 & 0.03 & 0.15 & 0.00 & 0.00 & 0.02 \\
BPI20 & 0.85 & 0.02 & 0.10 & 0.02 & 0.00 & 0.01 \\
BPI17 & 0.93 & 0.01 & 0.02 & 0.01 & 0.02 & 0.01 \\
\hline
\end{tabular}
\begin{tablenotes}
\item ActNoEOS denotes activity accuracy excluding \texttt{EOS}. SeqCat and EvtCat denote sequence and event categorical accuracy. SeqNum, EvtNum, and Time are converted from RMSE to bounded scores before weighting.
\end{tablenotes}
\end{threeparttable}
\end{table}

All GGATN variants are trained for a maximum of $300$ epochs with early stopping patience set to $40$. Training uses a batch size of $64$, while validation uses a batch size of $128$. At each epoch, the model is evaluated on the validation split, and the best checkpoint is selected using a dataset specific weighted validation score that combines activity accuracy without \texttt{EOS}, event attribute accuracy, sequence attribute accuracy, numerical attribute scores, and time prediction quality. The components and dataset specific weights are reported in Table~\ref{tab:val_score_weights}. Training history, validation metrics, and the best model checkpoint are saved for subsequent evaluation. For interpretability analysis, attention tensors are extracted from the best validation checkpoint, including global GAT attention, Transformer self-attention, and graph grounded cross attention weights. Aggregation is component specific. Transformer self-attention is averaged across all layers and heads to summarize cumulative sequence contextualization. For the global GAT encoder, the analysis uses only the final layer attention, which reflects the most task adapted structural representation. Graph grounded cross attention is implemented as a single layer, so its attention weights are analyzed directly.

\subsection{Evaluation Metrics}
\label{subsec:eva}
The primary evaluation uses sequence and control flow metrics: sequence similarity (SS), Damerau Levenshtein similarity (DL), bigram Jensen Shannon distance (JSD), and duration Wasserstein distance (WD). Sequence coverage is included because full sequence generation requires a valid and parsable output for each target sequence. We also measure aggregated event level and sequence level attribute performance, activity accuracy, and timestamp mean absolute error. Generation validity is assessed through activity vocabulary diagnostics, including hallucinated activities, activity recall, and sequence level attribute consistency. Detailed attribute level performance and sequence variable consistency results are provided in \ref{app:full_results}. All metrics are computed under an unconditional evaluation protocol, where the full set of holdout sequences is considered. Missing, empty, or unparsable generated outputs are treated as failed generations and penalized accordingly in the metric computation.

\subsubsection{Sequence and Control Flow Metrics}
\label{susbsubsec:control_metr}

For each generated sequence, activities are ordered by generated position, whereas ground truth activities are ordered by timestamp. Sequence coverage is defined as
\[
\mathrm{Cov}=\frac{|\mathcal{S}_{gen}\cap \mathcal{S}_{gt}|}{|\mathcal{S}_{gt}|},
\]
where $\mathcal{S}_{gen}$ and $\mathcal{S}_{gt}$ denote the generated and ground truth sequence identifier sets.

Sequence similarity (SS) is computed as the normalized matching ratio between generated and ground truth activity sequences:
\[
\mathrm{SS}
=
\frac{1}{|\mathcal{S}_{gt}|}
\sum_{\sigma_i \in \mathcal{S}_{gt}}
r(\sigma_i,\hat{\sigma}_i),
\]
where $r(\cdot,\cdot)$ denotes the normalized sequence matching ratio. If no valid generated sequence exists for $\sigma_i$, then $r(\sigma_i,\hat{\sigma}_i)=0$.

Damerau Levenshtein similarity (DL) measures edit based activity sequence similarity:
\[
\mathrm{DL}(\sigma,\hat{\sigma})=
1-\frac{d_{DL}(\sigma,\hat{\sigma})}{\max(|\sigma|,|\hat{\sigma}|)},
\]
where $d_{DL}$ denotes the Damerau Levenshtein distance.
The reported DL score is averaged over all ground truth sequences. If no valid generated sequence exists for $\sigma_i$, its DL similarity is set to $0$.

Bigram Jensen Shannon distance (JSD) compares the normalized distributions of adjacent activity pairs in the generated and ground truth logs. For each log, activity bigrams are extracted from each sequence and aggregated into a normalized frequency distribution. Let $P_{bg}$ and $Q_{bg}$ denote the generated and ground truth bigram distributions, respectively. JSD is computed as the Jensen Shannon distance between $P_{bg}$ and $Q_{bg}$, where lower values indicate closer control flow distributional similarity at the log level. To penalize failed generation, we report a coverage adjusted score:
\[
\mathrm{JSD}^{*}
=
\mathrm{JSD}
+
\lambda_{cov}(1-\mathrm{Cov}),
\]
where $\lambda_{cov}=1.0$ in all experiments. This assigns a unit penalty to missing sequences, so generation failure is reflected in the distributional score rather than being ignored.

Duration Wasserstein distance (WD) compares the distributions of sequence durations, where duration is measured as the time difference between the first and last events in a sequence. To penalize failed generation, we report a coverage adjusted score:
\[
\mathrm{WD}^{*}
=
\mathrm{WD}
+
(1-\mathrm{Cov})\cdot \mathrm{median}(D_{gt}),
\]
where $D_{gt}$ denotes the ground truth sequence duration distribution. This assigns each missing generated sequence a penalty proportional to the typical temporal scale of the dataset.

These metrics evaluate complementary requirements of full sequence generation. Coverage captures output completeness, SS measures direct activity order agreement, and DL provides an edit based similarity measure that accounts for insertions, deletions, substitutions, and adjacent transpositions. At the log level, bigram JSD evaluates preservation of local control flow transition distributions, while WD measures temporal plausibility through sequence duration distributions. The coverage adjusted variants prevent models from achieving favorable distributional scores by generating only a subset of easier sequences. Overall, the metric set jointly captures completeness, activity order fidelity, edit robustness, transition distribution similarity, and temporal plausibility.

\subsubsection{Activity, Temporal, and Aggregated Attribute Performance Metrics}
\label{subsubsec:agg-metrics}

Attribute and timestamp metrics are computed after aligning generated and ground truth events by sequence identifier and within sequence event position. Event level categorical and Boolean attributes are evaluated using accuracy, while event level numerical attributes are evaluated using mean absolute error. Activity accuracy is computed under the same aligned event level setting. Sequence level attributes are collapsed to one value per generated sequence, with inconsistent generated values treated as missing. Sequence level categorical and Boolean attributes are evaluated using accuracy, and sequence level numerical attributes are evaluated using mean absolute error. Event level and sequence level scores are averaged over their corresponding attributes. Timestamp quality is measured by mean absolute error in seconds.

For unconditional evaluation, all ground truth events and sequences are preserved. Missing categorical or Boolean predictions are counted as incorrect. Missing numerical attributes and timestamps are assigned penalty values equal to the maximum observed valid absolute error for the corresponding variable; if no valid matched error exists, a fallback penalty of $1.0$ is used. These localized metrics complement sequence and control flow metrics by assessing whether generated activities, timestamps, and attributes remain consistent after positional alignment. They are reported as auxiliary indicators of event level and sequence level reconstruction quality, rather than standalone measures of full sequence generation performance.

\subsubsection{Generation Validity Metrics}
\label{subsubsec:val-metrics}

Generation validity is evaluated using activity vocabulary diagnostics and sequence level attribute consistency. Hallucinated activities are counted as generated activity labels that do not appear in the ground truth activity vocabulary. Activity recall is defined as the proportion of ground truth activity labels that appear at least once in the generated log. For sequence level attributes, consistency is measured by checking whether each generated sequence assigns a single value to each sequence level variable across all generated event rows. We report the aggregate inconsistency ratio across sequence level variables, where lower values indicate better consistency.

\section{Results}
\label{sec:Res}
Table~\ref{tab:result} reports full sequence generation performance across the six datasets, including primary generation metrics and averaged attribute results. Complete dataset specific metric tables, including individual event level and sequence level attribute scores, are provided in \ref{app:full_results}.

Before discussing the results, we define the table abbreviations. \textbf{M} denotes model, where \textbf{G} is the main GGATN model with a frozen GAT attention encoder, \textbf{G\_j} is GGATN with fully unfrozen joint training, and \textbf{G\_s5} and \textbf{G\_s10} are staged training variants in which the GAT attention encoder is frozen for the first $5$ and $10$ epochs, respectively, and unfrozen afterwards. \textbf{L(4k)} and \textbf{L(32k)} denote Llama with 4k and 32k context windows, respectively, while \textbf{M(4k)} denotes Mistral with a 4k context window. \textbf{c(S)} denotes sequence coverage, \textbf{JSD} bigram Jensen Shannon distance, \textbf{SS} sequence similarity, \textbf{WD} duration Wasserstein distance, and \textbf{DL} Damerau Levenshtein similarity. \textbf{H} denotes the number of hallucinated activities, \textbf{re(A)} activity recall, and \textbf{Ac(A)}, \textbf{Ac(E)}, and \textbf{Ac(S)} activity accuracy, averaged event level attribute accuracy, and averaged sequence level attribute accuracy, respectively. \textbf{MA(E)}, \textbf{MA(S)}, and \textbf{MA(T)} denote averaged event level numerical MAE, averaged sequence level numerical MAE, and timestamp MAE, respectively. \textbf{I(c)} and \textbf{I(n)} denote aggregate sequence level inconsistency rate indicators for categorical and numerical attributes, and \textbf{\#S} denotes the number of generated sequences. For JSD, WD, MAE based metrics, and hallucinated activities, lower values are better; for coverage, SS, DL, recall, and accuracy metrics, higher values are better. Bold values indicate the best result for each dataset and metric.

\begin{table}[pos = htbp]
\centering
\caption{Performance Comparison across Different Datasets and Models}
\label{tab:result}
\footnotesize
\setlength{\tabcolsep}{3pt}
\begin{tabularx}{\textwidth}{@{} l *{16}{C} @{}}
\toprule
\multicolumn{17}{c}{\textbf{Helpdesk}} \\
\midrule
\textbf{M} & \textbf{c(S)} & \textbf{JSD} & \textbf{SS} & \textbf{WD} & \textbf{DL} & \textbf{H} & \textbf{re(A)} & \textbf{Ac(A)} & \textbf{Ac(E)} & \textbf{MA(E)} & \textbf{Ac(S)} & \textbf{MA(S)} & \textbf{MA(T)} & \textbf{I(c)} & \textbf{I(n)} & \textbf{\#S} \\
\midrule
\textbf{G} & 1.0000 & 0.1779 & 0.9498 & 502069 & 0.9276 & 0 & 0.6250 & 0.8969 & 0.4114 & & \textbf{0.7380} & & 589028 & 0 & & 458 \\
\textbf{G\_j} & 1.0000 & \textbf{0.1724} & 0.9502 & 481164 & 0.9280 & 0 & 0.6250 & 0.8974 & 0.4073 & & \textbf{0.7380} & & 615277 & 0 & & 458 \\
\textbf{G\_s5} & 1.0000 & 0.1872 & \textbf{0.9505} & 711122 & \textbf{0.9289} & 0 & 0.6250 & \textbf{0.8992} & \textbf{0.4156} & & 0.7358 & & 641977 & 0 & & 458 \\
\textbf{G\_s10} & 1.0000 & 0.1853 & 0.9496 & \textbf{433090} & 0.9275 & 0 & 0.6250 & 0.8951 & 0.4142 & & 0.7358 & & \textbf{586232} & 0 & & 458 \\
\textbf{L(4k)} & 1.0000 & 0.4297 & 0.6488 & 8446087 & 0.4876 & 2 & \textbf{1.0000} & 0.2821 & 0.0782 & & 0.0044 & & 1102316 & 1 & & 458 \\
\textbf{M(4k)} & 0.8908 & 0.3492 & 0.7619 & 4539597 & 0.7257 & 71 & 0.7500 & 0.6130 & 0.1252 & & 0.3734 & & 66648683 & 1 & & 408 \\
\textbf{L(32k)} & 1.0000 & 0.2561 & 0.8960 & 5254845 & 0.8526 & 1 & 0.8750 & 0.7662 & 0.1551 & & 0.1179 & & 2478145 & 0 & & 458 \\
\midrule
\multicolumn{17}{c}{\textbf{BPI20}} \\
\midrule
\textbf{M} & \textbf{c(S)} & \textbf{JSD} & \textbf{SS} & \textbf{WD} & \textbf{DL} & \textbf{H} & \textbf{re(A)} & \textbf{Ac(A)} & \textbf{Ac(E)} & \textbf{MA(E)} & \textbf{Ac(S)} & \textbf{MA(S)} & \textbf{MA(T)} & \textbf{I(c)} & \textbf{I(n)} & \textbf{\#S} \\
\midrule
\textbf{G} & 1.0000 & \textbf{0.2012} & 0.8856 & 1814361 & 0.8422 & 0 & 0.6667 & 0.7999 & 0.8286 &  & 0.5861 & 811.72 & 1334086 & 0& 0 & 209 \\
\textbf{G\_j} & 1.0000 & 0.2033 & 0.8854 & \textbf{1462130} & 0.8429 & 0 & 0.6667 & 0.8047 & 0.8312 &  & 0.6029 & 820.74 & \textbf{1331992} & 0 & 0 & 209 \\
\textbf{G\_s5} & 1.0000 & 0.2054 & \textbf{0.8962} & 1914524 & \textbf{0.8583} & 0 & 0.6667 & \textbf{0.8153} & \textbf{0.8376} &  & 0.5957 & \textbf{794.31} & 1336757 & 0 & 0 & 209 \\
\textbf{G\_s10} & 1.0000 & 0.2108 & 0.8899 & 1826933 & 0.8506 & 0 & 0.6667 & 0.8100 & 0.8339 &  & \textbf{0.6220} & 782.03 & 1348357 & 0 & 0 & 209\\
\textbf{L(4k)} & 0.756 & 0.7147 & 0.3398 & 2707083 & 0.2964 & 421 & 0.5926 & 0.1476 & 0.1783 &  & 0.1747 & 12547.55 & 74200704 & 1 & 1 & 158 \\
\textbf{M(4k)} & 0.3158 & 1.3999 & 0.1003 & 4011717 & 0.0882 & 227 & 0.5556 & 0.0552 & 0.0398 &  & 0.0359 & 4227.15 & 55912405 & 1 & 1 & 66 \\
\textbf{L(32k)} & 1.0000 & 0.2584 & 0.8211 & 1987999 & 0.7396 & 19 & \textbf{0.8519} & 0.5435 & 0.6253 &  & 0.4857 & 1265.46 & 3012057 & 0 & 1 & 209 \\
\midrule
\multicolumn{17}{c}{\textbf{Sepsis}} \\
\midrule
\textbf{M} & \textbf{c(S)} & \textbf{JSD} & \textbf{SS} & \textbf{WD} & \textbf{DL} & \textbf{H} & \textbf{re(A)} & \textbf{Ac(A)} & \textbf{Ac(E)} & \textbf{MA(E)} & \textbf{Ac(S)} & \textbf{MA(S)} & \textbf{MA(T)} & \textbf{I(c)} & \textbf{I(n)} & \textbf{\#S} \\
\midrule
\textbf{G} & 1.0000 & 0.2977 & 0.7169 & 1089448 & 0.6271 & 0 & 0.8125 & 0.4816 & 0.7311 & 20.02 & 0.8177 & \textbf{12.77} & 446981 & 0 & 0 & 104 \\
\textbf{G\_j} & 1.0000 & 0.3046 & 0.7184 & 1073339 & 0.6314 & 0 & 0.8125 & 0.4884 & 0.7273 & 20.05 & 0.8131 & 13.08 & 345565 & 0 & 0 & 104 \\
\textbf{G\_s5} & 1.0000 & \textbf{0.2921} & \textbf{0.7321} & 1361757 & 0.6401 & 0 & 0.8125 & 0.4878 & 0.7367 & 20.10 & 0.8181 & 12.97 & \textbf{334917} & 0 & 0 & 104 \\
\textbf{G\_s10} & 1.0000 & 0.3004 & 0.7272 & \textbf{894149} & \textbf{0.6490} & 0 & 0.8125 & \textbf{0.4991} & \textbf{0.7473} & \textbf{19.75} & \textbf{0.8207} & 12.97 & 621142 & 0 & 0 & 104 \\
\textbf{L(4k)} & 0.6346 & 0.9833 & 0.1061 & 1353228 & 0.0867 & 320 & 0.6250 & 0.0269 & 0.0350 & 73.51 & 0.1463 & 48.41 & 22346748 & 1 & 1 & 66 \\
\textbf{M(4k)} & 0.1635 & 1.5442 & 0.0201 & 2984395 & 0.0155 & 623 & 0.6250 & 0.0119 & 0.0113 & 353.71 & 0.0727 & 44.60 & 22264333 & 1 & 1 & 17 \\
\textbf{L(32k)} & 0.8654 & 0.5240 & 0.5181 & 1670784 & 0.4467 & 24 & \textbf{0.9375} & 0.2489 & 0.3258 & 81.46 & 0.6158 & 31.15 & 14337061 & 1 & 1 & 90 \\
\midrule
\multicolumn{17}{c}{\textbf{BPI13C}} \\
\midrule
\textbf{M} & \textbf{c(S)} & \textbf{JSD} & \textbf{SS} & \textbf{WD} & \textbf{DL} & \textbf{H} & \textbf{re(A)} & \textbf{Ac(A)} & \textbf{Ac(E)} & \textbf{MA(E)} & \textbf{Ac(S)} & \textbf{MA(S)} & \textbf{MA(T)} & \textbf{I(c)} & \textbf{I(n)} & \textbf{\#S} \\
\midrule
\textbf{G} & 1.0000 & 0.2945 & \textbf{0.8237} & 12449630 & \textbf{0.8017} & 0 & 0.8333 & 0.6933 & 0.3697 & & 0.4054 & & \textbf{8915835} & 0 & & 148 \\
\textbf{G\_j} & 1.0000 & 0.3412 & 0.8113 & 12326789 & 0.7984 & 0 & 0.8333 & \textbf{0.6975} & \textbf{0.3782} & & 0.4032 & & 9321481 & 0 & & 148 \\
\textbf{G\_s5} & 1.0000 & 0.3228 & 0.8166 & 12825437 & 0.7999 & 0 & 0.8333 & 0.6933 & 0.3756 & & \textbf{0.4122} & & 9073631 & 0 & & 148 \\
\textbf{G\_s10} & 1.0000 & 0.3385 & 0.8113 & 12790696 & 0.7972 & 0 & 0.8333 & 0.6891 & 0.3762 & & 0.4099 & & 9127482 & 0 & & 148 \\
\textbf{L(4k)} & 0.8311 & 0.5903 & 0.4840 & 14198777 & 0.4681 & 162 & \textbf{1.0000} & 0.3081 & 0.0784 & & 0.1374 & & 181447374 & 1 & & 123 \\
\textbf{M(4k)} & 0.6149 & 0.9099 & 0.3645 & 13194896 & 0.3528 & 63 & 0.8333 & 0.1765 & 0.0305 & & 0.0946 & & 41116259 & 1 & & 91 \\
\textbf{L(32k)} & 1.0000 & \textbf{0.2071} & 0.7939 & \textbf{11440188} & 0.7566 & 1 & 0.8333 & 0.6204 & 0.1160 & & 0.2635 & & 10513514 & 1 & & 148 \\
\midrule
\multicolumn{17}{c}{\textbf{BPI13I}} \\
\midrule
\textbf{M} & \textbf{c(S)} & \textbf{JSD} & \textbf{SS} & \textbf{WD} & \textbf{DL} & \textbf{H} & \textbf{re(A)} & \textbf{Ac(A)} & \textbf{Ac(E)} & \textbf{MA(E)} & \textbf{Ac(S)} & \textbf{MA(S)} & \textbf{MA(T)} & \textbf{I(c)} & \textbf{I(n)} & \textbf{\#S} \\
\midrule
\textbf{G} & 1.0000 & \textbf{0.2133} & \textbf{0.8325} & 748775 & 0.8190 & 0 & \textbf{0.9167} & 0.6552 & \textbf{0.3738} &  & 0.3731 &  & 459020 & 0 &  & 755 \\
\textbf{G\_j} & 1.0000 & 0.2547 & 0.8301 & 855279 & \textbf{0.8232} & 0 & 0.7500 & 0.6603 & 0.3695 &  & 0.3757 &  & 459483 & 0 &  & 755 \\
\textbf{G\_s5} & 1.0000 & 0.2586 & 0.8289 & \textbf{624653} & 0.8218 & 0 & 0.7500 & \textbf{0.6630} & 0.3649 &  & 0.3854 &  & \textbf{454386} & 0 &  & 755 \\
\textbf{G\_s10} & 1.0000 & 0.2208 & 0.8307 & 652814 & 0.8229 & 0 & \textbf{0.9167} & 0.6540 & 0.3655 &  & \textbf{0.3885} &  & 454590 & 0 &  & 755 \\
\textbf{L(4k)} & 0.7642 & 0.7407 & 0.3714 & 950069 & 0.3585 & 1133 & 0.6667 & 0.1462 & 0.0494 & & 0.1020 & & 266296975 & 1 & & 577 \\
\textbf{M(4k)} & 0.3510 & 1.3093 & 0.0833 & 1113165 & 0.0698 & 708 & 0.7500 & 0.0421 & 0.0236 & & 0.0406 & & 5338296 & 1 & & 265 \\
\textbf{L(32k)} & 1.0000 & 0.2605 & 0.7582 & 897646 & 0.7227 & 8 & 0.7500 & 0.5007 & 0.1547 & & 0.1995 & & 1175517 & 1 & & 755 \\
\midrule
\multicolumn{17}{c}{\textbf{BPI17}} \\
\midrule
\textbf{M} & \textbf{c(S)} & \textbf{JSD} & \textbf{SS} & \textbf{WD} & \textbf{DL} & \textbf{H} & \textbf{re(A)} & \textbf{Ac(A)} & \textbf{Ac(E)} & \textbf{MA(E)} & \textbf{Ac(S)} & \textbf{MA(S)} & \textbf{MA(T)} & \textbf{I(c)} & \textbf{I(n)} & \textbf{\#S} \\
\midrule
\textbf{G} & 1.0000 & 0.2238 & 0.7526 & 418785 & 0.6352 & 0 & \textbf{0.8833} & 0.3353 & 0.4997 & 2991.16 & 0.5792 & 11547.99 & 429831 & 0 & 0 & 3150 \\
\textbf{G\_j} & 1.0000 & 0.2157 & 0.7542 & 445256 & 0.6359 & 0 & \textbf{0.8833} & 0.3366 & \textbf{0.5013} & 3024.25 & 0.5813 & 11375.47 & \textbf{419696} & 0 & 0 & 3150 \\
\textbf{G\_s5} & 1.0000 & \textbf{0.2140} & 0.7531 & 539976 & 0.6339 & 0 & \textbf{0.8833} & 0.3319 & 0.4995 & 3099.29 & \textbf{0.5848} & \textbf{10479.63} & 432232 & 0 & 0 & 3150 \\
\textbf{G\_s10} & 1.0000 & 0.2243 & \textbf{0.7551} & \textbf{375357} & \textbf{0.6379} & 0 & 0.8667 & \textbf{0.3386} & 0.5007 & \textbf{2980.69} & 0.5805 & 11342.45 & 521810 & 0 & 0 & 3150 \\
\textbf{L(4k)} & 0.7419 & 0.9814 & 0.0461 & 1724366 & 0.0323 & 13195 & 0.5167 & 0.0023 & 0.0112 & 78770.87 & 0.0437 & 176838.04 & 216544974 & 1 & 1 & 2337 \\
\textbf{M(4k)} & 0.2251 & 1.5629 & 0.0073 & 3530745 & 0.0041 & 5110 & 0.5333 & 0.0007 & 0.0201 & 71698623.83 & 0.0157 & 281286.40 & 30360903 & 1 & 1 & 709 \\
\textbf{L(32k)} &0.5889  & 0.9157 & 0.1424	 & 5691146 & 0.1125 & 4321	 & 0.7833 &0.042	 & 0.0689	& 74352.56	& 0.1686 & 249466.68	 & 223751169	&1 &1 	&1855\\
\bottomrule
\end{tabularx}
\end{table}

Overall, the GGATN variants achieve the strongest and most stable performance across datasets. They obtain full sequence coverage, generate no hallucinated activities, and maintain zero inconsistency for sequence level attributes on all datasets. They also consistently outperform the LLM baselines in sequence similarity and Damerau Levenshtein similarity. These results indicate that the structured graph based model is substantially more reliable for end to end full sequence generation.

\subsection{Performance Patterns and Dataset Specific Effects}
\label{subsec:performance_patterns}
The advantage of GGATN is especially clear on the primary sequence and control flow metrics. Across Helpdesk, BPI20, Sepsis, BPI13I, and BPI17, GGATN variants obtain the best values on JSD, SS, WD, and DL. The only exception is BPI13C. On this dataset, GGATN still achieves stronger SS and DL, higher activity accuracy, better event level and sequence level attribute accuracy, zero hallucinated activities, and zero sequence level inconsistency, whereas the 32k context Llama obtains the best JSD and WD.

This exception is mainly explained by the data characteristics of BPI13C. It is the sparsest dataset in the benchmark, with a median sequence length of $4.5$ and a length range from $1$ to $35$, while also having the largest temporal span, with a median sequence duration of $179$ days and a median inter event time of $9$ days. Under this setting, JSD and WD can favor models that reproduce coarse log level statistics, because JSD compares adjacent activity pair distributions and WD compares marginal duration distributions. Neither metric requires the generated sequence to match the ground truth event order, activity positions, or attributes. The LLM setting also receives the absolute initial timestamp as text, together with start weekday and start hour giving it a richer calendar level temporal anchor for duration generation. This can improve coarse duration distribution matching on BPI13C without implying better sequence level fidelity.

This behavior may also be related to the different decoding mechanisms. The LLM generates events autoregressively, so each event is conditioned on the previously generated prefix. In a sparse dataset with short median sequences, this stepwise mechanism can reproduce frequent adjacent transitions and coarse duration patterns effectively, directly improving JSD and WD. By contrast, GGATN performs single pass generation with Viterbi style graph constrained decoding, optimizing a globally valid activity path under learned unary and transition scores. This favors structural validity, sequence alignment, and hallucination control, but can be less sensitive to sparse local transition frequencies or highly dispersed duration patterns when these statistics are weakly represented in the learned global objective.
 
A further factor is the evaluation asymmetry induced by dataset size and length conditioning. BPI13C is relatively small, and some length buckets contain only around $150$ training sequences. The LLM baseline generates within length buckets, meaning that the prompt provides explicit length conditioned examples. Therefore, 10 shot prompting represents a relatively strong prior for BPI13C, especially in short sparse buckets where few examples can cover many frequent local patterns. GGATN does not receive such bucket specific examples during generation; it must generate from the learned representation and decoding constraints cross different lengths. Therefore, the BPI13C result should not be read as a reversal of the overall trend. The 32k context LLM appears to better approximate the sparse local transition and duration distributions of this particular dataset, but GGATN remains stronger on sequence level alignment, event level reconstruction, attribute consistency, and hallucination control. 

The LLM baselines show clear limitations under the evaluation protocol. The 4k context settings frequently exhibit reduced coverage, hallucinated activities, low activity accuracy, and poor timestamp accuracy, especially on BPI20, Sepsis, BPI13I, and BPI17. Mistral is particularly unstable, with low coverage and weak sequence similarity on several datasets. Increasing the Llama context window to 32k improves coverage and sequence similarity, but the improvement remains insufficient to close the gap with GGATN on most datasets. The BPI17 results further show that these gains are not uniform, as several metrics decline under the 32k setting, indicating unstable behavior on long and complex sequence structures. By contrast, GGATN variants maintain consistently strong performance across datasets. Their generation process is guided by learned sequence representations, graph based transition constraints, and structured decoding rather than by prompt length or context window size. In several cases, the LLM baselines obtain higher activity recall, but this is often accompanied by lower activity accuracy and more hallucinated activities, indicating broader but less controlled activity generation. GGATN instead achieves a stronger balance between activity coverage and correctness. Attribute and temporal metrics further support this pattern. GGATN variants generally obtain higher event level and sequence level accuracy while maintaining zero sequence level inconsistency across datasets. Overall, the proposed structured model provides more reliable full sequence generation by preserving control flow quality together with stronger alignment among activities, timestamps, and attributes.

\subsection{Computational Cost and Metric Interpretation}
\label{subsec:cost_metric_interpretation}

Computational cost further separates GGATN from the LLM baselines. GGATN is substantially smaller and completes BPI17 generation in less than four hours, whereas Mistral with a 4k context requires more than 96 hours and Llama with a 32k context requires more than 240 hours. The same pattern holds on the remaining datasets: GGATN typically completes generation within 20 minutes, while the LLM baselines require at least 30 minutes and often several hours. This shows that the proposed structured model achieves stronger generation quality with much lower inference cost, making it more practical for repeated evaluation and deployment on event sequence generation tasks.

The activity and attribute results should be interpreted in light of the full sequence generation setting. Unlike next event prediction, GGATN does not condition on a ground truth prefix at each generation step. It produces latent representations for all positions in a single pass and decodes the activity path under graph based transition constraints. Consequently, local event level accuracy is stricter than global sequence similarity, especially for long and high cardinality datasets such as BPI17, where small local deviations can accumulate across many positions. This effect is visible in the activity and attribute metrics: GGATN substantially outperforms the LLM baselines in activity accuracy, event level accuracy, and sequence level accuracy, but the absolute values remain lower on long and complex logs.

The multi task design also introduces tradeoffs among activity, timestamp, event attribute, and sequence attribute heads, particularly on attribute rich datasets such as Sepsis. This reflects the difficulty of jointly aligning activities, timestamps, and attributes at every generated event position under a full sequence generation setting. Nevertheless, the stronger SS and DL scores indicate that the generated sequences remain globally close to the target sequences even when exact event level matches are harder to obtain. We observed a similar distinction in additional training analyses. Validation under teacher forcing improved local validation metrics, but did not materially improve final generation quality when the same Viterbi based decoding procedure was used during generation. Conversely, preliminary experiments that incorporated teacher forcing into the generation procedure degraded performance across datasets. These observations suggest that full sequence generation should be evaluated primarily through global sequence quality, coverage, transition realism, and temporal distribution metrics rather than local position wise accuracy alone.

\section{Ablation Study}
\label{sec:ablation}
The ablation study examines the stability and role of the global GAT encoder in GGATN. The graph encoder is constructed from the global process graph and produces activity embeddings that serve as structural memory for graph grounded cross attention, activity prototype decoding, and constrained generation. When the graph encoder is trainable, both the activity embeddings and graph attention weights can adapt to the full sequence generation objective. This setting allows us to assess whether the global graph representation remains a stable structural prior within the full pipeline. If performance varies substantially across graph encoder training settings, the graph representation is sensitive to optimization strategy and therefore less reliable as a structural prior. Conversely, if performance remains within a narrow range, with only marginal improvements or degradations, the global graph design can be interpreted as a stable structural foundation. The remaining question is whether the graph encoder should follow a uniform training strategy or be adjusted according to dataset characteristics.
\begin{figure}[pos=htbp]
  \centering
  \begin{tabular}{cc}
    \includegraphics[width=0.45\textwidth]{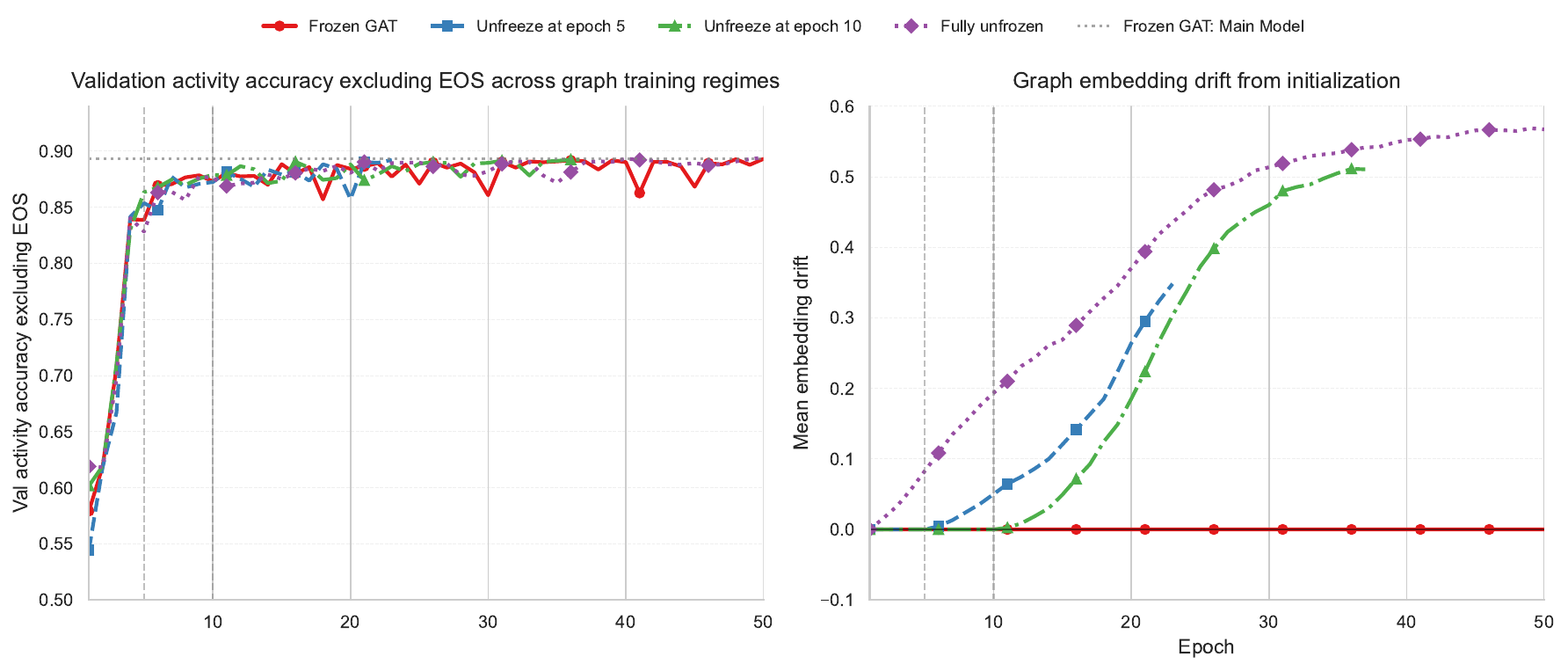} &
    \includegraphics[width=0.45\textwidth]{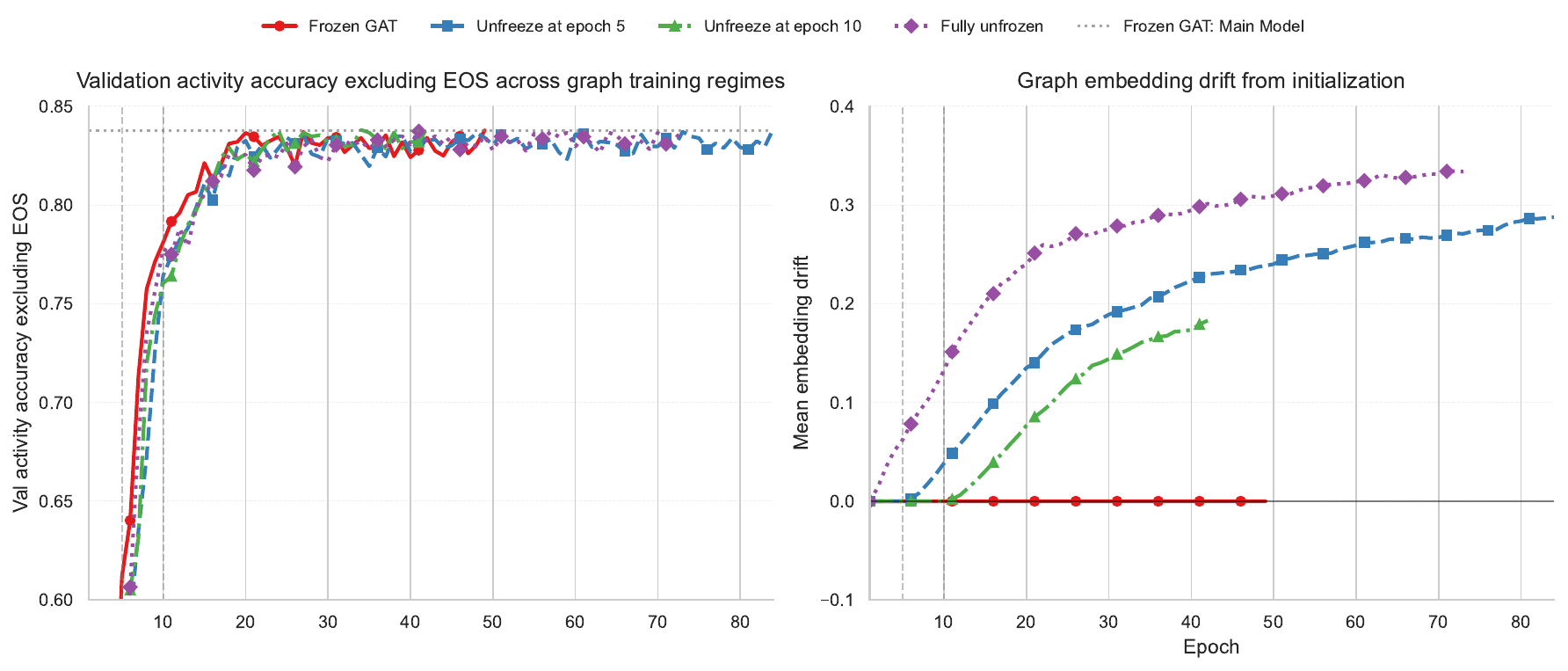} \\
    {\footnotesize\textbf{(a) Helpdesk}} & {\footnotesize\textbf{(b) BPI20}} \\

    \includegraphics[width=0.45\textwidth,trim={0 0 0 19.5mm},clip]{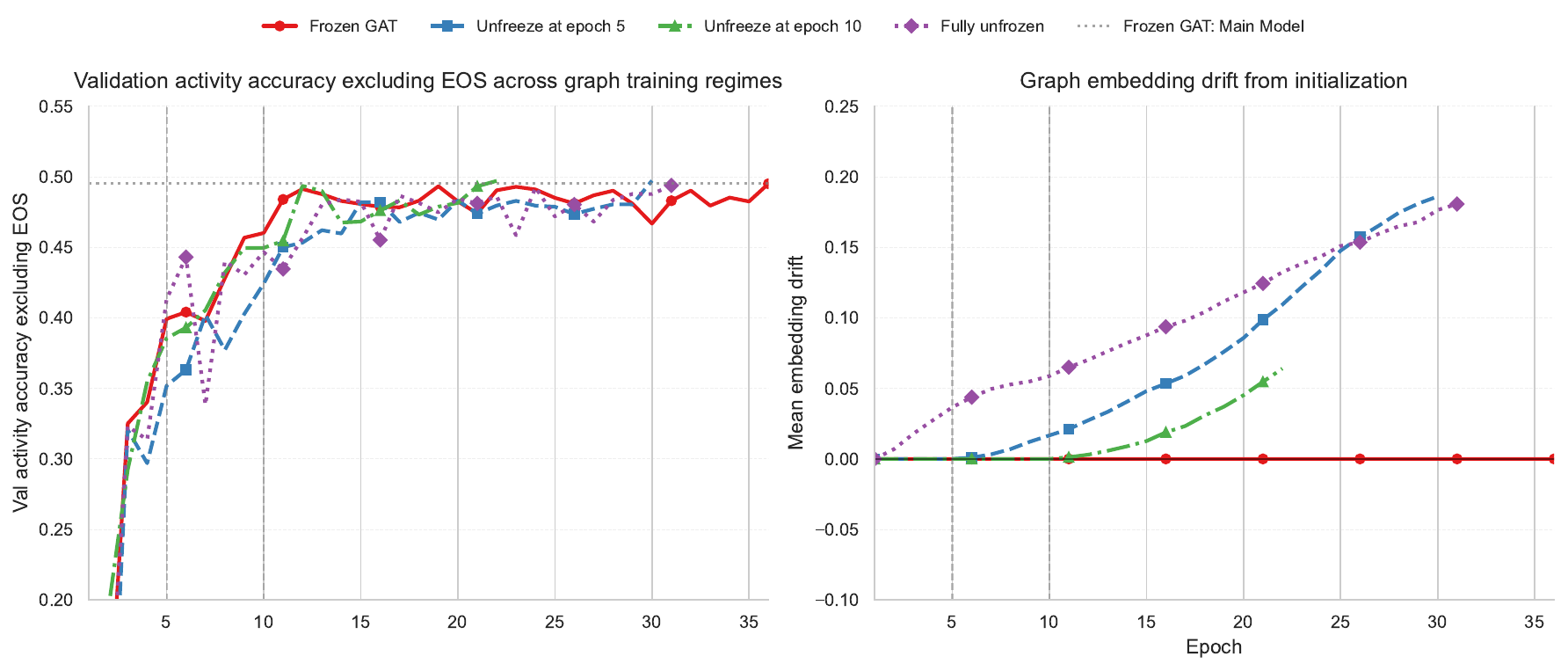} &
    \includegraphics[width=0.45\textwidth,trim={0 0 0 19.5mm},clip]{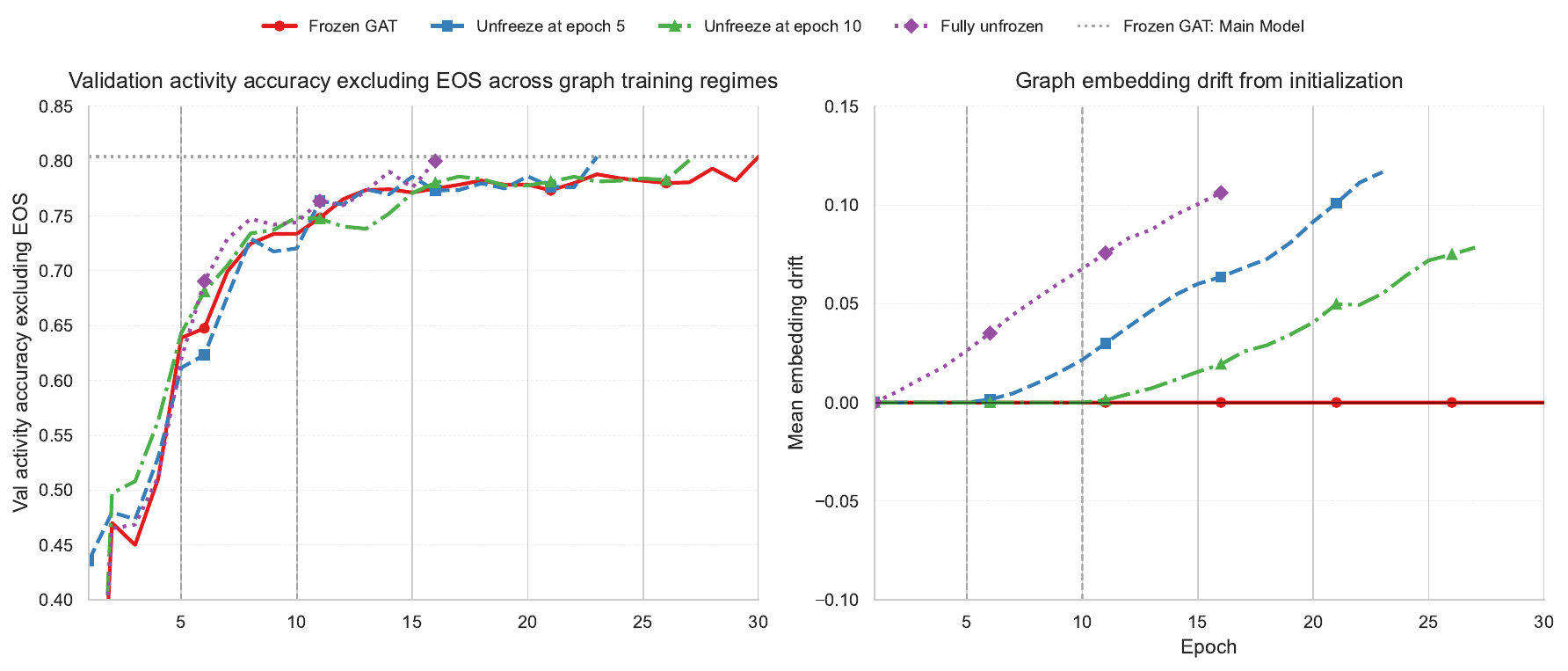} \\
    {\footnotesize\textbf{(c) Sepsis}} & {\footnotesize\textbf{(d) BPI13C}} \\

    \includegraphics[width=0.45\textwidth,trim={0 0 0 19.5mm},clip]{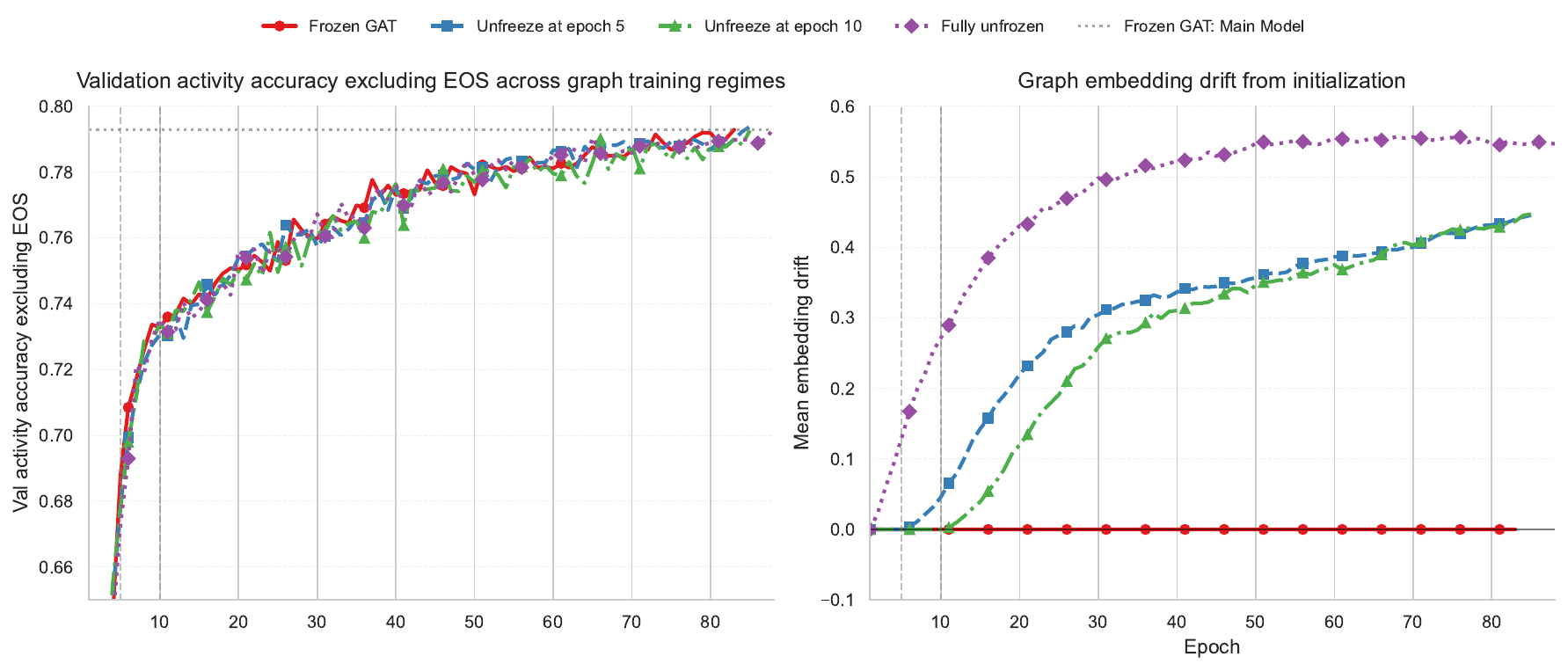} &
    \includegraphics[width=0.45\textwidth,trim={0 0 0 19.5mm},clip]{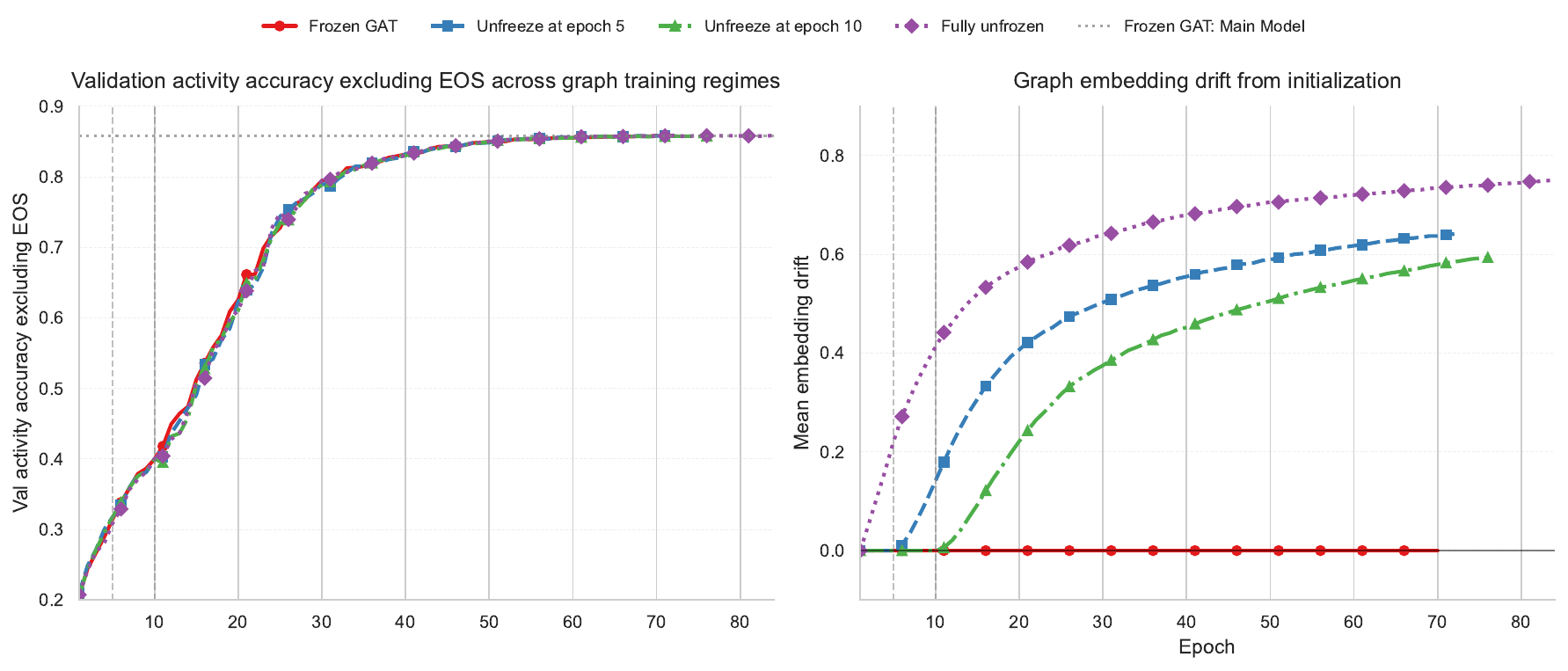} \\
    {\footnotesize\textbf{(e) BPI13I}} & {\footnotesize\textbf{(f) BPI17}}
  \end{tabular}
  \caption{Ablation Analysis of GAT Encoder Training Regimes across Six Datasets}
  \label{fig:ablation_curve}
    \vspace{-2mm}
\end{figure}
 
\begin{figure}[pos=htbp]
  \centering
  \includegraphics[width=0.98\textwidth]{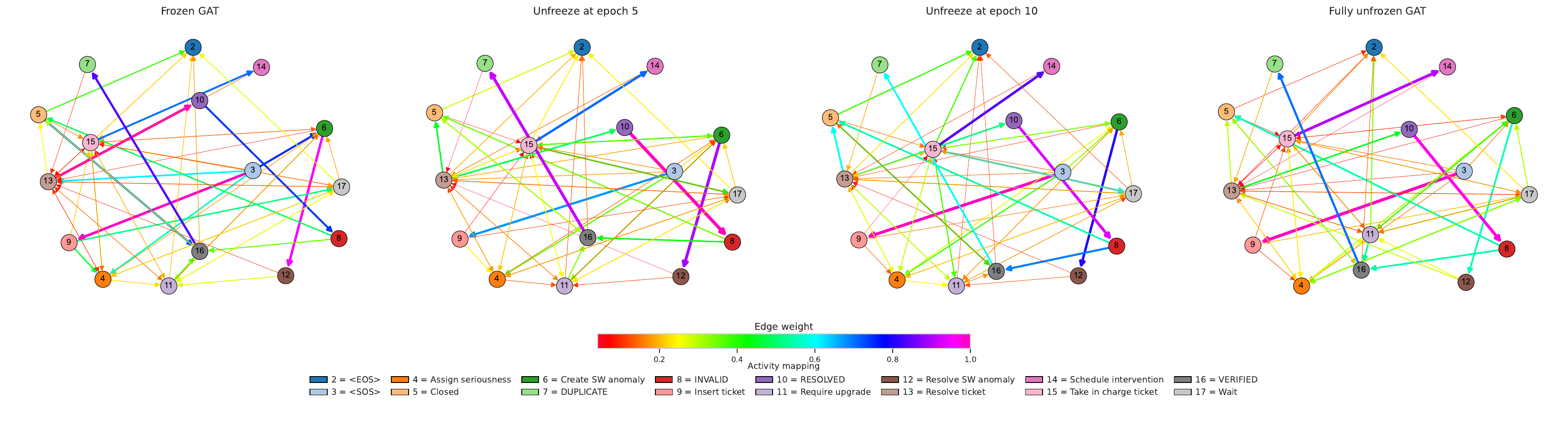}\\[-4mm]
  {\footnotesize\textbf{(a) Helpdesk}}\\
  \includegraphics[width=0.98\textwidth]{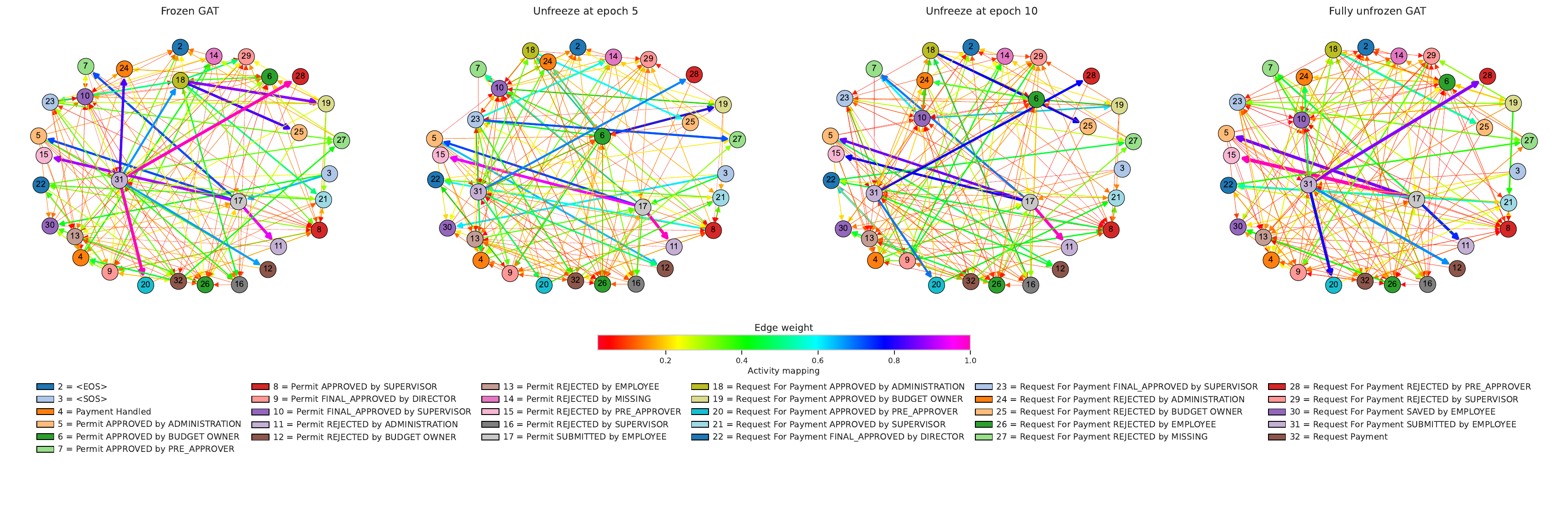}\\[-4mm]
  {\footnotesize\textbf{(b) BPI20}}\\
  \includegraphics[width=0.98\textwidth]{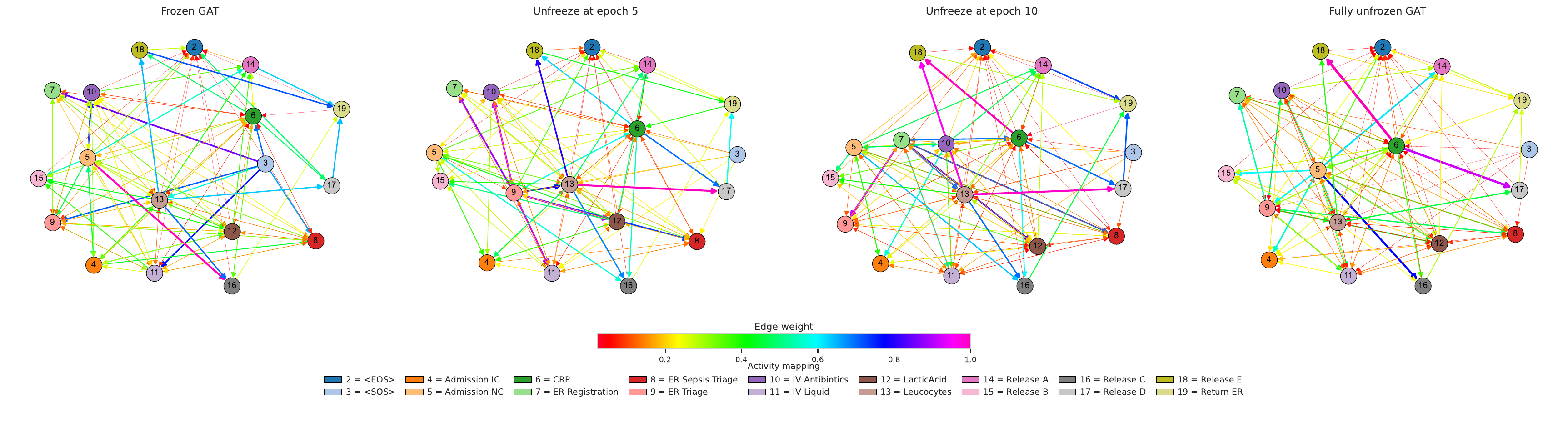}\\[-4mm]
  {\footnotesize\textbf{(c) Sepsis}}
  \caption{Ablation Analysis of GAT Attention Structure Shifts for Four Training Regimes across Helpdesk, BPI20, and Sepsis Datasets}
  \label{fig:ablation_att}
\end{figure}

We compare four GGATN variants. In the frozen setting, used as the main model, the graph encoder is fixed while the remaining GGATN modules are trained. In the fully trainable setting, the graph encoder and the remaining modules are optimized jointly from the beginning, allowing activity embeddings and graph attention weights to adapt throughout training. In the staged settings, the graph encoder is frozen for the first $5$ or $10$ epochs and then unfrozen, so training first uses fixed graph representations before moving to joint optimization. These variants are denoted as \textbf{G}, \textbf{G\_j}, \textbf{G\_s5}, and \textbf{G\_s10}, respectively, in Table~\ref{tab:result}.

\begin{figure}[pos=htbp]
  \centering

  \includegraphics[width=0.98\textwidth]{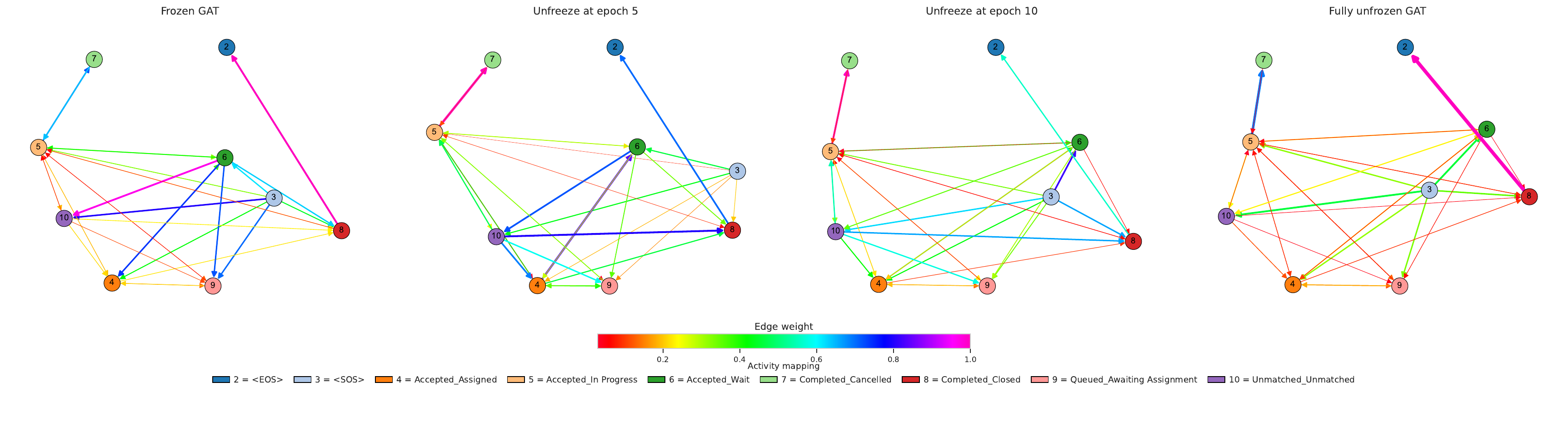}\\[-4mm]
  {\footnotesize\textbf{(d) BPI13C}}\\

  \includegraphics[width=0.98\textwidth]{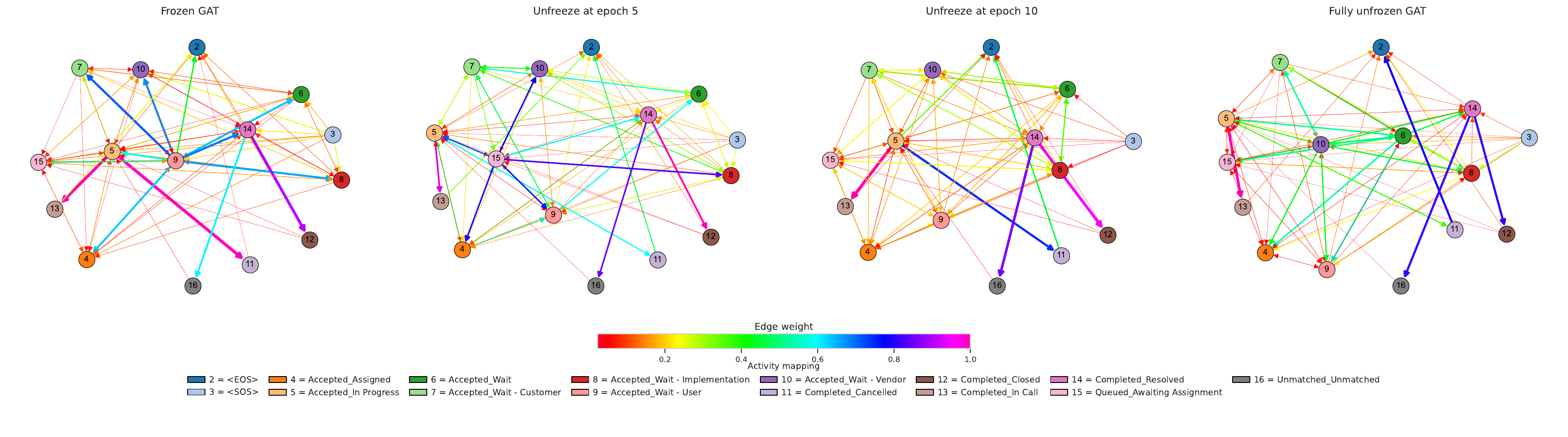}\\[-4mm]
  {\footnotesize\textbf{(e) BPI13I}}\\
  \includegraphics[width=0.98\textwidth]{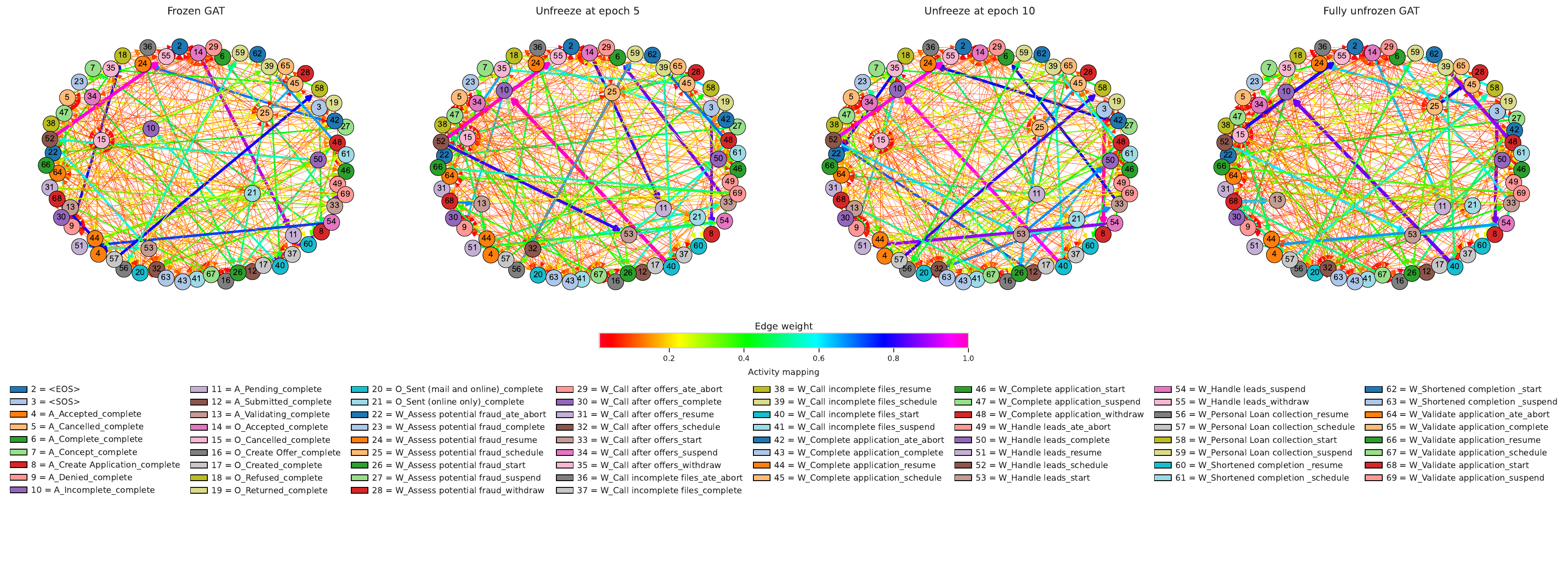}\\[-4mm]
  {\footnotesize\textbf{(f) BPI17}}
  \caption{Ablation Analysis of GAT Attention Structure Shifts for Four Training Regimes across BPI13C, BPI13I and BPI17 Datasets}
  \label{fig:ablation_att_p2}
\end{figure}

To examine whether graph encoder updates change validation behavior, Figure~\ref{fig:ablation_curve} reports validation activity accuracy excluding \texttt{EOS} together with the embedding drift of graph activity representations across datasets. The drift curves measure how far activity embeddings move from their initialization during training. Validation activity accuracy excluding \texttt{EOS} is used because it directly reflects event label prediction influenced by graph representations, whereas generation metrics such as SS, DL, JSD, and WD also depend on decoding and log level aggregation. Thus, generation metrics are less suitable for isolating graph encoder dynamics. The accuracy curves characterize validation dynamics during training, while final generation quality is reported separately in Table~\ref{tab:result}. Each subplot traces training from initialization to the epoch of the selected model. Dataset specific axis ranges are used to show the separation and convergence of the four training regimes, together with the full scale of embedding drift. The attention panels in Figures~\ref{fig:ablation_att} and~\ref{fig:ablation_att_p2} provide qualitative evidence of how graph attention weights are redistributed under frozen, staged, and fully trainable regimes.

As expected, the frozen model shows zero embedding drift in Figure~\ref{fig:ablation_curve}, whereas the fully trainable and staged variants exhibit increasing drift once the graph encoder is updated. However, validation activity accuracy remains close across the four regimes, even when learned graph embeddings move substantially. This indicates that graph encoder adaptation changes the embedding space, but does not consistently produce large validation gains. Fixed global graph representations therefore provide a stable structural basis for graph grounded cross attention, activity prototype decoding, and constrained generation. Although graph encoder adaptation can be beneficial in some cases, comparable validation behavior shows that stable predictive quality does not depend on continuous graph encoder updates.

The embedding drift curves quantify the magnitude of representation change, while the attention panels in Figures~\ref{fig:ablation_att} and~\ref{fig:ablation_att_p2} indicate where these changes occur in the global graph. Across datasets, attention weights are visibly redistributed when the graph encoder is unfrozen, showing that adaptive training reallocates importance over activity transitions. This reallocation occurs without substantial variation in validation activity accuracy. This suggests that adaptation refines attention over an already informative graph structure, rather than changing the structural role of the graph within GGATN. For example, in Helpdesk, the attention weight on transition $13 \rightarrow 10$ decreases as the graph encoder becomes more adaptive, whereas transition $15 \rightarrow 14$ receives higher attention. In BPI13I, by contrast, the attention weight on transition $5 \rightarrow 11$ varies across frozen, staged, and fully unfrozen regimes without a consistent direction. These observations show that attention reallocation is transition specific and dataset dependent, rather than uniformly directional. The key point is that such internal rearrangements do not produce performance instability, supporting the use of the global GAT encoder as a stable structural component within GGATN.

Table~\ref{tab:result} shows the same pattern at the final evaluation level. Across datasets, the four GGATN variants show highly similar performance, with most primary metric differences remaining below $0.05$. This indicates that the global GAT encoder provides a stable structural component whose contribution is not dependent on a specific training regime. However, the preferred regime is not uniform across datasets. On the sparse BPI13C and BPI13I datasets, the frozen graph encoder achieves stronger results on several metrics, suggesting that fixed graph representations can preserve a stable structural prior when observed transitions are limited. In contrast, BPI17, which has the largest activity space and longest average sequence length, benefits more from adaptive graph encoder training, with staged or fully trainable variants improving several sequence, attribute, and temporal metrics. These observations indicate that freezing provides a stable default, while joint or staged adaptation may be useful depending on sequence length, activity space size, and transition complexity.

\section{Interpretability Analysis}
\label{sec:interpre}

The interpretability analysis examines GGATN along the full generation pipeline rather than treating the model as a black box. The four analyses correspond to the main decision stages of the architecture. First, the dual stage attention panel shows how Transformer self-attention builds sequence context and how graph grounded cross attention injects activity level structural evidence when forming position level activity predictions. Second, the graph grounded cross attention decomposition analysis examines graph attention from two perspectives: its alignment with transition admissibility in the global process graph and its attention to the true and predicted activities at each sequence position. Third, the refinement analysis examines how provisional activity distributions are reshaped before final decoding through soft activity feedback. Finally, the structured decoding analysis examines how structured decoding converts raw local activity preferences into process aware sequence decisions through graph based transition constrained path selection. Together, these analyses connect internal attention, graph conditioning, distribution reshaping, and final path selection, providing a stage wise mechanistic account of how GGATN produces process aware sequence generation. For each analysis, one or two representative examples are reported in the main text, while the remaining dataset specific interpretability plots are provided in \ref{app:inter}.

\subsection{Position Level Dual Stage Attention Analysis}
\label{subsec:dualstage}

\begin{figure}[pos=htbp]
  \centering

  \includegraphics[width=0.98\textwidth]{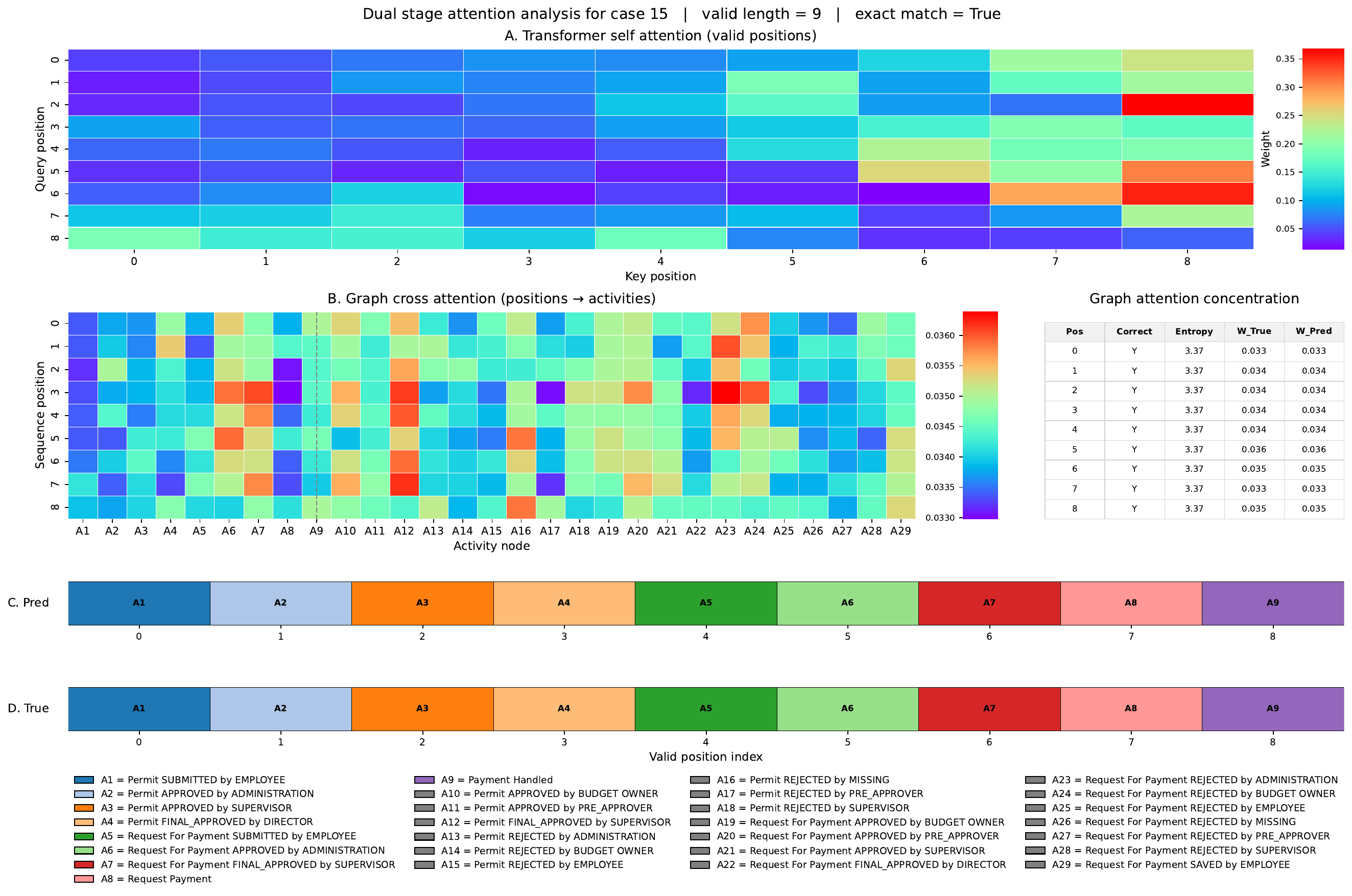}\\[-4mm]
  {\footnotesize\textbf{(a) BPI20}}\\
  
  \includegraphics[width=0.98\textwidth]{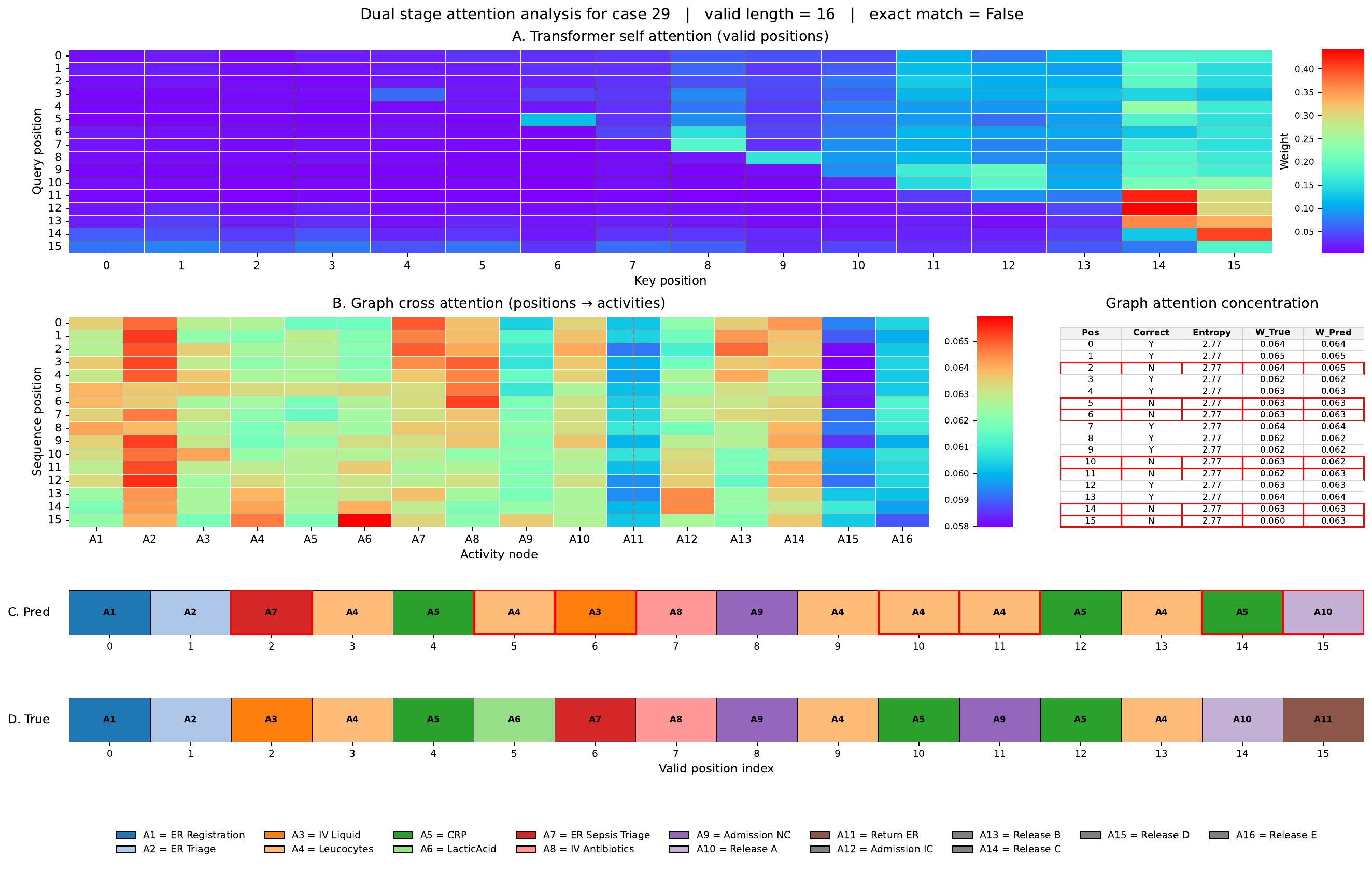}\\[-4mm]
  {\footnotesize\textbf{(b) Sepsis}}\\

  \caption{Dual Stage Attention Analysis Panel of Exactly Match Sample (BPI20) and Non Exactly Match Sample (Sepsis)}
  \label{fig:dual_attn}
\end{figure}

Figure~\ref{fig:dual_attn} presents a dual stage attention analysis for representative generated sequences. The purpose of this visualization is to examine how GGATN combines sequential context with graph grounded activity information when forming position level activity predictions. Transformer self-attention is averaged over all heads and layers, while graph grounded cross attention is averaged over heads in its single cross attention layer. Thus, each panel reports the mean attention pattern for the selected sequence. The representative sequences are selected from valid generated outputs with sufficient length and nontrivial activity variation, and include both exact match and non exact match examples. Exact match is a strict sequence level indicator that holds only when all valid predicted activities match the ground truth sequence.

Panel A reports Transformer self-attention over valid sequence positions. Panel B reports graph grounded cross attention from sequence positions to activity nodes in the global process graph, together with a concentration summary that includes entropy and the attention weights assigned to the true and predicted activities at each position. Entropy is computed from the row normalized graph grounded cross attention distribution over valid activity nodes, with higher values indicating more distributed attention and lower values indicating stronger concentration on fewer activity nodes. Panels C and D compare the predicted and ground truth activity sequences, with mismatched positions highlighted. The figure links the two internal attention stages with the final activity output, allowing prediction behavior to be inspected at the position level rather than only through aggregate metrics.

The two examples illustrate different prediction outcomes under the same dual stage attention mechanism. In the BPI20 example, the selected sequence has valid length $9$ and achieves an exact match. Panels C and D are identical, indicating that all valid activity positions are correctly predicted. Panel A shows a clear sequence contextualization pattern rather than local neighbor attention: earlier query positions assign stronger attention to later key positions, whereas later query positions emphasize earlier positions. This indicates that Transformer self-attention integrates information across the full valid sequence when forming position representations. Panel B shows that graph grounded cross attention behaves differently from positional self-attention. It does not follow a monotonic sequence pattern or a one to one alignment between positions and activity nodes. Instead, attention is distributed over a broad set of graph nodes, with position specific variation across the activity space. The concentration summary supports this reading: entropy remains high, and the weights assigned to the true and predicted activities vary across positions. Since the sequence is predicted correctly at every position, correct generation does not require sharply concentrated graph attention. Rather, graph grounded cross attention provides distributed structural conditioning that is combined with sequence contextualization and activity decoding to produce the final activity sequence.

In the Sepsis example, the selected sequence has valid length $16$ and does not achieve an exact match. Several positions are incorrectly predicted, as indicated by the red outlines in Panel C and the corresponding rows in the concentration summary. Panel A shows a markedly different self-attention pattern from the BPI20 example. Attention is shifted toward later key positions across most query positions, with the strongest responses concentrated in the final part of the sequence. This indicates that the sequence representation relies strongly on late sequence context. Panel B remains structurally informative despite the prediction errors. Graph grounded cross attention is not uniformly spread over the full activity space. Instead, the subset of graph nodes corresponding to activities that appear in the true sequence receives consistently stronger attention than the remaining activity nodes. Thus, even in a non exact match case, graph grounded cross attention still emphasizes process relevant regions of the global graph rather than irrelevant activities. The concentration summary further clarifies the error pattern. At correctly predicted positions, the weights assigned to the true and predicted activities are identical. At mismatched positions, these values diverge because the predicted label differs from the true label. However, both \texttt{W\_True} and \texttt{W\_Pred} remain within a relatively narrow range and do not span the full range of graph grounded cross attention values in Panel B. This suggests that incorrect predictions are not caused by a complete loss of graph support for the true activity. Instead, the graph component continues to provide structured but distributed evidence, while final errors arise from how this evidence is combined with sequence representation and decoding.

Overall, the examples show that graph grounded cross attention functions as distributed structural conditioning rather than hard activity selection. In both exact and non exact cases, it preserves process relevant graph evidence, while final prediction correctness depends on how this structural evidence interacts with sequence self-attention and decoding.

\subsection{Graph Grounded Cross Attention Decomposition Analysis}
\label{subsec:atten-ta}

\begin{figure}[pos=htbp]
  \centering
  
  \begin{subfigure}[b]{0.48\textwidth}
    \centering
    \includegraphics[width=\linewidth, trim={0 0 0 7mm}, clip]{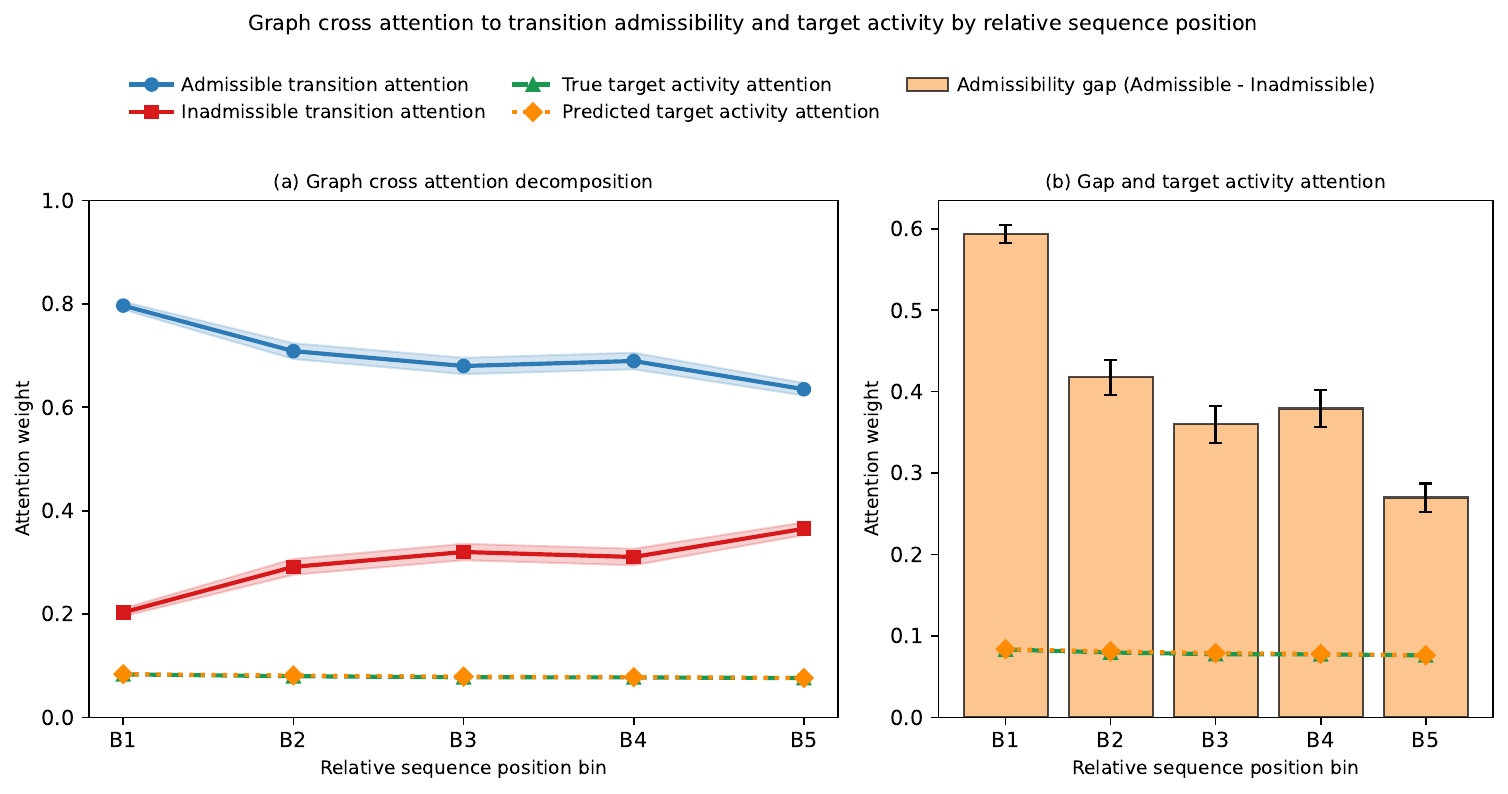}
    \label{subfig:attnw_a}
  \end{subfigure}
  \hfill
  \begin{subfigure}[b]{0.48\textwidth}
    \centering
    \includegraphics[width=\linewidth, trim={0 0 0 7mm}, clip]{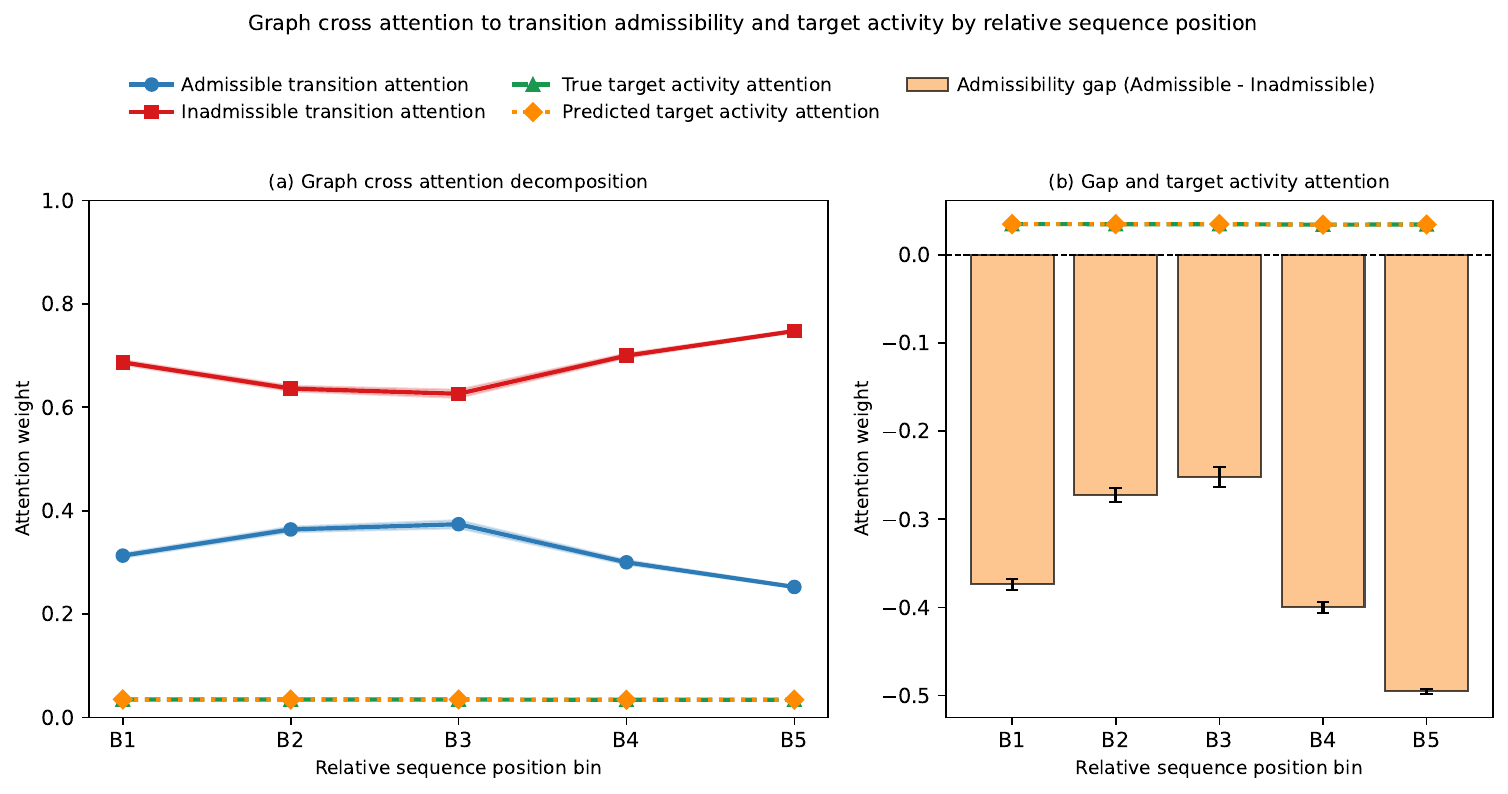}
    \label{subfig:attnw_b}
  \end{subfigure}  \\
  \vspace{-8pt}
  \footnotesize\textbf{(a) Graph Grounded Cross Attention to Transition Admissibility and Target Activity for BPI13I (left) and BPI20 (right)}
  
    \begin{subfigure}[b]{0.48\textwidth}
    \centering
    \includegraphics[width=\linewidth]{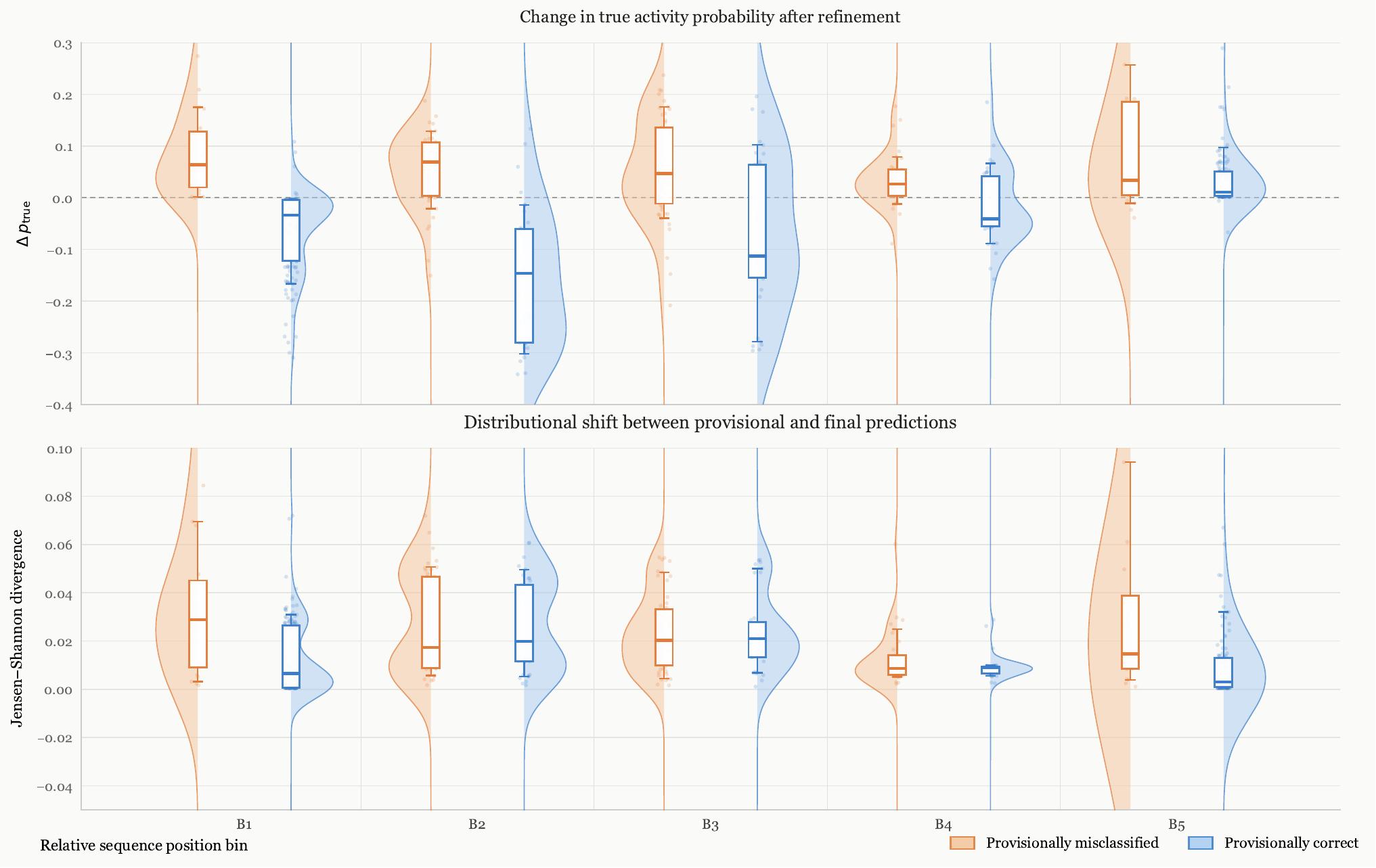}
    \label{subfig:refine_c}
  \end{subfigure}
  \hfill
  \begin{subfigure}[b]{0.48\textwidth}
    \centering
    \includegraphics[width=\linewidth]{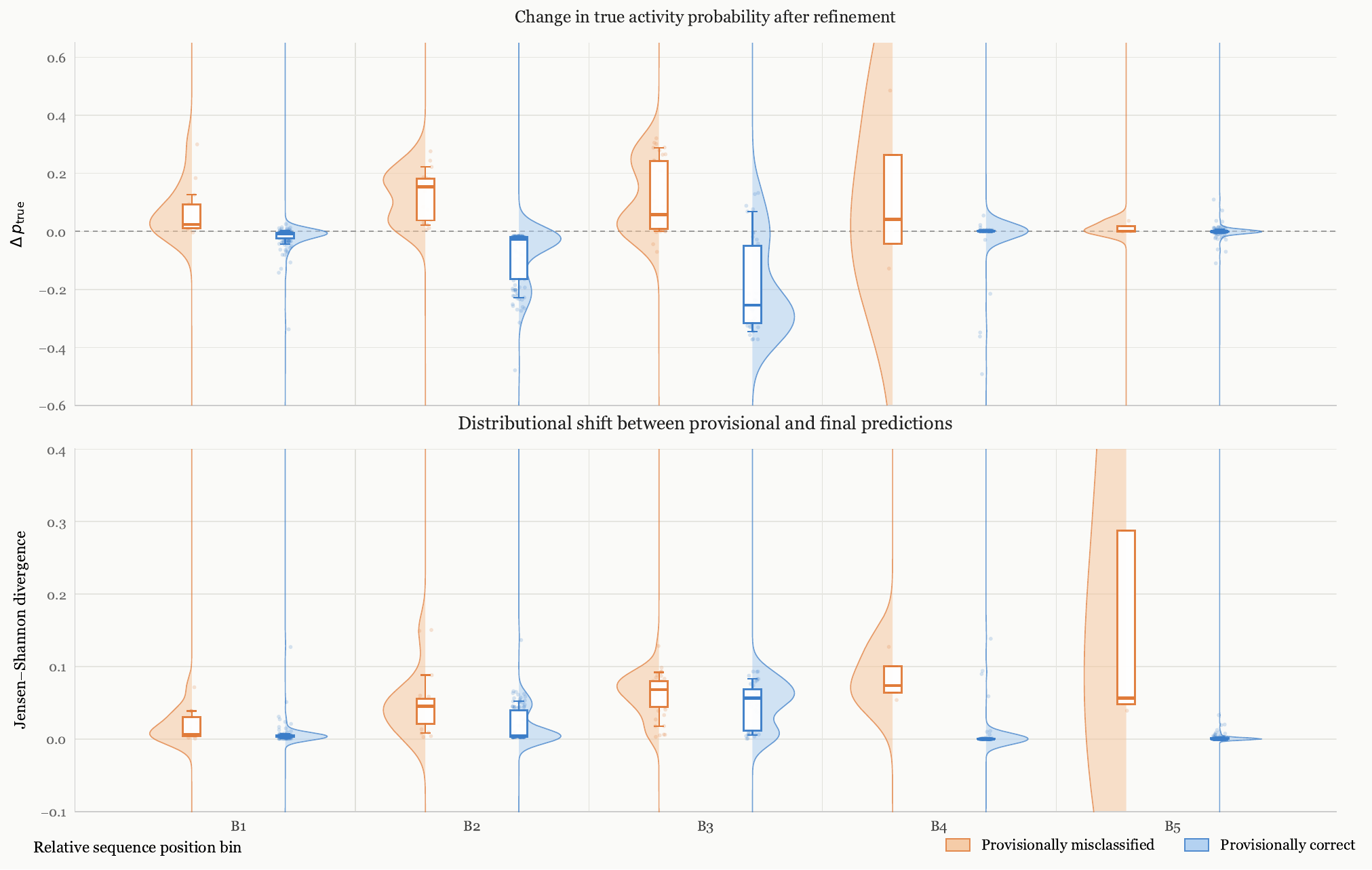}
    \label{subfig:refine_d}
  \end{subfigure}  \\
  \vspace{-8pt}
  \footnotesize\textbf{(b) Refinement Based Activity Distribution Reshaping for BPI13C (left) and Helpdesk (right)}
 
 \begin{subfigure}[b]{\textwidth}
    \centering
    \includegraphics[width=\linewidth, trim={20mm 20mm 20mm 20mm}, clip]{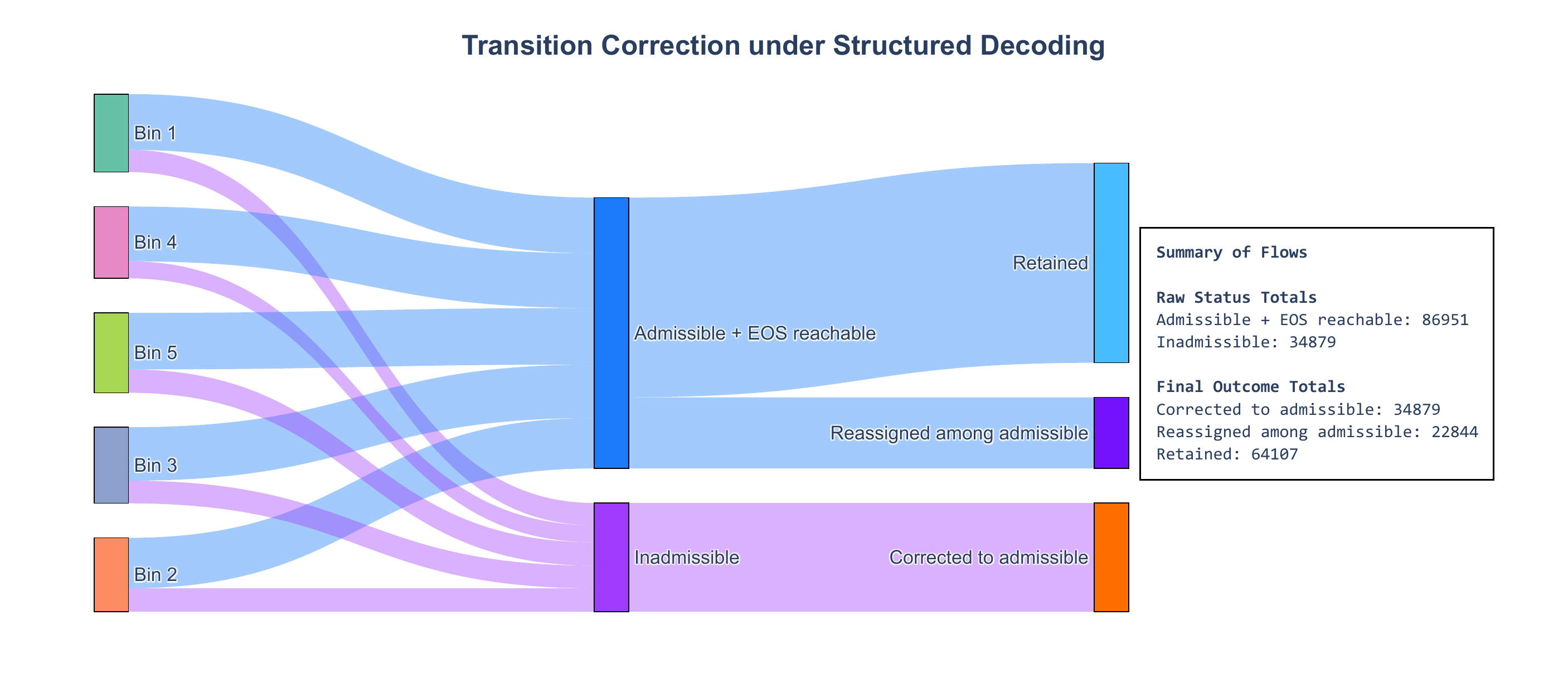}
    \label{subfig:dec_e}
  \end{subfigure}\\
    \vspace{-8pt}
  \footnotesize\textbf{(c) Transition Correction under Graph Constrained Structured Decoding for BPI17}
  
  % --- Main caption for the entire figure ---
  \caption{Stage Wise Interpretability Analysis of GGATN}
  \label{fig:inter_5plots}
\end{figure}

Figure~\ref{fig:inter_5plots}(a) analyzes graph grounded cross attention from two perspectives: its alignment with transition admissibility in the global process graph and its attention to the ground truth and model selected activities at the following sequence position. The purpose is to determine whether the graph attention signal supports admissible activity transitions, disperses over broader graph regions, or separates between ground truth and model selected activities during full sequence generation. The analysis is aggregated by relative source position, where valid source positions in each sequence are divided into five ordered bins from early to late regions. For each valid source position, the activity at that position defines the admissible outgoing transitions in the observed process graph. We decompose the graph attention distribution into four quantities: attention assigned to admissible outgoing activities, attention assigned outside this admissible set, attention assigned to the ground truth activity at the following position, and attention assigned to the model selected activity at the following position. The admissibility gap is defined as admissible transition attention minus inadmissible transition attention. Positive values indicate that graph attention favors admissible outgoing activities, whereas negative values indicate stronger attention outside the admissible transition set.

This analysis does not treat graph grounded cross attention as a hard admissibility filter. In GGATN, graph grounded cross attention produces the structural representation $\mathbf{H}_{struct}$, which is decoded against the graph activity embeddings $\mathbf{E}_{act}$ and then passed through graph constrained decoding. The purpose of this figure is therefore to examine whether learned graph attention directly aligns with admissible outgoing activities or provides broader structural conditioning before the final decoding constraint. Strong positive gaps indicate direct alignment with admissible transitions, while weak or negative gaps indicate that graph attention is used more broadly as distributed structural evidence rather than as a local transition filter.

Figure~\ref{fig:inter_5plots}(a) shows two contrasting patterns. In BPI13I, graph attention is strongly aligned with admissible transitions. Across all relative position bins, admissible transition attention remains substantially higher than inadmissible transition attention, producing a consistently positive gap. The gap is largest in the first bin and remains positive through later bins, although it gradually decreases. This indicates that graph grounded cross attention places greater weight on process admissible outgoing activities throughout the sequence. The ground truth and model selected activity attention curves are also nearly identical, showing that the graph signal assigned to the model selected activity at the following position closely follows the signal assigned to the ground truth activity at that position.

BPI20 presents the opposite pattern. Inadmissible transition attention dominates admissible transition attention across all bins, yielding a consistently negative gap. At the same time, the ground truth and model selected activity attention curves remain close to each other and stay near the lower range of the graph attention scale. This indicates that, in a larger and more complex activity space with $29$ activities, graph grounded cross attention does not operate as a local admissible transition selector. Instead, it provides distributed structural conditioning that is combined with graph constrained decoding to support final activity selection.

\subsection{Refinement Based Activity Distribution Reshaping Analysis}
\label{subsect:refine}
Figure~\ref{fig:inter_5plots}(b) analyzes the role of the refinement block by comparing the provisional activity distribution with the final activity distribution. GGATN first produces provisional activity logits, converts the resulting distribution into a soft activity embedding, and injects this feedback into the hidden representation before producing the final activity logits. This design introduces an explicit activity distribution reshaping stage before structured decoding, allowing the model to revise position level activity distributions rather than relying only on the first decoded logits.

For each valid non \texttt{EOS} position, we compute the change in probability assigned to the ground truth activity, $\Delta p_{\mathrm{true}} = p_{\mathrm{true}}^{final} - p_{\mathrm{true}}^{first}$, and the Jensen Shannon (JS) divergence between the provisional and final activity distributions. Positions are grouped by relative sequence position and separated according to whether the provisional prediction is correct or misclassified. The figure therefore captures both the directional effect of refinement on the true activity and the magnitude of distributional change induced by the feedback stage.

The two examples in Figure~\ref{fig:inter_5plots}(b) show that refinement does not behave as a uniformly corrective operation. In BPI13C, provisionally misclassified positions generally show positive $\Delta p_{\mathrm{true}}$ across relative position bins, indicating that refinement often increases the probability assigned to the ground truth activity after the first pass. Provisionally correct positions, in contrast, frequently show negative or near zero $\Delta p_{\mathrm{true}}$, especially in the middle bins. This suggests that refinement does not simply amplify already correct predictions. Instead, it redistributes probability mass after the provisional prediction, increasing support for many initially misclassified positions while recalibrating some initially correct ones. The JS divergence panel supports this interpretation. In BPI13C, provisionally misclassified positions show visible distributional shifts across bins, while provisionally correct positions also exhibit nontrivial shifts in several regions. Thus, refinement changes the full activity distribution, not only the probability assigned to the true activity. This is consistent with the feedback design of GGATN, where the provisional distribution is converted into a soft activity embedding and used to update the hidden representation before final decoding.

The Helpdesk example shows a more selective refinement pattern. For provisionally misclassified positions, $\Delta p_{\mathrm{true}}$ is positive in several bins, especially in the middle and later parts of the sequence, but the effect is not uniform. For provisionally correct positions, $\Delta p_{\mathrm{true}}$ is closer to zero and often slightly negative, suggesting that refinement introduces only limited adjustment when the first pass is already correct. The lower panel shows that such changes are accompanied by larger distributional shifts for misclassified positions than for correct positions, particularly in later bins. This suggests that the refinement block applies stronger reshaping when the provisional distribution is less reliable, while leaving some already stable predictions nearly unchanged.

Overall, the refinement analysis shows that the feedback block acts as a distribution reshaping and calibration mechanism before final decoding. Positive $\Delta p_{\mathrm{true}}$ indicates direct improvement in the probability assigned to the ground truth activity, whereas negative or near zero values can reflect recalibration among plausible competing activities. Because GGATN performs full sequence generation with structured decoding, the refined distribution is not used as an isolated position wise decision. Instead, refinement reshapes local activity evidence before the final sequence level decoding step, allowing the model to balance activity probabilities under broader contextual and graph based constraints.

\subsection{Structured Decoding and Transition Correction}
\label{subsec:decoding}

Figure~\ref{fig:inter_5plots}(c) examines the effect of structured decoding on GGATN activity generation. During generation, GGATN first performs a single forward pass and produces activity logits for all positions. These logits define local activity preferences, but the final sequence is not obtained by independent argmax decoding. Instead, GGATN combines unary scores with learned transition bias, optional bigram prior information, and graph based admissibility constraints when the adjacency matrix is used. A Viterbi style dynamic program then selects the highest scoring admissible path. This makes generation explicitly process aware: local scores are filtered through the transition structure of the global process graph. The figure analyzes this effect by comparing the raw unary top candidate at each position with the final activity selected in the decoded sequence.

For this analysis, each generated real event position is compared before and after structured decoding. The raw candidate is defined as the highest scoring activity under the unary activity logits at that position, after excluding special tokens. Its transition status is evaluated with respect to the previously decoded activity in the final path. A raw candidate is classified as \emph{admissible and EOS reachable} if the transition is allowed by the process graph and the candidate can still reach \texttt{EOS} through at least one directed path. It is classified as \emph{admissible but EOS unreachable} if the immediate transition is allowed but the candidate cannot reach \texttt{EOS}, and as \emph{inadmissible} if the immediate transition is not allowed by the process graph. The final outcome records how structured decoding handles the raw candidate. A position is marked as \emph{retained} when the final decoded activity is the same as the raw candidate. It is marked as \emph{corrected to admissible} when an inadmissible or EOS unreachable raw candidate is replaced by an admissible activity. It is marked as \emph{reassigned among admissible} when the raw candidate is already admissible and EOS reachable, but structured decoding selects a different admissible activity to improve the full path. The Sankey diagram aggregates these transitions across relative sequence position bins, showing how local unary preferences are transformed into process admissible decoded activities.

We use BPI17 for this analysis because it has the largest activity space in the benchmark, with $66$ activities, and therefore presents the most complex transition structure. Figure~\ref{fig:inter_5plots}(c) shows that a substantial number of raw unary top candidates are not directly retained. Among all generated positions, $86{,}951$ raw candidates are admissible and EOS reachable, while $34{,}879$ are inadmissible. Structured decoding corrects all inadmissible candidates to admissible transitions and additionally reassigns $22{,}844$ candidates among admissible alternatives. The remaining $64{,}107$ candidates are retained. This shows that structured decoding is not a cosmetic post processing step: it actively transforms local activity preferences into process admissible paths by correcting inadmissible local choices and selecting among admissible alternatives for path consistency.

\section{Conclusion}
\label{sec:conclusion}

This paper proposed the Graph Grounded Cross Attention Transformer Neural Network (GGATN) for structurally constrained full event sequence generation in PPM. GGATN combines global process graph learning, Transformer based sequence contextualization, graph grounded cross attention, feedback refinement, and Viterbi style graph constrained decoding to generate complete event sequences with activities, timestamps, sequence length, and event and sequence attributes. Unlike autoregressive language model baselines, GGATN performs single pass generation over all positions and combines sequence representations with activity embeddings learned from the global process graph. 

Experiments on six benchmark datasets show that GGATN overall provides more reliable generation quality than practical LLM baselines. Across most datasets, GGATN variants provide stronger overall generation quality, with competitive or leading results in sequence similarity, DL similarity, bigram JSD based control flow distribution, and duration distribution. They also maintain higher activity accuracy, stronger attribute reconstruction, zero hallucinated activities, and zero sequence level attribute inconsistency. The LLM baselines improve under larger context settings in some cases, but remain limited by coverage failures, hallucinated outputs, unstable behavior on complex datasets, and high computational cost.

The ablation analyses further confirm that the global GAT encoder remains stable across frozen, jointly trainable, and staged training regimes, indicating that the process graph provides a robust structural basis for generation. In addition, the interpretability analyses provide a stage wise view of model behavior, showing how sequence context, graph structure, activity refinement, and transition constrained decoding jointly shape the generated sequences.

Future work can extend GGATN in four directions. First, the global process graph can be enriched with more informative edge representations, such as transition distributions, temporal gap statistics, attribute changes, and context dependent transition patterns. Second, because the LLM baselines benefit from bucket based prompting while GGATN uses a single global graph setting, future work can investigate bucket specific or scale adaptive graph construction to better reflect sequence length, sparsity, and transition density. Third, although the present experiments use mostly global hyperparameter settings, dataset specific tuning of the graph encoder, cross attention module, refinement module, transition bias, and decoding parameters may further improve performance. Fourth, the tradeoff among activity, temporal, event attribute, and sequence attribute heads should be studied more systematically, since stronger control flow generation may not always coincide with optimal timestamp or attribute reconstruction.

\section{Declaration of generative AI and AI-assisted technologies in the writing process}
Statement: During the preparation of this work the authors used ChatGPT in order to improve the readability and language of the manuscript in the writing process. After using this tool/service, the authors reviewed and edited the content as needed and take full responsibility for the content of the published article.

\bibliographystyle{cas-model2-names}
\bibliography{ref}

@misc{Steeman2013BPI,
  author    = {Steeman, Ward},
  title     = {{BPI Challenge 2013: Incidents (Event Log)}},
  year      = {2013},
  publisher = {Ghent University},
  note      = {Dataset, Version 1},
  doi       = {10.4121/uuid:500573e6-accc-4b0c-9576-aa5468b10cee},
  url       = {https://doi.org/10.4121/uuid:500573e6-accc-4b0c-9576-aa5468b10cee}
}

@misc{Steeman2013BPIclosed,
  author    = {Steeman, Ward},
  title     = {{BPI Challenge 2013: Closed Problems (Event Log)}},
  year      = {2013},
  publisher = {4TU.ResearchData},
  note      = {Dataset, Version 1},
  doi       = {10.4121/uuid:c2c3b154-ab26-4b31-a0e8-8f2350ddac11},
  url       = {https://doi.org/10.4121/uuid:c2c3b154-ab26-4b31-a0e8-8f2350ddac11}
}

@inproceedings{wang2025auto,
  title={Auto-ML Graph Neural Network Hypermodels for Outcome Prediction in Event-Sequence Data},
  author={Wang, Fang and Kosca, Lance and Kosca, Adrienne and Gacesa, Marko and Damiani, Ernesto},
  booktitle={2025 IEEE 19th International Conference on Application of Information and Communication Technologies (AICT)},
  pages={1--6},
  year={2025},
  organization={IEEE}
}

@inproceedings{wang2025leveraging,
  title={Leveraging Duration Pseudo-embeddings in Multilevel LSTM and GCN Hypermodels for Outcome-Oriented PPM},
  author={Wang, Fang and Ceravolo, Paolo and Damiani, Ernesto},
  booktitle={International Conference on Business Process Management},
  pages={94--106},
  year={2025},
  organization={Springer}
}

@misc{Polato2017HelpDesk,
  author    = {Polato, Mirko},
  title     = {{Helpdesk: Ticketing Management Process (Event Log)}},
  year      = {2017},
  publisher = {4TU.ResearchData},
  note      = {Dataset, Version 1},
  doi       = {10.4121/uuid:0c60edf1-6f83-4e75-9367-4c63b3e9d5bb},
  url       = {https://doi.org/10.4121/uuid:0c60edf1-6f83-4e75-9367-4c63b3e9d5bb}
}

@misc{vanDongen2020BPI,
  author    = {Van Dongen, Boudewijn F.},
  title     = {{BPI Challenge 2020: Prepaid Travel Cost (Event Log)}},
  year      = {2020},
  publisher = {4TU.ResearchData},
  note      = {Dataset, Version 1},
  doi       = {10.4121/uuid:52fb97d4-4588-43c9-9d04-3604d4613b51},
  url       = {https://doi.org/10.4121/uuid:52fb97d4-4588-43c9-9d04-3604d4613b51}
}

@misc{Sepsis,
  doi = {10.4121/UUID:915D2BFB-7E84-49AD-A286-DC35F063A460},
  url = {https://data.4tu.nl/articles/_/12707639/1},
  author = {Mannhardt, Felix},
  language = {en},
  title = {Sepsis Cases - Event Log},
  publisher = {Eindhoven University of Technology},
  year = {2016},
  copyright = {4TU General Terms of Use}
}

@misc{BPI17,
  doi = {10.4121/UUID:5F3067DF-F10B-45DA-B98B-86AE4C7A310B},
  url = {https://data.4tu.nl/articles/_/12696884/1},
  author = {van Dongen, Boudewijn},
  language = {en},
  title = {BPI Challenge 2017},
  publisher = {Eindhoven University of Technology},
  year = {2017},
  copyright = {4TU General Terms of Use}
}

@article{rama2021deep,
  title={Deep learning for predictive business process monitoring: Review and benchmark},
  author={Rama-Maneiro, Efr{\'e}n and Vidal, Juan C and Lama, Manuel},
  journal={IEEE Transactions on Services Computing},
  volume={16},
  number={1},
  pages={739--756},
  year={2021},
  publisher={IEEE}
}

@article{rama2023embedding,
  title={Embedding graph convolutional networks in recurrent neural networks for predictive monitoring},
  author={Rama-Maneiro, Efr{\'e}n and Vidal, Juan C and Lama, Manuel},
  journal={IEEE Transactions on Knowledge and Data Engineering},
  volume={36},
  number={1},
  pages={137--151},
  year={2023},
  publisher={IEEE}
}

@inproceedings{wuyts2024sutran,
  title={Sutran: an encoder-decoder transformer for full-context-aware suffix prediction of business processes},
  author={Wuyts, Brecht and Vanden Broucke, Seppe and De Weerdt, Jochen},
  booktitle={2024 6th International Conference on Process Mining (ICPM)},
  pages={17--24},
  year={2024},
  organization={IEEE}
}

@article{bukhsh2021processtransformer,
  title={Processtransformer: Predictive business process monitoring with transformer network},
  author={Bukhsh, Zaharah A and Saeed, Aaqib and Dijkman, Remco M},
  journal={arXiv preprint arXiv:2104.00721},
  year={2021}
}

@article{marquez2017predictive,
  title={Predictive monitoring of business processes: a survey},
  author={M{\'a}rquez-Chamorro, Alfonso Eduardo and Resinas, Manuel and Ruiz-Cort{\'e}s, Antonio},
  journal={IEEE Transactions on Services Computing},
  volume={11},
  number={6},
  pages={962--977},
  year={2017},
  publisher={IEEE}
}

@article{verenich2019survey,
  title={Survey and cross-benchmark comparison of remaining time prediction methods in business process monitoring},
  author={Verenich, Ilya and Dumas, Marlon and Rosa, Marcello La and Maggi, Fabrizio Maria and Teinemaa, Irene},
  journal={ACM Transactions on Intelligent Systems and Technology (TIST)},
  volume={10},
  number={4},
  pages={1--34},
  year={2019},
  publisher={ACM New York, NY, USA}
}

@article{teinemaa2019outcome,
  title={Outcome-oriented predictive process monitoring: Review and benchmark},
  author={Teinemaa, Irene and Dumas, Marlon and Rosa, Marcello La and Maggi, Fabrizio Maria},
  journal={ACM Transactions on Knowledge Discovery from Data (TKDD)},
  volume={13},
  number={2},
  pages={1--57},
  year={2019},
  publisher={ACM New York, NY, USA}
}

@inproceedings{taymouri2020predictive,
  title={Predictive business process monitoring via generative adversarial nets: the case of 
 event prediction},
  author={Taymouri, Farbod and Rosa, Marcello La and Erfani, Sarah and Bozorgi, Zahra Dasht and Verenich, Ilya},
  booktitle={International Conference on Business Process Management},
  pages={237--256},
  year={2020},
  organization={Springer}
}

@inproceedings{weinzierl2020prescriptive,
  title={Prescriptive business process monitoring for recommending next best actions},
  author={Weinzierl, Sven and Dunzer, Sebastian and Zilker, Sandra and Matzner, Martin},
  booktitle={International conference on business process management},
  pages={193--209},
  year={2020},
  organization={Springer}
}

@inproceedings{tax2017predictive,
  title={Predictive business process monitoring with LSTM neural networks},
  author={Tax, Niek and Verenich, Ilya and La Rosa, Marcello and Dumas, Marlon},
  booktitle={International conference on advanced information systems engineering},
  pages={477--492},
  year={2017},
  organization={Springer}
}

@inproceedings{weinzierl2021exploring,
  title={Exploring gated graph sequence neural networks for predicting next process activities},
  author={Weinzierl, Sven},
  booktitle={International conference on business process management},
  pages={30--42},
  year={2021},
  organization={Springer}
}

@article{pasquadibisceglie2024sf,
  title={PROPHET: Explainable Predictive Process Monitoring With Heterogeneous Graph Neural Networks},
  author={Pasquadibisceglie, Vincenzo and Scaringi, Raffaele and Appice, Annalisa and Castellano, Giovanna and Malerba, Donato},
  journal={IEEE Transactions on Services Computing},
  volume={17},
  number={6},
  pages={4111--4124},
  year={2024},
  publisher={IEEE}
}

@inproceedings{navarin2017lstm,
  title={LSTM networks for data-aware remaining time prediction of business process instances},
  author={Navarin, Nicolo and Vincenzi, Beatrice and Polato, Mirko and Sperduti, Alessandro},
  booktitle={2017 IEEE symposium series on computational intelligence (SSCI)},
  pages={1--7},
  year={2017},
  organization={IEEE}
}

@article{wang2025hgcn,
  title={Hgcn (o): A self-tuning gcn hypermodel toolkit for outcome prediction in event-sequence data},
  author={Wang, Fang and Ceravolo, Paolo and Damiani, Ernesto},
  journal={arXiv preprint arXiv:2507.22524},
  year={2025}
}

@article{guo2024explainable,
  title={Explainable and effective process remaining time prediction using feature-informed cascade prediction model},
  author={Guo, Na and Liu, Cong and Li, Caihong and Zeng, Qingtian and Ouyang, Chun and Liu, Qingzhi and Lu, Xixi},
  journal={IEEE Transactions on Services Computing},
  volume={17},
  number={3},
  pages={949--962},
  year={2024},
  publisher={IEEE}
}

@article{wang2025comprehensive,
  title={Comprehensive attribute encoding and dynamic lstm hypermodels for outcome oriented predictive business process monitoring},
  author={Wang, Fang and Ceravolo, Paolo and Damiani, Ernesto},
  journal={arXiv preprint arXiv:2506.03696},
  year={2025}
}

@inproceedings{rivera2022multi,
  title={Multi-attribute transformers for sequence prediction in business process management},
  author={Rivera Lazo, Gonzalo and {\~N}anculef, Ricardo},
  booktitle={International Conference on Discovery Science},
  pages={184--194},
  year={2022},
  organization={Springer}
}

@article{pasquadibisceglie2021multi,
  title={A multi-view deep learning approach for predictive business process monitoring},
  author={Pasquadibisceglie, Vincenzo and Appice, Annalisa and Castellano, Giovanna and Malerba, Donato},
  journal={IEEE Transactions on Services Computing},
  volume={15},
  number={4},
  pages={2382--2395},
  year={2021},
  publisher={IEEE}
}

@inproceedings{pasquadibisceglie2019using,
  title={Using convolutional neural networks for predictive process analytics},
  author={Pasquadibisceglie, Vincenzo and Appice, Annalisa and Castellano, Giovanna and Malerba, Donato},
  booktitle={2019 international conference on process mining (ICPM)},
  pages={129--136},
  year={2019},
  organization={IEEE}
}

@article{elman1990finding,
  title={Finding structure in time},
  author={Elman, Jeffrey L},
  journal={Cognitive science},
  volume={14},
  number={2},
  pages={179--211},
  year={1990},
  publisher={Wiley Online Library}
}

@article{medsker2001recurrent,
  title={Recurrent neural networks},
  author={Medsker, Larry R and Jain, Lakhmi and others},
  journal={Design and applications},
  volume={5},
  number={64-67},
  pages={2},
  year={2001}
}

@article{bahdanau2014neural,
  title={Neural machine translation by jointly learning to align and translate},
  author={Bahdanau, Dzmitry and Cho, Kyunghyun and Bengio, Yoshua},
  journal={arXiv preprint arXiv:1409.0473},
  year={2014}
}

@inproceedings{cho2014learning,
  title={Learning phrase representations using RNN encoder--decoder for statistical machine translation},
  author={Cho, Kyunghyun and Van Merri{\"e}nboer, Bart and Gul{\c{c}}ehre, {\c{C}}a{\u{g}}lar and Bahdanau, Dzmitry and Bougares, Fethi and Schwenk, Holger and Bengio, Yoshua},
  booktitle={Proceedings of the 2014 conference on empirical methods in natural language processing (EMNLP)},
  pages={1724--1734},
  year={2014}
}

@inproceedings{radford2021learning,
  title={Learning transferable visual models from natural language supervision},
  author={Radford, Alec and Kim, Jong Wook and Hallacy, Chris and Ramesh, Aditya and Goh, Gabriel and Agarwal, Sandhini and Sastry, Girish and Askell, Amanda and Mishkin, Pamela and Clark, Jack and others},
  booktitle={International conference on machine learning},
  pages={8748--8763},
  year={2021},
  organization={PmLR}
}

@article{brown2020language,
  title={Language models are few-shot learners},
  author={Brown, Tom and Mann, Benjamin and Ryder, Nick and Subbiah, Melanie and Kaplan, Jared D and Dhariwal, Prafulla and Neelakantan, Arvind and Shyam, Pranav and Sastry, Girish and Askell, Amanda and others},
  journal={Advances in neural information processing systems},
  volume={33},
  pages={1877--1901},
  year={2020}
}

@article{raffel2020exploring,
  title={Exploring the limits of transfer learning with a unified text-to-text transformer},
  author={Raffel, Colin and Shazeer, Noam and Roberts, Adam and Lee, Katherine and Narang, Sharan and Matena, Michael and Zhou, Yanqi and Li, Wei and Liu, Peter J},
  journal={Journal of machine learning research},
  volume={21},
  number={140},
  pages={1--67},
  year={2020}
}

@article{touvron2023llama,
  title={Llama: Open and efficient foundation language models},
  author={Touvron, Hugo and Lavril, Thibaut and Izacard, Gautier and Martinet, Xavier and Lachaux, Marie-Anne and Lacroix, Timoth{\'e}e and Rozi{\`e}re, Baptiste and Goyal, Naman and Hambro, Eric and Azhar, Faisal and others},
  journal={arXiv preprint arXiv:2302.13971},
  year={2023}
}

@inproceedings{fan2019using,
  title={Using local knowledge graph construction to scale seq2seq models to multi-document inputs},
  author={Fan, Angela and Gardent, Claire and Braud, Chlo{\'e} and Bordes, Antoine},
  booktitle={Proceedings of the 2019 Conference on Empirical Methods in Natural Language Processing and the 9th International Joint Conference on Natural Language Processing (EMNLP-IJCNLP)},
  pages={4186--4196},
  year={2019}
}

@inproceedings{cao2025graphinsight,
  title={Graphinsight: Unlocking insights in large language models for graph structure understanding},
  author={Cao, Yukun and Han, Shuo and Gao, Zengyi and Ding, Zezhong and Xie, Xike and Zhou, S Kevin},
  booktitle={Proceedings of the 63rd Annual Meeting of the Association for Computational Linguistics (Volume 1: Long Papers)},
  pages={12096--12134},
  year={2025}
}

@inproceedings{kasner2024beyond,
  title={Beyond traditional benchmarks: Analyzing behaviors of open LLMs on data-to-text generation},
  author={Kasner, Zden{\v{e}}k and Du{\v{s}}ek, Ond{\v{r}}ej},
  booktitle={Proceedings of the 62nd Annual Meeting of the Association for Computational Linguistics (Volume 1: Long Papers)},
  pages={12045--12072},
  year={2024}
}

@article{geng2025generating,
  title={Generating structured outputs from language models: Benchmark and studies},
  author={Geng, Saibo and Cooper, Hudson and Moskal, Micha{\l} and Jenkins, Samuel and Berman, Julian and Ranchin, Nathan and West, Robert and Horvitz, Eric and Nori, Harsha},
  journal={arXiv e-prints},
  pages={arXiv--2501},
  year={2025}
}

@inproceedings{wang2024tram,
  title={Tram: Benchmarking temporal reasoning for large language models},
  author={Wang, Yuqing and Zhao, Yun},
  booktitle={Findings of the Association for Computational Linguistics: ACL 2024},
  pages={6389--6415},
  year={2024}
}

@inproceedings{li2025exposing,
  title={Exposing numeracy gaps: A benchmark to evaluate fundamental numerical abilities in large language models},
  author={Li, Haoyang and Chen, Xuejia and Xu, Zhanchao and Li, Darian and Hu, Nicole and Teng, Fei and Li, Yiming and Qiu, Luyu and Zhang, Chen Jason and Qing, Li and others},
  booktitle={Findings of the Association for Computational Linguistics: ACL 2025},
  pages={20004--20026},
  year={2025}
}

@inproceedings{kandpal2023large,
  title={Large language models struggle to learn long-tail knowledge},
  author={Kandpal, Nikhil and Deng, Haikang and Roberts, Adam and Wallace, Eric and Raffel, Colin},
  booktitle={International conference on machine learning},
  pages={15696--15707},
  year={2023},
  organization={PMLR}
}

@article{kipf2016semi,
  title={Semi-supervised classification with graph convolutional networks},
  author={Kipf, Thomas N and Welling, Max},
  journal={arXiv preprint arXiv:1609.02907},
  year={2016}
}

@article{hamilton2017inductive,
  title={Inductive representation learning on large graphs},
  author={Hamilton, Will and Ying, Zhitao and Leskovec, Jure},
  journal={Advances in neural information processing systems},
  volume={30},
  year={2017}
}

@article{bronstein2021geometric,
  title={Geometric deep learning: Grids, groups, graphs, geodesics, and gauges},
  author={Bronstein, Michael M and Bruna, Joan and Cohen, Taco and Veli{\v{c}}kovi{\'c}, Petar},
  journal={arXiv preprint arXiv:2104.13478},
  year={2021}
}

@inproceedings{gilmer2017neural,
  title={Neural message passing for quantum chemistry},
  author={Gilmer, Justin and Schoenholz, Samuel S and Riley, Patrick F and Vinyals, Oriol and Dahl, George E},
  booktitle={International conference on machine learning},
  pages={1263--1272},
  year={2017},
  organization={Pmlr}
}

@inproceedings{schmidt2019generalization,
  title={Generalization in generation: A closer look at exposure bias},
  author={Schmidt, Florian},
  booktitle={Proceedings of the 3rd Workshop on Neural Generation and Translation},
  pages={157--167},
  year={2019}
}

@inproceedings{zhou2021structure,
  title={Structure-aware fine-tuning of sequence-to-sequence transformers for transition-based AMR parsing},
  author={Zhou, Jiawei and Naseem, Tahira and Astudillo, Ram{\'o}n Fernandez and Lee, Young-Suk and Florian, Radu and Roukos, Salim},
  booktitle={Proceedings of the 2021 conference on empirical methods in natural language processing},
  pages={6279--6290},
  year={2021}
}

@inproceedings{zhu2019modeling,
  title={Modeling graph structure in transformer for better AMR-to-text generation},
  author={Zhu, Jie and Li, Junhui and Zhu, Muhua and Qian, Longhua and Zhang, Min and Zhou, Guodong},
  booktitle={Proceedings of the 2019 conference on empirical methods in natural language processing and the 9th international joint conference on natural language processing (EMNLP-IJCNLP)},
  pages={5459--5468},
  year={2019}
}

@article{vaswani2017attention,
  title={Attention is all you need},
  author={Vaswani, Ashish and Shazeer, Noam and Parmar, Niki and Uszkoreit, Jakob and Jones, Llion and Gomez, Aidan N and Kaiser, {\L}ukasz and Polosukhin, Illia},
  journal={Advances in neural information processing systems},
  volume={30},
  year={2017}
}

@inproceedings{zhou2021informer,
  title={Informer: Beyond efficient transformer for long sequence time-series forecasting},
  author={Zhou, Haoyi and Zhang, Shanghang and Peng, Jieqi and Zhang, Shuai and Li, Jianxin and Xiong, Hui and Zhang, Wancai},
  booktitle={Proceedings of the AAAI conference on artificial intelligence},
  volume={35},
  number={12},
  pages={11106--11115},
  year={2021}
}

@inproceedings{nguyen2020time,
  title={Time matters: Time-aware lstms for predictive business process monitoring},
  author={Nguyen, An and Chatterjee, Srijeet and Weinzierl, Sven and Schwinn, Leo and Matzner, Martin and Eskofier, Bjoern},
  booktitle={International Conference on Process Mining},
  pages={112--123},
  year={2020},
  organization={Springer}
}

@article{chen2022multi,
  title={Multi-task prediction method of business process based on BERT and transfer learning},
  author={Chen, Hang and Fang, Xianwen and Fang, Huan},
  journal={Knowledge-Based Systems},
  volume={254},
  pages={109603},
  year={2022},
  publisher={Elsevier}
}

@article{scarselli2008graph,
  title={The graph neural network model},
  author={Scarselli, Franco and Gori, Marco and Tsoi, Ah Chung and Hagenbuchner, Markus and Monfardini, Gabriele},
  journal={IEEE transactions on neural networks},
  volume={20},
  number={1},
  pages={61--80},
  year={2008},
  publisher={IEEE}
}

@article{velivckovic2017graph,
  title={Graph attention networks},
  author={Veli{\v{c}}kovi{\'c}, Petar and Cucurull, Guillem and Casanova, Arantxa and Romero, Adriana and Lio, Pietro and Bengio, Yoshua},
  journal={arXiv preprint arXiv:1710.10903},
  year={2017}
}

@article{li2015gated,
  title={Gated graph sequence neural networks},
  author={Li, Yujia and Tarlow, Daniel and Brockschmidt, Marc and Zemel, Richard},
  journal={arXiv preprint arXiv:1511.05493},
  year={2015}
}

@inproceedings{koncel2019text,
  title={Text generation from knowledge graphs with graph transformers},
  author={Koncel-Kedziorski, Rik and Bekal, Dhanush and Luan, Yi and Lapata, Mirella and Hajishirzi, Hannaneh},
  booktitle={Proceedings of the 2019 Conference of the North American Chapter of the Association for Computational Linguistics: Human Language Technologies, Volume 1 (Long and Short Papers)},
  pages={2284--2293},
  year={2019}
}

@article{alon2020bottleneck,
  title={On the bottleneck of graph neural networks and its practical implications},
  author={Alon, Uri and Yahav, Eran},
  journal={arXiv preprint arXiv:2006.05205},
  year={2020}
}

@article{dwivedi2020generalization,
  title={A generalization of transformer networks to graphs},
  author={Dwivedi, Vijay Prakash and Bresson, Xavier},
  journal={arXiv preprint arXiv:2012.09699},
  year={2020}
}

@article{ying2021transformers,
  title={Do transformers really perform badly for graph representation?},
  author={Ying, Chengxuan and Cai, Tianle and Luo, Shengjie and Zheng, Shuxin and Ke, Guolin and He, Di and Shen, Yanming and Liu, Tie-Yan},
  journal={Advances in neural information processing systems},
  volume={34},
  pages={28877--28888},
  year={2021}
}

@article{rampavsek2022recipe,
  title={Recipe for a general, powerful, scalable graph transformer},
  author={Ramp{\'a}{\v{s}}ek, Ladislav and Galkin, Michael and Dwivedi, Vijay Prakash and Luu, Anh Tuan and Wolf, Guy and Beaini, Dominique},
  journal={Advances in Neural Information Processing Systems},
  volume={35},
  pages={14501--14515},
  year={2022}
}

@inproceedings{beck2018graph,
  title={Graph-to-sequence learning using gated graph neural networks},
  author={Beck, Daniel and Haffari, Gholamreza and Cohn, Trevor},
  booktitle={Proceedings of the 56th Annual Meeting of the Association for Computational Linguistics (Volume 1: Long Papers)},
  pages={273--283},
  year={2018}
}

@inproceedings{lewis2020bart,
  title={BART: Denoising sequence-to-sequence pre-training for natural language generation, translation, and comprehension},
  author={Lewis, Mike and Liu, Yinhan and Goyal, Naman and Ghazvininejad, Marjan and Mohamed, Abdelrahman and Levy, Omer and Stoyanov, Veselin and Zettlemoyer, Luke},
  booktitle={Proceedings of the 58th annual meeting of the association for computational linguistics},
  pages={7871--7880},
  year={2020}
}

@article{hochreiter1997lstm,
  title={Long short-term memory},
  author={Hochreiter, Sepp and Schmidhuber, J{\"u}rgen},
  journal={Neural computation},
  volume={9},
  number={8},
  pages={1735--1780},
  year={1997},
  publisher={MIT press}
}

@article{xu2018graph2seq,
  title={Graph2seq: Graph to sequence learning with attention-based neural networks},
  author={Xu, Kun and Wu, Lingfei and Wang, Zhiguo and Feng, Yansong and Witbrock, Michael and Sheinin, Vadim},
  journal={arXiv preprint arXiv:1804.00823},
  year={2018}
}

@article{wang2025time,
  title={Time-aware and transition-semantic graph neural networks for interpretable predictive business process monitoring},
  author={Wang, Fang and Damiani, Ernesto},
  journal={Expert Systems with Applications},
  pages={130320},
  year={2025},
  publisher={Elsevier}
}

@article{yin2025stgnn,
  title={STGNN: A novel spatial-temporal graph neural network for predicting complicated business process performance under multi-event parallelism},
  author={Yin, Jun and Qiu, Aoxue and Fang, Lin and Wang, Nianxin and Dong, Chen and Ge, Shilun},
  journal={Expert Systems with Applications},
  volume={291},
  pages={128391},
  year={2025},
  publisher={Elsevier}
}

@inproceedings{wang2019outcome,
  title={Outcome-oriented predictive process monitoring with attention-based bidirectional LSTM neural networks},
  author={Wang, Jiaojiao and Yu, Dongjin and Liu, Chengfei and Sun, Xiaoxiao},
  booktitle={2019 IEEE international conference on web services (ICWS)},
  pages={360--367},
  year={2019},
  organization={IEEE}
}

@inproceedings{dissegna2024graph,
  title={Graph neural networks for PPM: review and benchmark for next activity predictions},
  author={Dissegna, Sebastiano and Di Francescomarino, Chiara},
  booktitle={International Conference on Business Process Management},
  pages={31--43},
  year={2024},
  organization={Springer}
}

@inproceedings{lin2019mm,
  title={Mm-pred: A deep predictive model for multi-attribute event sequence},
  author={Lin, Li and Wen, Lijie and Wang, Jianmin},
  booktitle={Proceedings of the 2019 SIAM international conference on data mining},
  pages={118--126},
  year={2019},
  organization={SIAM}
}

@inproceedings{evermann2016deep,
  title={A deep learning approach for predicting process behaviour at runtime},
  author={Evermann, Joerg and Rehse, Jana-Rebecca and Fettke, Peter},
  booktitle={International Conference on Business Process Management},
  pages={327--338},
  year={2016},
  organization={Springer}
}

@inproceedings{khan2021deepprocess,
  title={DeepProcess: supporting business process execution using a MANN-based recommender system},
  author={Khan, Asjad and Le, Hung and Do, Kien and Tran, Truyen and Ghose, Aditya and Dam, Hoa and Sindhgatta, Renuka},
  booktitle={International Conference on Service-Oriented Computing},
  pages={19--33},
  year={2021},
  organization={Springer}
}

@inproceedings{duong2023remaining,
  title={Remaining cycle time prediction with graph neural networks for predictive process monitoring},
  author={Duong, Le Toan and Trav{\'e}-Massuy{\`e}s, Louise and Subias, Audine and Merle, Christophe},
  booktitle={Proceedings of the 2023 8th international conference on machine learning technologies},
  pages={95--101},
  year={2023}
}

@article{kratsch2021machine,
  title={Machine Learning in Business Process Monitoring: A Comparison of Deep Learning and Classical Approaches Used for Outcome Prediction: W. Kratsch et al.},
  author={Kratsch, Wolfgang and Manderscheid, Jonas and R{\"o}glinger, Maximilian and Seyfried, Johannes},
  journal={Business \& Information Systems Engineering},
  volume={63},
  number={3},
  pages={261--276},
  year={2021},
  publisher={Springer}
}

@article{sutskever2014sequence,
  title={Sequence to sequence learning with neural networks},
  author={Sutskever, Ilya and Vinyals, Oriol and Le, Quoc V},
  journal={Advances in neural information processing systems},
  volume={27},
  year={2014}
}

@article{chung2014empirical,
  title={Empirical evaluation of gated recurrent neural networks on sequence modeling},
  author={Chung, Junyoung and Gulcehre, Caglar and Cho, KyungHyun and Bengio, Yoshua},
  journal={arXiv preprint arXiv:1412.3555},
  year={2014}
}

@article{harane2020comprehensive,
  title={Comprehensive survey on deep learning approaches in predictive business process monitoring},
  author={Harane, Nitin and Rathi, Sheetal},
  journal={Modern approaches in machine learning and cognitive science: A walkthrough: Latest trends in ai},
  pages={115--128},
  year={2020},
  publisher={Springer}
}

@inproceedings{camargo2019learning,
  title={Learning accurate LSTM models of business processes},
  author={Camargo, Manuel and Dumas, Marlon and Gonz{\'a}lez-Rojas, Oscar},
  booktitle={International Conference on Business Process Management},
  pages={286--302},
  year={2019},
  organization={Springer}
}

\appendix
\renewcommand{\thesection}{Appendix~\Alph{section}}

\section{Complete Results by Dataset}
\label{app:full_results}
The appendix tables follow the abbreviations defined in Section~\ref{sec:Res}. Briefly, \textbf{M} denotes model; \textbf{G} is the main GGATN model with a frozen graph attention encoder, \textbf{G\_j} is the fully joint training variant, and \textbf{G\_s5}/\textbf{G\_s10} are staged unfreezing variants. \textbf{L(4k)}, \textbf{L(32k)}, and \textbf{M(4k)} denote the Llama and Mistral baselines. The main generation metrics are coverage, SS, DL, bigram JSD, and duration WD. Dataset specific columns report activity, temporal, event level, and sequence level attribute metrics. For sequence level attributes, we also include \textbf{I-c} to report the number of generated sequences with inconsistent values for a given attribute, and \textbf{I-pct} for  the corresponding inconsistency percentage over generated sequences. Higher values are better for coverage, SS, DL, recall, and accuracy metrics, while lower values are better for JSD, WD, I-pct, I-c, MAE based metrics, and hallucinated activities. 

\begin{table}[pos = htbp]
\caption{Complete Results for Sepsis}
\label{appe:tab_1}
\centering
\setlength{\tabcolsep}{1.5pt}
\begin{tabularx}{\textwidth}{@{} l *{15}{C} @{}}
\toprule
\multicolumn{16}{c}{\textbf{Sepsis}} \\ \midrule
\textbf{M} & \textbf{c(S)} & \textbf{JSD} & \textbf{SS} & \textbf{WD} & \textbf{DL} & \textbf{H} & \textbf{re(A)} & \textbf{Ac(A)} & \textbf{MA(T)} & \textbf{\#S} & \textbf{Ac(E)} & \multicolumn{3}{c}{\textbf{MA(E)}} & \textbf{MA(S)} \\
\textbf{} & \textbf{} & \textbf{} & \textbf{} & \textbf{} & \textbf{} & \textbf{} & \textbf{} & \textbf{} & \textbf{} & \textbf{} & \textbf{O:G} & \textbf{LEU} & \textbf{CRP} & \textbf{LA} & \textbf{Age} \\ \midrule
\textbf{G} & 1.0000 & 0.2977 & 0.7169 & 1089448 & 0.6271 & 0 & 0.8125 & 0.4816 & 446981 & 104 & 0.7311 & 5.97 & 53.48 & 0.6040 & 12.77 \\
\textbf{G\_j} & 1.0000 & 0.3046 & 0.7184 & 1073339 & 0.6314 & 0 & 0.8125 & 0.4884 & 345565 & 104 & 0.7273 & 6.02 & 53.50 & 0.6175 & 13.08 \\
\textbf{G\_s5} & 1.0000 & 0.2921 & 0.7321 & 1361757 & 0.6401 & 0 & 0.8125 & 0.4878 & 334917 & 104 & 0.7367 & 6.12 & 53.57 & 0.6185 & 12.97 \\
\textbf{G\_s10} & 1.0000 & 0.3004 & 0.7272 & 894149 & 0.6490 & 0 & 0.8125 & 0.4991 & 621142 & 104 & 0.7473 & 6.03 & 52.63 & 0.6000 & 12.97 \\
\textbf{L(4k)} & 0.6346 & 0.9833 & 0.1061 & 1353228 & 0.0867 & 320 & 0.6250 & 0.0269 & 22346748 & 66 & 0.0350 & 19.28 & 197.08 & 4.1793 & 48.41 \\
\textbf{M(4k)} & 0.1635 & 1.5442 & 0.0201 & 2984395 & 0.0155 & 623 & 0.6250 & 0.0119 & 22264333 & 17 & 0.0113 & 141.05 & 914.12 & 5.9523 & 44.60 \\
\textbf{L(32k)} & 0.8654 & 0.5240 & 0.5181 & 1670784 & 0.4467 & 24 & 0.9375 & 0.2489 & 14337061 & 90 & 0.3258 & 52.85 & 187.92 & 3.6085 & 31.15\\
\midrule
\end{tabularx}

\begin{tabularx}{\textwidth}{@{} l *{15}{C} @{}}
\textbf{M} & \multicolumn{15}{c}{\textbf{Ac(S)}} \\
\textbf{} & \textbf{IS} & \textbf{DB} & \textbf{DO} & \textbf{SCT} & \textbf{HYP} & \textbf{SCRH} & \textbf{INF} & \textbf{DAA} & \textbf{DIC} & \textbf{DS} & \textbf{DiL} & \textbf{DOth} & \textbf{SC2+} & \textbf{DX} & \textbf{SCTmp} \\\midrule
\textbf{G} & 0.8558 & 0.8269 & 0.9038 & 0.6250 & 0.8942 & 0.8558 & 0.8654 & 0.6346 & 0.8365 & 0.9904 & 1.0000 & 0.9808 & 0.8462 & 0.8365 & 0.8365 \\
\textbf{G\_j} & 0.8558 & 0.8269 & 0.9038 & 0.6442 & 0.8942 & 0.8558 & 0.8750 & 0.6538 & 0.8365 & 0.9904 & 1.0000 & 0.9808 & 0.8462 & 0.8269 & 0.8269 \\
\textbf{G\_s5} & 0.8750 & 0.8462 & 0.9038 & 0.6250 & 0.8942 & 0.8750 & 0.8654 & 0.6346 & 0.8558 & 0.9904 & 1.0000 & 0.9808 & 0.8654 & 0.8558 & 0.8558 \\
\textbf{G\_s10} & 0.8654 & 0.8269 & 0.9038 & 0.6154 & 0.8942 & 0.8558 & 0.8654 & 0.6635 & 0.8365 & 0.9904 & 1.0000 & 0.9808 & 0.8654 & 0.8462 & 0.8462 \\
\textbf{L(4k)} & 0.1538 & 0.1346 & 0.1827 & 0.1250 & 0.1731 & 0.0481 & 0.0577 & 0.1250 & 0.1538 & 0.1923 & 0.1923 & 0.1827 & 0.1731 & 0.1442 & 0.1538 \\
\textbf{M(4k)} & 0.0673 & 0.0865 & 0.0962 & 0.0577 & 0.1058 & 0.0769 & 0.0673 & 0.0481 & 0.0769 & 0.0288 & 0.0962 & 0.1058 & 0.0481 & 0.0673 & 0.0865 \\
\textbf{L(32k)} & 0.6154 & 0.6154 & 0.7596 & 0.4808 & 0.7500 & 0.6346 & 0.5577 & 0.4615 & 0.6346 & 0.7981 & 0.7885 & 0.7885 & 0.6250 & 0.6346 & 0.5577 \\
\midrule
\end{tabularx}

\begin{tabularx}{\textwidth}{@{} l *{16}{C} @{}}
\textbf{M} & \multicolumn{8}{c}{\textbf{Ac(S)}} & \textbf{I-c} & \textbf{I-pct} & \textbf{I-c} & \textbf{I-pct} & \textbf{I-c} & \textbf{I-pct} & \textbf{I-c} & \textbf{I-pct} \\
\textbf{} & \textbf{DUC} & \textbf{SCL} & \textbf{OLI} & \textbf{DLA} & \textbf{DX} & \textbf{HYPX} & \textbf{DUS} & \textbf{DECG} & \multicolumn{2}{c}{\textbf{IS}} & \multicolumn{2}{c}{\textbf{DB}} & \multicolumn{2}{c}{\textbf{DO}} & \multicolumn{2}{c}{\textbf{SCT}} \\ \midrule
\textbf{G} & 0.5577 & 0.9423 & 0.9712 & 0.8365 & 0.2885 & 0.9808 & 0.6154 & 0.8269 & 0 & 0 & 0 & 0 & 0 & 0 & 0 & 0 \\
\textbf{G\_j} & 0.5385 & 0.9423 & 0.9712 & 0.8269 & 0.2404 & 0.9808 & 0.5577 & 0.8269 & 0 & 0 & 0 & 0 & 0 & 0 & 0 & 0 \\
\textbf{G\_s5} & 0.5192 & 0.9423 & 0.9712 & 0.8558 & 0.2596 & 0.9808 & 0.5192 & 0.8462 & 0 & 0 & 0 & 0 & 0 & 0 & 0 & 0 \\
\textbf{G\_s10} & 0.5865 & 0.9423 & 0.9712 & 0.8269 & 0.2596 & 0.9808 & 0.6250 & 0.8269 & 0 & 0 & 0 & 0 & 0 & 0 & 0 & 0  \\
\textbf{L(4k)} & 0.1538 & 0.1635 & 0.1923 & 0.1538 & 0.0192 & 0.1923 & 0.1442 & 0.1538 & 7 & 0.1061 & 5 & 0.0758 & 3 & 0.0455 & 5 & 0.0758  \\
\textbf{M(4k)} & 0.0481 & 0.1058 & 0.0962 & 0.0865 & 0.0096 & 0.1058 & 0.0385 & 0.0673 & 5 & 0.2941 & 4 & 0.2353 & 3 & 0.1765 & 3 & 0.1765 \\
\textbf{L(32k)} & 0.4231 & 0.7596 & 0.7885 & 0.6827 & 0.0962 & 0.7885 & 0.3750 & 0.5481 & 1 & 0.0111 & 1 & 0.0111 & 1 & 0.0111 & 1 & 0.0111 \\
\midrule
\end{tabularx}

\begin{threeparttable}
\begin{tabularx}{\textwidth}{@{} l *{20}{C} @{}}
\textbf{} & \textbf{I-c} & \textbf{I-pct} & \textbf{I-c} & \textbf{I-pct} & \textbf{I-c} & \textbf{I-pct} & \textbf{I-c} & \textbf{I-pct} & \textbf{I-c} & \textbf{I-pct} & \textbf{I-c} & \textbf{I-pct} & \textbf{I-c} & \textbf{I-pct} & \textbf{I-c} & \textbf{I-pct} & \textbf{I-c} & \textbf{I-pct} & \textbf{I-c} & \textbf{I-pct} \\ \midrule
\textbf{M} & \multicolumn{2}{c}{\textbf{HYP}} & \multicolumn{2}{c}{\textbf{SCRH}} & \multicolumn{2}{c}{\textbf{INF}} & \multicolumn{2}{c}{\textbf{DAA}} & \multicolumn{2}{c}{\textbf{DIC}} & \multicolumn{2}{c}{\textbf{DS}} & \multicolumn{2}{c}{\textbf{DiL}} & \multicolumn{2}{c}{\textbf{DOth}} & \multicolumn{2}{c}{\textbf{SC2+}} & \multicolumn{2}{c}{\textbf{DX}} \\ \midrule
\textbf{G} & 0 & 0 & 0 & 0 & 0 & 0 & 0 & 0 & 0 & 0 & 0 & 0 & 0 & 0 & 0 & 0 & 0 & 0 & 0 & 0 \\
\textbf{G\_j} & 0 & 0 & 0 & 0 & 0 & 0 & 0 & 0 & 0 & 0 & 0 & 0 & 0 & 0 & 0 & 0 & 0 & 0 & 0 & 0 \\
\textbf{G\_s5} & 0 & 0 & 0 & 0 & 0 & 0 & 0 & 0 & 0 & 0 & 0 & 0 & 0 & 0 & 0 & 0 & 0 & 0 & 0 & 0 \\
\textbf{G\_s10} & 0 & 0 & 0 & 0 & 0 & 0 & 0 & 0 & 0 & 0 & 0 & 0 & 0 & 0 & 0 & 0 & 0 & 0 & 0 & 0 \\
\textbf{L(4k)} & 3 & 0.0455 & 3 & 0.0455 & 4 & 0.0606 & 3 & 0.0455 & 3 & 0.0455 & 3 & 0.0455 & 3 & 0.0455 & 3 & 0.0455 & 5 & 0.0758 & 3 & 0.0455 \\
\textbf{M(4k)} & 2 & 0.1176 & 3 & 0.1765 & 5 & 0.2941 & 4 & 0.2353 & 3 & 0.1765 & 3 & 0.1765 & 4 & 0.2353 & 4 & 0.2353 & 7 & 0.4118 & 4 & 0.2353 \\
\textbf{L(32k)} & 1 & 0.0111 & 1 & 0.0111 & 1 & 0.0111 & 1 & 0.0111 & 1 & 0.0111 & 1 & 0.0111 & 1 & 0.0111 & 1 & 0.0111 & 1 & 0.0111 & 1 & 0.0111 \\
\midrule
\textbf{M} & \multicolumn{2}{c}{\textbf{SCTmp}} & \multicolumn{2}{c}{\textbf{DUC}} & \multicolumn{2}{c}{\textbf{SCL}} & \multicolumn{2}{c}{\textbf{OLI}} & \multicolumn{2}{c}{\textbf{DLA}} & \multicolumn{2}{c}{\textbf{DX}} & \multicolumn{2}{c}{\textbf{HYPX}} & \multicolumn{2}{c}{\textbf{DUS}} & \multicolumn{2}{c}{\textbf{DECG}} & \multicolumn{2}{c}{\textbf{Age}} \\ \midrule
\textbf{G} & 0 & 0 & 0 & 0 & 0 & 0 & 0 & 0 & 0 & 0 & 0 & 0 & 0 & 0 & 0 & 0 & 0 & 0 & 0 & 0 \\
\textbf{G\_j} & 0 & 0 & 0 & 0 & 0 & 0 & 0 & 0 & 0 & 0 & 0 & 0 & 0 & 0 & 0 & 0 & 0 & 0 & 0 & 0 \\
\textbf{G\_s5} & 0 & 0 & 0 & 0 & 0 & 0 & 0 & 0 & 0 & 0 & 0 & 0 & 0 & 0 & 0 & 0 & 0 & 0 & 0 & 0 \\
\textbf{G\_s10} & 0 & 0 & 0 & 0 & 0 & 0 & 0 & 0 & 0 & 0 & 0 & 0 & 0 & 0 & 0 & 0 & 0 & 0 & 0 & 0 \\
\textbf{L(4k)} & 4 & 0.0606 & 3 & 0.0455 & 4 & 0.0606 & 3 & 0.0455 & 3 & 0.0455 & 4 & 0.0606 & 3 & 0.0455 & 3 & 0.0455 & 3 & 0.0455 & 3 & 0.0455 \\
\textbf{M(4k)} & 5 & 0.2941 & 5 & 0.2941 & 5 & 0.2941 & 5 & 0.2941 & 6 & 0.3529 & 6 & 0.3529 & 5 & 0.2941 & 5 & 0.2941 & 5 & 0.2941 & 4 & 0.2353 \\
\textbf{L(32k)} & 1 & 0.0111 & 1 & 0.0111 & 1 & 0.0111 & 1 & 0.0111 & 1 & 0.0111 & 1 & 0.0111 & 1 & 0.0111 & 1 & 0.0111 & 1 & 0.0111 & 1 & 0.0111 \\ \bottomrule
\end{tabularx}
\begin{tablenotes}
\footnotesize
\item \textbf{Attributes Abbreviations:} \textbf{O:G} - org:group; \textbf{LEU} - Leucocytes; \textbf{LA} - LacticAcid; \textbf{IS} - InfectionSuspected; \textbf{DB} - DiagnosticBlood; \textbf{DO} - DisfuncOrg; \textbf{SCT} - SIRSCritTachypnea; \textbf{HYP} - Hypotensie; \textbf{SCRH} - SIRSCritHeartRate; \textbf{INF} - Infusion; \textbf{DAA} - DiagnosticArtAstrup; \textbf{DIC} - DiagnosticIC; \textbf{DS} - DiagnosticSputum; \textbf{DiL} - DiagnosticLiquor; \textbf{DOth} - DiagnosticOther; \textbf{SC2+} - SIRSCriteria2OrMore; \textbf{DX} - DiagnosticXthorax; \textbf{SCTmp} - SIRSCritTemperature; \textbf{DUC} - DiagnosticUrinaryCulture; \textbf{SCL} - SIRSCritLeucos; \textbf{OLI} - Oligurie; \textbf{DLA} - DiagnosticLacticAcid; \textbf{HYPX} - Hypoxie; \textbf{DUS} - DiagnosticUrinarySediment; \textbf{DECG} - DiagnosticECG.
 \end{tablenotes}
\end{threeparttable}
\end{table}

\begin{table}[pos = htbp]
\caption{Complete Results for Helpdesk, BPI13C, and BPI13I}
\label{appe:tab_2}
\centering
\setlength{\tabcolsep}{1.5pt}
\begin{threeparttable}
\begin{tabularx}{\textwidth}{@{} l *{14}{C} @{}}
\toprule[1.5pt]
\multicolumn{15}{c}{\textbf{Helpdesk}} \\
\midrule
\textbf{M} & \textbf{c(S)} & \textbf{JSD} & \textbf{SS} & \textbf{WD} & \textbf{DL} & \textbf{H} & \textbf{re(A)} & \textbf{Ac(A)} & \textbf{MA(T)} & \textbf{\#S} & \textbf{Ac(E)} & \textbf{Ac(S)} & \textbf{I-c} & \textbf{I-pct} \\
 &  &  &  &  &  & & & & & & \textbf{O:RE} & \textbf{C:VA} & \multicolumn{2}{c}{\textbf{C:VA}}\\
\midrule
\textbf{G} & 1.0000 & 0.1779 & 0.9498 & 502069 & 0.9276 & 0 & 0.6250 & 0.8969 & 589028 & 458 & 0.4114 & 0.7380 & 0 & 0.0000 \\
\textbf{G\_j} & 1.0000 & 0.1724 & 0.9502 & 481164 & 0.9280 & 0 & 0.6250 & 0.8974 & 615277 & 458 & 0.4073 & 0.7380 & 0 & 0.0000 \\
\textbf{G\_s5} & 1.0000 & 0.1872 & 0.9505 & 711122 & 0.9289 & 0 & 0.6250 & 0.8992 & 641977 & 458 & 0.4156 & 0.7358 & 0 & 0.0000 \\
\textbf{G\_s10} & 1.0000 & 0.1853 & 0.9496 & 433090 & 0.9275 & 0 & 0.6250 & 0.8951 & 586232 & 458 & 0.4142 & 0.7358 & 0 & 0.0000 \\
\textbf{L(4k)} & 1.0000 & 0.4297 & 0.6488 & 8446087 & 0.4876 & 2 & 1.0000 & 0.2821 & 1102316 & 458 & 0.0782 & 0.0044 & 424 & 0.9258 \\
\textbf{M(4k)} & 0.8908 & 0.3492 & 0.7619 & 4539597 & 0.7257 & 71 & 0.7500 & 0.6130 & 66648683 & 408 & 0.1252 & 0.3734 & 37 & 0.0907 \\
\textbf{L(32k)} & 1.0000 & 0.2561 & 0.8960 & 5254845 & 0.8526 & 1 & 0.8750 & 0.7662 & 2478145 & 458 & 0.1551 & 0.1179 & 0 & 0.0000\\
\bottomrule
\end{tabularx}
\footnotesize
\begin{tablenotes}
 \item \textbf{Attributes Abbreviations:} \textbf{O:RE} - org:resource; \textbf{C:VA} - case:variant.
 \end{tablenotes}
\end{threeparttable}

\begin{tabularx}{\textwidth}{@{} l *{10}{C} @{}}
\toprule[1.5pt]
\multicolumn{11}{c}{\textbf{BPI13C}} \\ \midrule
\textbf{M} & \textbf{c(S)} & \textbf{JSD} & \textbf{SS} & \textbf{WD} & \textbf{DL} & \textbf{H} & \textbf{re(A)} & \textbf{Ac(A)} & \textbf{MA(T)} & \textbf{\#S} \\ \midrule
\textbf{G} & 1.0000 & 0.2945 & 0.8237 & 12449630 & 0.8017 & 0 & 0.8333 & 0.6933 & 8915835 & 148 \\
\textbf{G\_j} & 1.0000 & 0.3412 & 0.8113 & 12326789 & 0.7984 & 0 & 0.8333 & 0.6975 & 9321481 & 148 \\
\textbf{G\_s5} & 1.0000 & 0.3228 & 0.8166 & 12825437 & 0.7999 & 0 & 0.8333 & 0.6933 & 9073631 & 148 \\
\textbf{G\_s10} & 1.0000 & 0.3385 & 0.8113 & 12790696 & 0.7972 & 0 & 0.8333 & 0.6891 & 9127482 & 148 \\
\textbf{L(4k)} & 0.8311 & 0.5903 & 0.4840 & 14198777 & 0.4681 & 162 & 1.0000 & 0.3081 & 181447374 & 123 \\
\textbf{M(4k)} & 0.6149 & 0.9099 & 0.3645 & 13194896 & 0.3528 & 63 & 0.8333 & 0.1765 & 41116259 & 91 \\
\textbf{L(32k)} & 1.0000 & 0.2071 & 0.7939 & 11440188 & 0.7566 & 1 & 0.8333 & 0.6204 & 10513514 & 148\\
\midrule
\end{tabularx}

\begin{tabularx}{\textwidth}{@{} l *{14}{C} @{}}
\textbf{M} & \multicolumn{5}{c}{\textbf{Ac(E)}} & \multicolumn{3}{c}{\textbf{Ac(S)}} & \textbf{I-c} & \textbf{I-pct} & \textbf{I-c} & \textbf{I-pct} & \textbf{I-c} & \textbf{I-pct} \\
\textbf{} & \textbf{O:G} & \textbf{RC} & \textbf{O:RE} & \textbf{OI} & \textbf{O:RO} & \textbf{OC} & \textbf{IMP} & \textbf{PROD} & \multicolumn{2}{c}{\textbf{OC}} & \multicolumn{2}{c}{\textbf{IMP}} & \multicolumn{2}{c}{\textbf{PROD}} \\ \midrule
\textbf{G} & 0.4538 & 0.6232 & 0.1737 & 0.2227 & 0.3754 & 0.5338 & 0.5743 & 0.1081 & 0 & 0 & 0 & 0 & 0 & 0 \\
\textbf{G\_j} & 0.4846 & 0.6345 & 0.1723 & 0.2381 & 0.3613 & 0.5270 & 0.5743 & 0.1081 & 0 & 0 & 0 & 0 & 0 & 0 \\
\textbf{G\_s5} & 0.4776 & 0.6317 & 0.1751 & 0.2311 & 0.3627 & 0.5338 & 0.5743 & 0.1284 & 0 & 0 & 0 & 0 & 0 & 0 \\
\textbf{G\_s10} & 0.4734 & 0.6232 & 0.1751 & 0.2395 & 0.3697 & 0.5338 & 0.5811 & 0.1149 & 0 & 0 & 0 & 0 & 0 & 0 \\
\textbf{L(4k)} & 0.0952 & 0.1919 & 0.0126 & 0.0154 & 0.077 & 0.1689 & 0.2365 & 0.0068 & 8 & 0.065 & 6 & 0.0488 & 6 & 0.0488 \\
\textbf{M(4k)} & 0.021 & 0.056 & 0.0028 & 0.0028 & 0.07 & 0.1149 & 0.1689 & 0 & 4 & 0.044 & 1 & 0.011 & 1 & 0.011 \\
\textbf{L(32k)} & 0.1597 & 0.3235 & 0.0084 & 0.0266 & 0.0616 & 0.2973 & 0.4797 & 0.0135 & 6 & 0.0405 & 1 & 0.0068 & 0 & 0 
\end{tabularx}
\begin{tabularx}{\textwidth}{@{} l *{10}{C} @{}}
\toprule[1.5pt]
\multicolumn{11}{c}{\textbf{BPI13I}} \\ \midrule
\textbf{M} & \textbf{c(S)} & \textbf{JSD} & \textbf{SS} & \textbf{WD} & \textbf{DL} & \textbf{H} & \textbf{re(A)} & \textbf{Ac(A)} & \textbf{MA(T)} & \textbf{\#S} \\ \midrule
\textbf{G} & 1.0000 & 0.2133 & 0.8325 & 748775 & 0.8190 & 0 & 0.9167 & 0.6552 & 459020 & 755 \\
\textbf{G\_j} & 1.0000 & 0.2547 & 0.8301 & 855279 & 0.8232 & 0 & 0.7500 & 0.6603 & 459483 & 755 \\
\textbf{G\_s5} & 1.0000 & 0.2586 & 0.8289 & 624653 & 0.8218 & 0 & 0.7500 & 0.6630 & 454386 & 755 \\
\textbf{G\_s10} & 1.0000 & 0.2208 & 0.8307 & 652814 & 0.8229 & 0 & 0.9167 & 0.6540 & 454590 & 755 \\
\textbf{L(4k)} & 0.7642 & 0.7407 & 0.3714 & 950069 & 0.3585 & 1133 & 0.6667 & 0.1462 & 266296975 & 577 \\
\textbf{M(4k)} & 0.3510 & 1.3093 & 0.0833 & 1113165 & 0.0698 & 708 & 0.7500 & 0.0421 & 5338296 & 265 \\
\textbf{L(32k)} & 1.0000 & 0.2605 & 0.7582 & 897646 & 0.7227 & 8 & 0.7500 & 0.5007 & 1175517 & 755\\
\midrule
\end{tabularx}

\begin{threeparttable}
\begin{tabularx}{\textwidth}{@{} l *{14}{C} @{}}
\textbf{M} & \multicolumn{5}{c}{\textbf{Ac(E)}} & \multicolumn{3}{c}{\textbf{Ac(S)}} & \textbf{I-c} & \textbf{I-pct} & \textbf{I-c} & \textbf{I-pct} & \textbf{I-c} & \textbf{I-pct} \\
\textbf{} & \textbf{O:G} & \textbf{RC} & \textbf{O:RE} & \textbf{OI} & \textbf{O:RO} & \textbf{OC} & \textbf{IMP} & \textbf{PROD} & \multicolumn{2}{c}{\textbf{OC}} & \multicolumn{2}{c}{\textbf{IMP}} & \multicolumn{2}{c}{\textbf{PROD}}\\ \midrule
\textbf{G} & 0.1961 & 0.4062 & 0.1132 & 0.6478 & 0.4727 & 0.4318 & 0.5894 & 0.1404 & 0 & 0 & 0 & 0 & 0 & 0 \\
\textbf{G\_j} & 0.2019 & 0.4107 & 0.1182 & 0.6561 & 0.4804 & 0.4172 & 0.5894 & 0.1470 & 0 & 0 & 0 & 0 & 0 & 0 \\
\textbf{G\_s5} & 0.1963 & 0.3976 & 0.1164 & 0.6523 & 0.4846 & 0.4159 & 0.5881 & 0.1417 & 0 & 0 & 0 & 0 & 0 & 0 \\
\textbf{G\_s10} & 0.2034 & 0.4046 & 0.1134 & 0.6506 & 0.4781 & 0.4331 & 0.5974 & 0.1391 & 0 & 0 & 0 & 0 & 0 & 0 \\
\textbf{L(4k)} & 0.0071 & 0.0431 & 0.0144 & 0.1266 & 0.056 & 0.0901 & 0.2132 & 0.0026 & 34 & 0.0589 & 25 & 0.0433 & 26 & 0.0451 \\
\textbf{M(4k)} & 0.0022 & 0.0164 & 0.0006 & 0.0738 & 0.0251 & 0.0371 & 0.0808 & 0.004 & 16 & 0.0604 & 22 & 0.083 & 21 & 0.0792 \\
\textbf{L(32k)} & 0.0155 & 0.1421 & 0.0314 & 0.3805 & 0.2038 & 0.1364 & 0.4437 & 0.0185 & 44 & 0.0583 & 13 & 0.0172 & 14 & 0.0185 \\ \bottomrule
\end{tabularx}
\begin{tablenotes}
\footnotesize
\item \textbf{Attributes Abbreviations:} \textbf{O:G} - org:group; \textbf{RC} - resource country; \textbf{O:RE} - org:resource; \textbf{OI} - organization involved; \textbf{O:RO} - org:role; \textbf{OC} - organization country; \textbf{IMP} - impact; \textbf{PROD} - product
\end{tablenotes}    
\end{threeparttable}
\end{table}

\begin{table}[pos = htbp]
\caption{Complete Results for BPI20 and BPI17}
\label{appe:tab_3}
\centering
\setlength{\tabcolsep}{1.5pt}
\begin{tabularx}{\textwidth}{@{} l *{10}{C} @{}}
\toprule[1.5pt]
\multicolumn{11}{c}{\textbf{BPI120}} \\ \midrule
\textbf{M} & \textbf{c(S)} & \textbf{JSD} & \textbf{SS} & \textbf{WD} & \textbf{DL} & \textbf{H} & \textbf{re(A)} & \textbf{Ac(A)} & \textbf{MA(T)} & \textbf{\#S} \\ \midrule
\textbf{G} & 1.0000 & 0.2012 & 0.8856 & 1814361 & 0.8422 & 0 & 0.6667 & 0.7999 & 1334086 & 209 \\
\textbf{G\_j} & 1.0000 & 0.2033 & 0.8854 & 1462130 & 0.8429 & 0 & 0.6667 & 0.8047 & 1331992 & 209 \\
\textbf{G\_s5} & 1.0000 & 0.2054 & 0.8962 & 1914524 & 0.8583 & 0 & 0.6667 & 0.8153 & 1336757 & 209 \\
\textbf{G\_s10} & 1.0000 & 0.2108 & 0.8899 & 1826933 & 0.8506 & 0 & 0.6667 & 0.8100 & 1348357 & 209 \\
\textbf{L(4k)} & 0.756 & 0.7147 & 0.3398 & 2707083 & 0.2964 & 421 & 0.5926 & 0.1476 & 74200704 & 158 \\
\textbf{M(4k)} & 0.3158 & 1.3999 & 0.1003 & 4011717 & 0.0882 & 227 & 0.5556 & 0.0552 & 55912405 & 66 \\
\textbf{L(32k)} & 1.0000 & 0.2584 & 0.8211 & 1987999 & 0.7396 & 19 & 0.8519 & 0.5435 & 3012057 & 209 \\
\midrule
\end{tabularx}

\begin{threeparttable}
\begin{tabularx}{\textwidth}{@{} l *{13}{C} @{}}
\textbf{} & \textbf{Ac(E)} & \multicolumn{2}{c}{\textbf{Ac(S)}} & \multicolumn{2}{c}{\textbf{MA(S)}} & \textbf{I-c} & \textbf{I-pct} & \textbf{I-c} & \textbf{I-pct} & \textbf{I-c} & \textbf{I-pct} & \textbf{I-c} & \textbf{I-pct} \\
\textbf{} & \textbf{O:RE} & \textbf{C:OE} & \textbf{C:PR} & \textbf{C:RA} & \textbf{C:PRB} & \multicolumn{2}{c}{\textbf{C:OE}} & \multicolumn{2}{c}{\textbf{C:PR}} & \multicolumn{2}{c}{\textbf{C:CRA}} & \multicolumn{2}{c}{\textbf{C:PRB}} \\ 
\midrule
\textbf{G} & 0.8286 & 0.1962 & 0.9761 & 459.03 & 1164.42 & 0 & 0 & 0 & 0 & 0 & 0 & 0 & 0 \\
\textbf{G\_j} & 0.8312 & 0.2297 & 0.9761 & 467.60 & 1173.89 & 0 & 0 & 0 & 0 & 0 & 0 & 0 & 0 \\
\textbf{G\_s5} & 0.8376 & 0.2153 & 0.9761 & 455.14 & 1133.47 & 0 & 0 & 0 & 0 & 0 & 0 & 0 & 0 \\
\textbf{G\_s10} & 0.8339 & 0.2679 & 0.9761 & 461.43 & 1102.62 & 0 & 0 & 0 & 0 & 0 & 0 & 0 & 0 \\
\textbf{L(4k)} & 0.1783 & 0.0048 & 0.3445 & 15637.64 & 9457.47 & 13 & 0.0823 & 12 & 0.0759 & 12 & 0.0759 & 10 & 0.0633 \\
\textbf{M(4k)} & 0.0398 & 0.0144 & 0.0574 & 2450.51 & 6003.79 & 1 & 0.0152 & 1 & 0.0152 & 2 & 0.0303 & 5 & 0.0758 \\
\textbf{L(32k)} & 0.6253 & 0.0431 & 0.9282 & 732.80 & 1798.13 & 0 & 0 & 0 & 0 & 1 & 0.0048 & 1 & 0.0048 \\ \bottomrule
\end{tabularx}
\begin{tablenotes}
\footnotesize
\item \textbf{Attributes Abbreviations:} \textbf{O:RE} - org:resource; \textbf{C:OE} - case:OrganizationalEntity; \textbf{C:PR} - case:Project; \textbf{C:RA} - case:RequestedAmount; \textbf{C:PRB} - case:Permit RequestedBudget.     
 \end{tablenotes}
\end{threeparttable}
\begin{tabularx}{\textwidth}{@{} l *{10}{C} @{}}
\toprule[1.5pt]
\multicolumn{11}{c}{\textbf{BPI17}} \\ \midrule
\textbf{M} & \textbf{c(S)} & \textbf{JSD} & \textbf{SS} & \textbf{WD} & \textbf{DL} & \textbf{H} & \textbf{re(A)} & \textbf{Ac(A)} & \textbf{MA(T)} & \textbf{\#S} \\ \midrule
\textbf{G} & 1.0000 & 0.2238 & 0.7526 & 418785 & 0.6352 & 0 & 0.8833 & 0.3353 & 429831 & 3150 \\
\textbf{G\_j} & 1.0000 & 0.2157 & 0.7542 & 445256 & 0.6359 & 0 & 0.8833 & 0.3366 & 419696 & 3150 \\
\textbf{G\_s5} & 1.0000 & 0.2140 & 0.7531 & 539976 & 0.6339 & 0 & 0.8833 & 0.3319 & 432232 & 3150 \\
\textbf{G\_s10} & 1.0000 & 0.2243 & 0.7551 & 375357 & 0.6379 & 0 & 0.8667 & 0.3386 & 521810 & 3150 \\
\textbf{L(4k)} & 0.7419 & 0.9814 & 0.0461 & 1724366 & 0.0323 & 13195 & 0.5167 & 0.0023 & 216544974 & 2337 \\
\textbf{M(4k)} & 0.2251 & 1.5629 & 0.0073 & 3530745 & 0.0041 & 5110 & 0.5333 & 0.0007 & 30360903 & 709\\
\textbf{M(32k)} & 0.5889 & 0.9157 & 0.1424 & 5691146 & 0.1125 & 4321 & 0.7833 & 0.042 & 223751169 & 1855\\
\midrule
\end{tabularx}

\begin{tabularx}{\textwidth}{@{} l *{11}{C} @{}}
\textbf{M} & \multicolumn{6}{c}{\textbf{Ac(E)}} & \multicolumn{5}{c}{\textbf{MA(E)}} \\
\textbf{} &\textbf{ACT} & \textbf{O:RE} & \textbf{EO} & \textbf{ACC} & \textbf{SEL} & \textbf{OID} & \textbf{FWA} & \textbf{NOT} & \textbf{MC} & \textbf{CS} & \textbf{OA} \\ \midrule
\textbf{G} & 0.4574 & 0.1338 & 0.6943 & 0.7024 & 0.7914 & 0.2175 & 5903.11 & 30.59 & 123.80 & 316.22 & 9122.74 \\
\textbf{G\_j} & 0.4620 & 0.1353 & 0.6957 & 0.7046 & 0.7908 & 0.2192 & 5945.11 & 30.23 & 114.31 & 341.23 & 8690.36 \\
\textbf{G\_s5} & 0.4627 & 0.1341 & 0.6956 & 0.7040 & 0.7917 & 0.2163 & 5969.32 & 30.05 & 113.80 & 341.20 & 8449.05 \\
\textbf{G\_s10} & 0.4618 & 0.1306 & 0.6971 & 0.7034 & 0.7912 & 0.2140 & 5955.95 & 29.99 & 115.74 & 339.09 & 8515.03 \\
\textbf{L(4k)} & 0.0107 & 0.0019 & 0.0114 & 0.0167 & 0.0230 & 0.0037 & 195038.56 & 351.96 & 2033.93 & 1106.19 & 195323.68 \\
\textbf{M(4k)} & 0.0131 & 0.0009 & 0.0326 & 0.0335 & 0.0388 & 0.0018 & 54289.57 & 510.75 & 3432.69 & 1064.33 & 358433821.80 \\
\textbf{M(32k)} & 0.0609 & 0.0115 & 0.0871 & 0.1039 & 0.1049 & 0.0453 & 69287.93 & 346.28 & 1932.85 & 996.84 & 299198.92 \\
\midrule
\end{tabularx}

\begin{threeparttable}
\begin{tabularx}{\textwidth}{@{} l *{9}{C} @{}}
\textbf{M} & \multicolumn{2}{c}{\textbf{Ac(S)}} & \textbf{MA(S)} & \textbf{I-c} & \textbf{I-pct} & \textbf{I-c} & \textbf{I-pct} & \textbf{I-c} & \textbf{I-pct} \\
\textbf{} &\textbf{C:LG} & \textbf{C:AT} & \textbf{C:RA} & \multicolumn{2}{c}{\textbf{C:LG}} & \multicolumn{2}{c}{\textbf{C:AT}} & \multicolumn{2}{c}{\textbf{C:RA}} \\ \midrule
\textbf{G} & 0.2781 & 0.8914 & 10479.63 & 0 & 0 & 0 & 0 & 0 & 0 \\
\textbf{G\_j} & 0.2740 & 0.8886 & 11375.47 & 0 & 0 & 0 & 0 & 0 & 0 \\
\textbf{G\_s5} & 0.2746 & 0.8863 & 11342.45 & 0 & 0 & 0 & 0 & 0 & 0 \\
\textbf{G\_s10} & 0.2717 & 0.8867 & 11547.99 & 0 & 0 & 0 & 0 & 0 & 0 \\
\textbf{L(4k)} & 0.0165 & 0.0708 & 176838.037 & 33 & 0.0141 & 31 & 0.0133 & 38 & 0.0163 \\
\textbf{M(4k)} & 0.0057 & 0.0257 & 281286.396 & 40 & 0.0564 & 62 & 0.0874 & 62 & 0.0874\\
\textbf{M(32k)} & 0.0714 & 0.2657 & 249466.684 & 142 & 0.0765 & 133 & 0.0717 & 215 & 0.1159\\
\bottomrule
\end{tabularx}
\begin{tablenotes}
\footnotesize
\item \textbf{Attributes Abbreviations:} \textbf{ACT} - Action; \textbf{O:RE} - org:resource; \textbf{EO} - EventOrigin; \textbf{ACC} - Accepted; \textbf{SEL} - Selected; \textbf{OID} - OfferID; \textbf{FWA} - FirstWithdrawalAmount; \textbf{NOT} - NumberOfTerms; \textbf{MC} - MonthlyCost; \textbf{CS} - CreditScore; \textbf{OA} - OfferedAmount; \textbf{C:LG} - case:LoanGoal; \textbf{C:AT} - case:ApplicationType; \textbf{C:RA} - case:RequestedAmount.
 \end{tablenotes}
\end{threeparttable}

\end{table}

\pagebreak
\section{Additional Interpretability Analysis Results}
\label{app:inter}

 \begin{figure}[pos=htbp]
  \centering
  \includegraphics[width=0.98\textwidth]{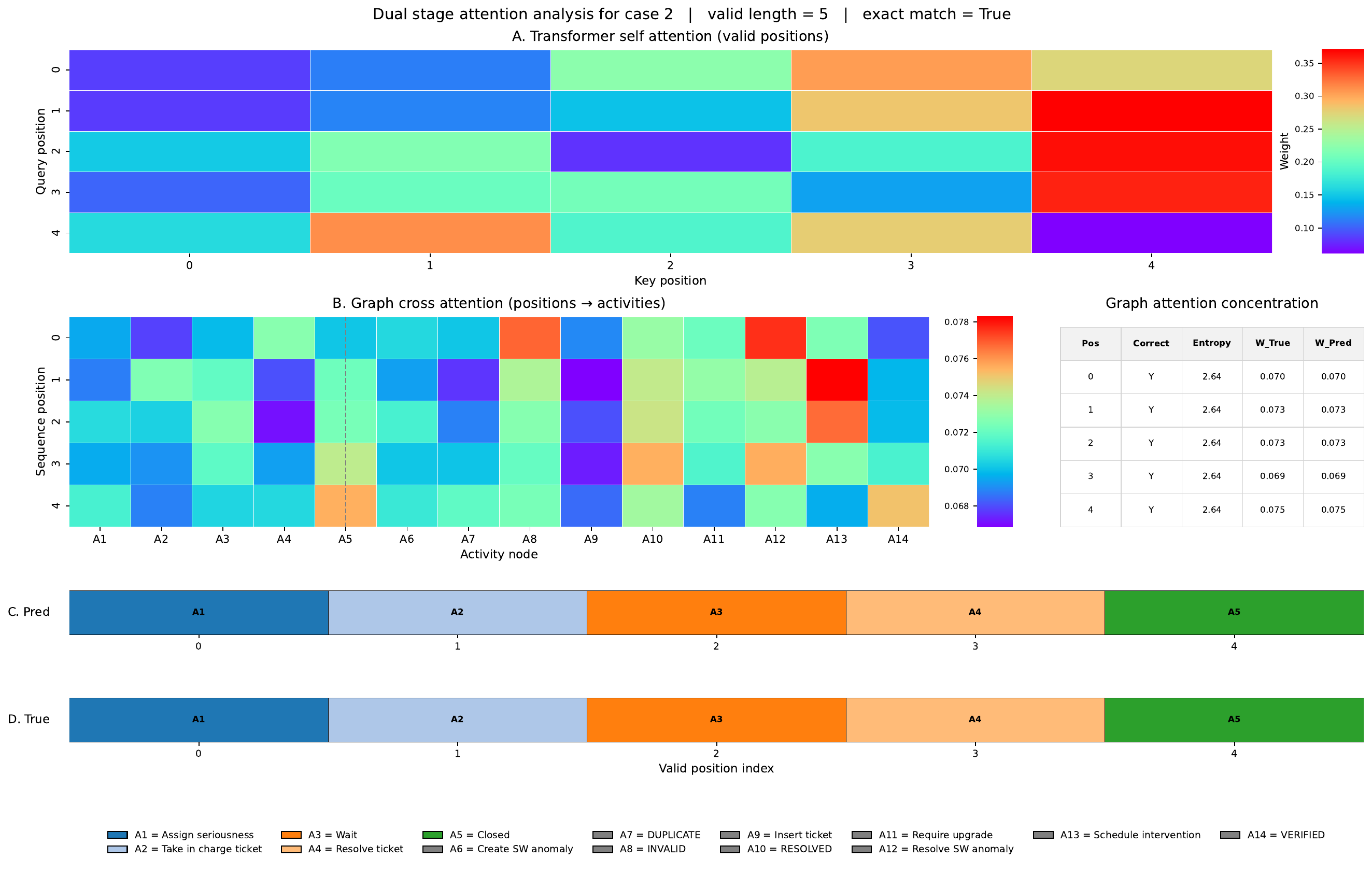}
  \caption{Dual Stage Attention Analysis Panel of Representative Samples from Helpdesk}
  \label{appfig:dual_attn_3}
 \end{figure}
 
\begin{figure}[pos=htbp]
  \centering
  \includegraphics[width=0.98\textwidth]{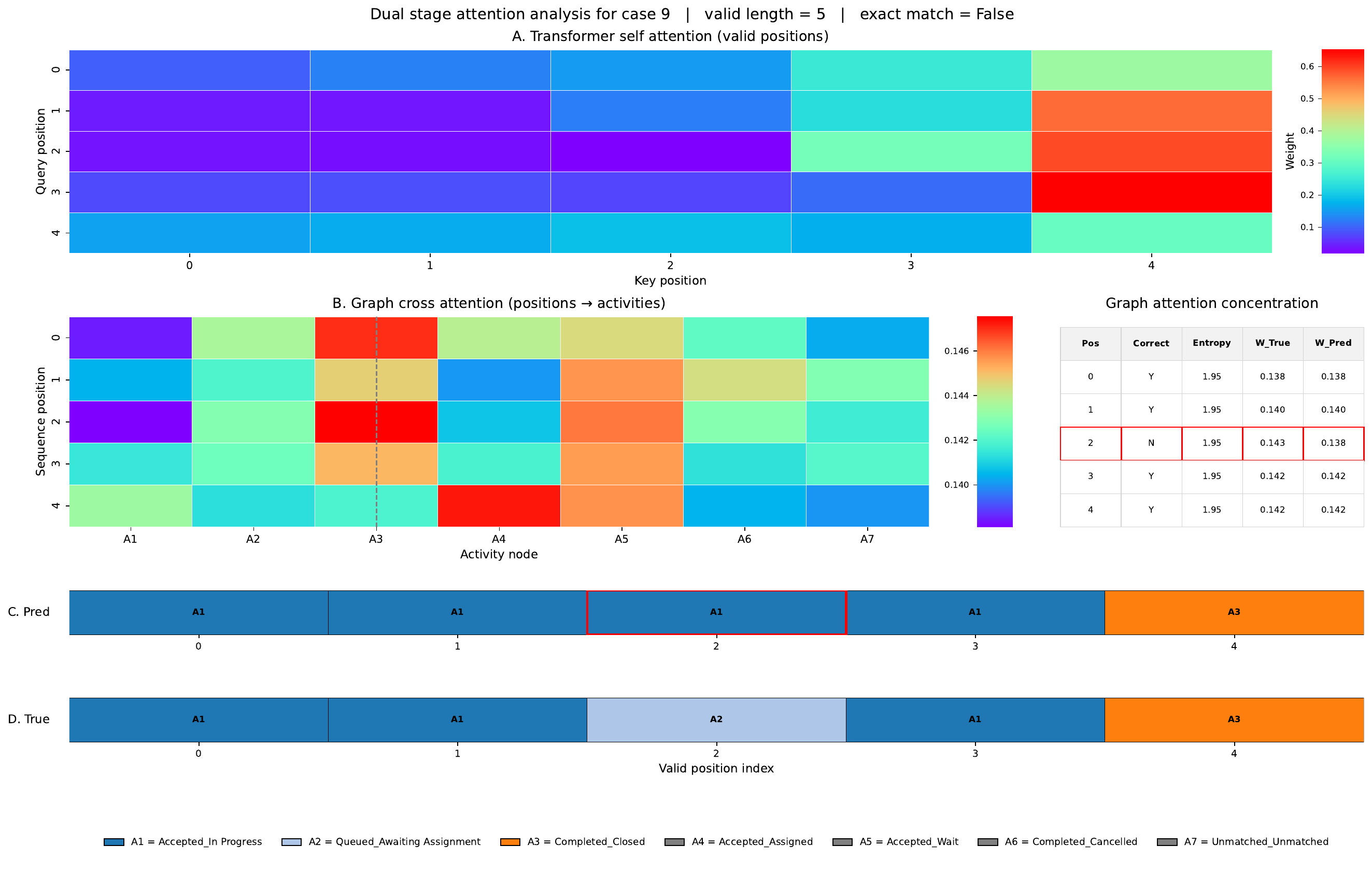}\\[-4mm]
  {\footnotesize\textbf{(a) BPI13C}}\\
  
  \includegraphics[width=0.98\textwidth]{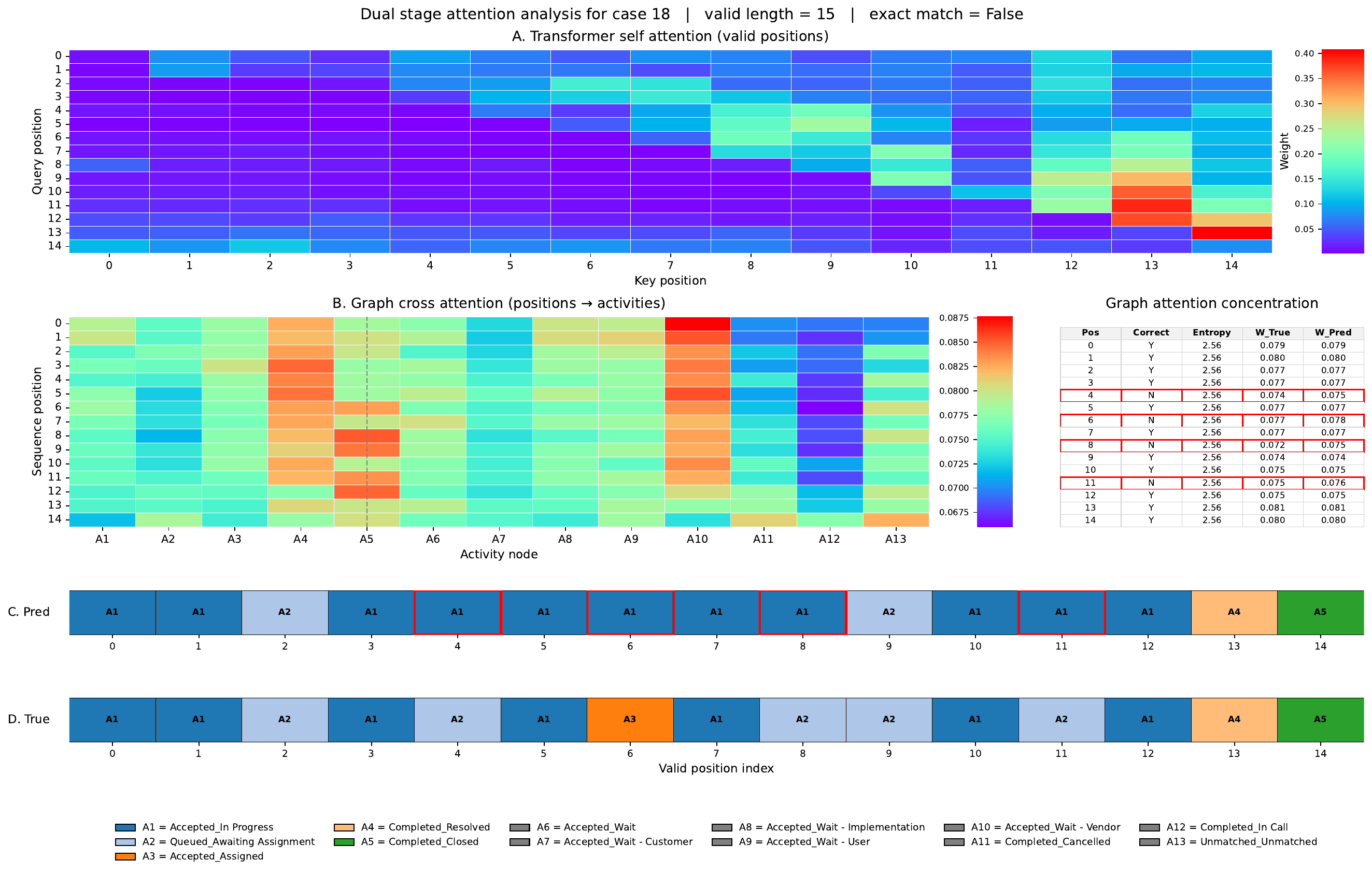}\\[-4mm]
  {\footnotesize\textbf{(b) BPI13I}}\\
  \caption{Dual Stage Attention Analysis Panel of Representative Samples from BPI13C and BPI13I}
  \label{appfig:dual_attn_1}
\end{figure}

\begin{figure}[pos=htbp]
  \centering
  \includegraphics[width=0.98\textwidth]{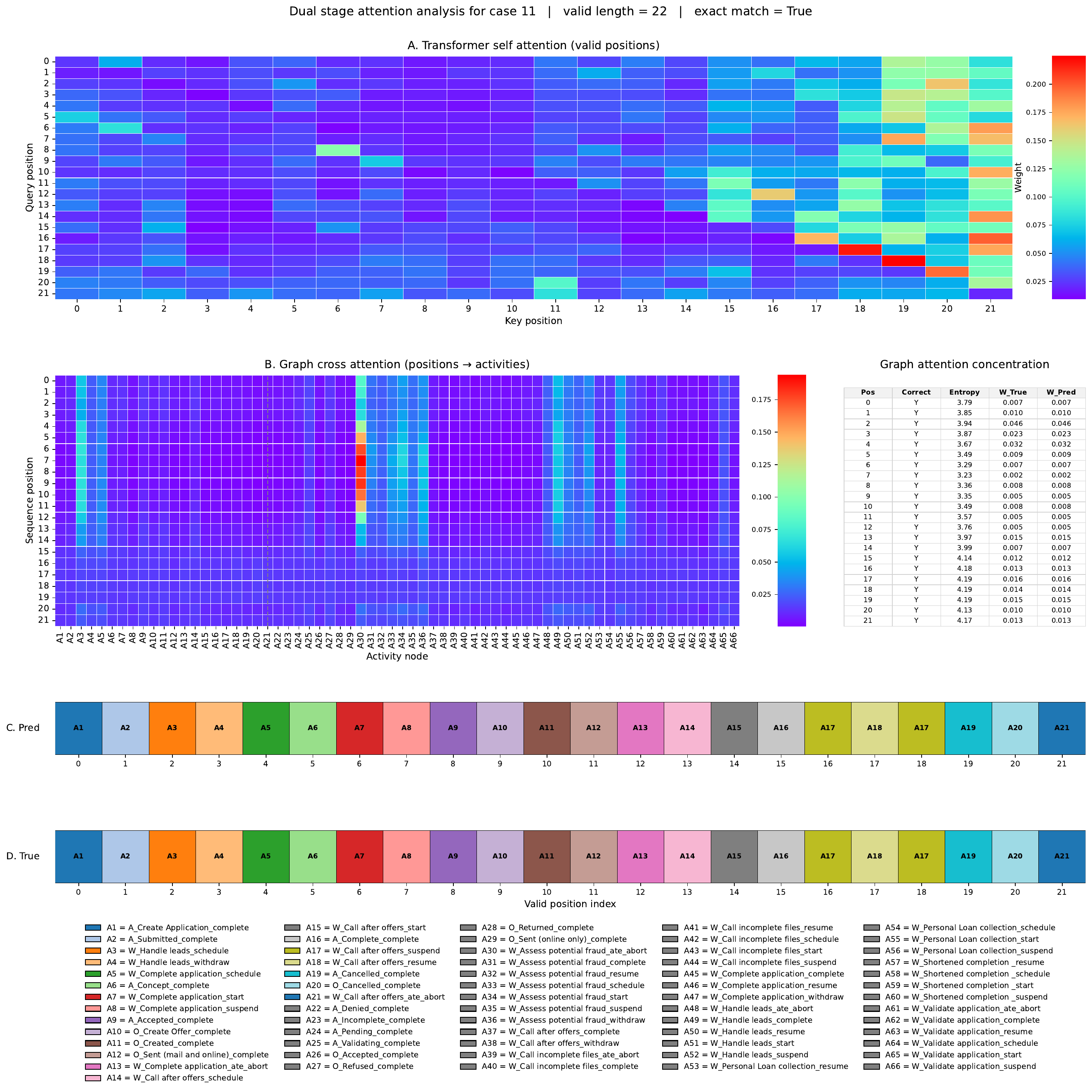}
  \caption{Dual Stage Attention Analysis Panel of Representative Samples from BPI17}
  \label{appfig:dual_attn_2}
 \end{figure}

\begin{figure}[pos=htbp]
  \centering
  \begin{subfigure}[b]{0.48\textwidth}
    \centering
    \includegraphics[width=\linewidth, trim={0 0 0 7mm}, clip]{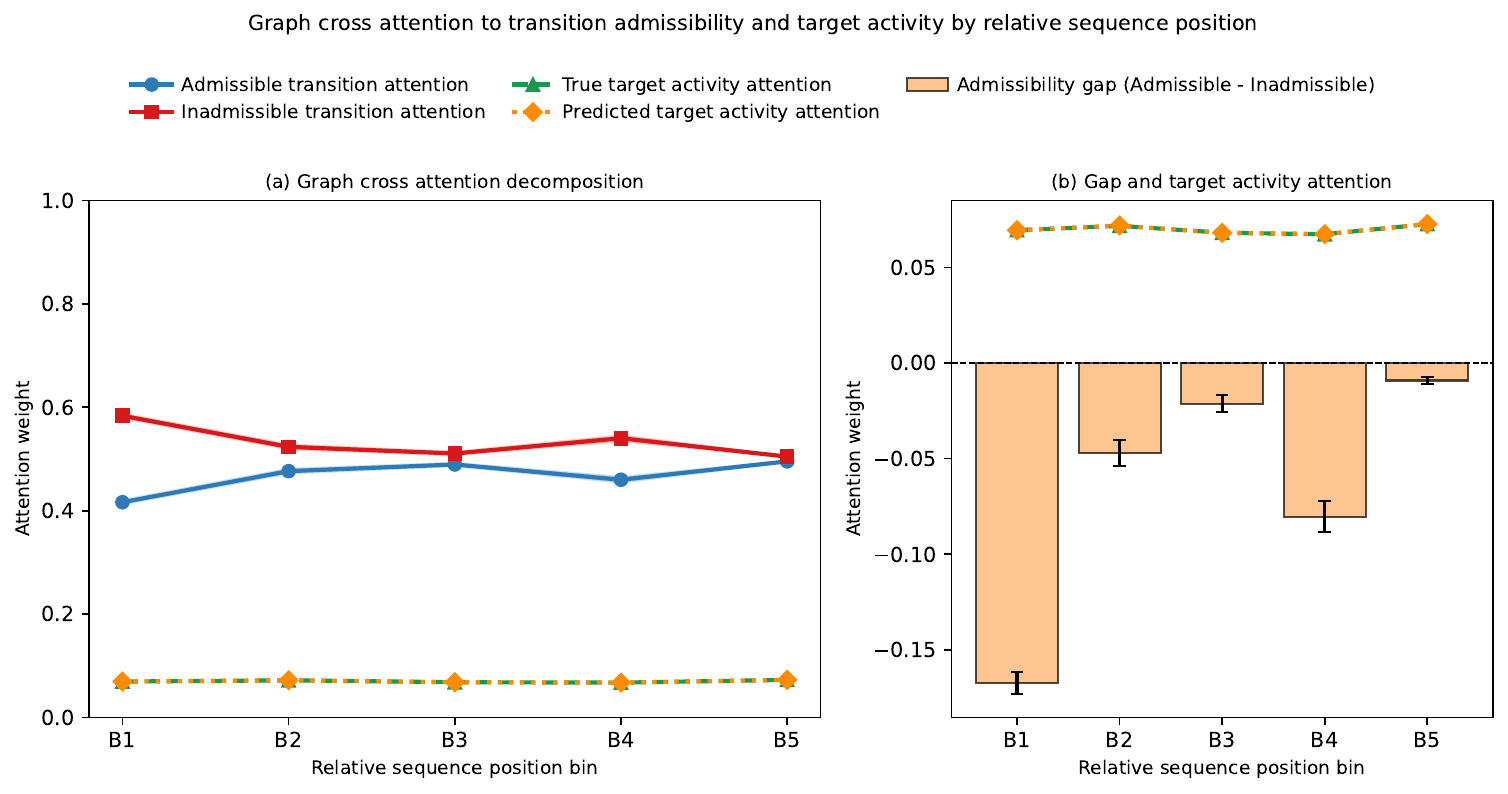}\\
    {\footnotesize\textbf{(a) Helpdesk}}
  \end{subfigure}
  \hfill
  \begin{subfigure}[b]{0.48\textwidth}
    \centering
    \includegraphics[width=\linewidth, trim={0 0 0 7mm}, clip]{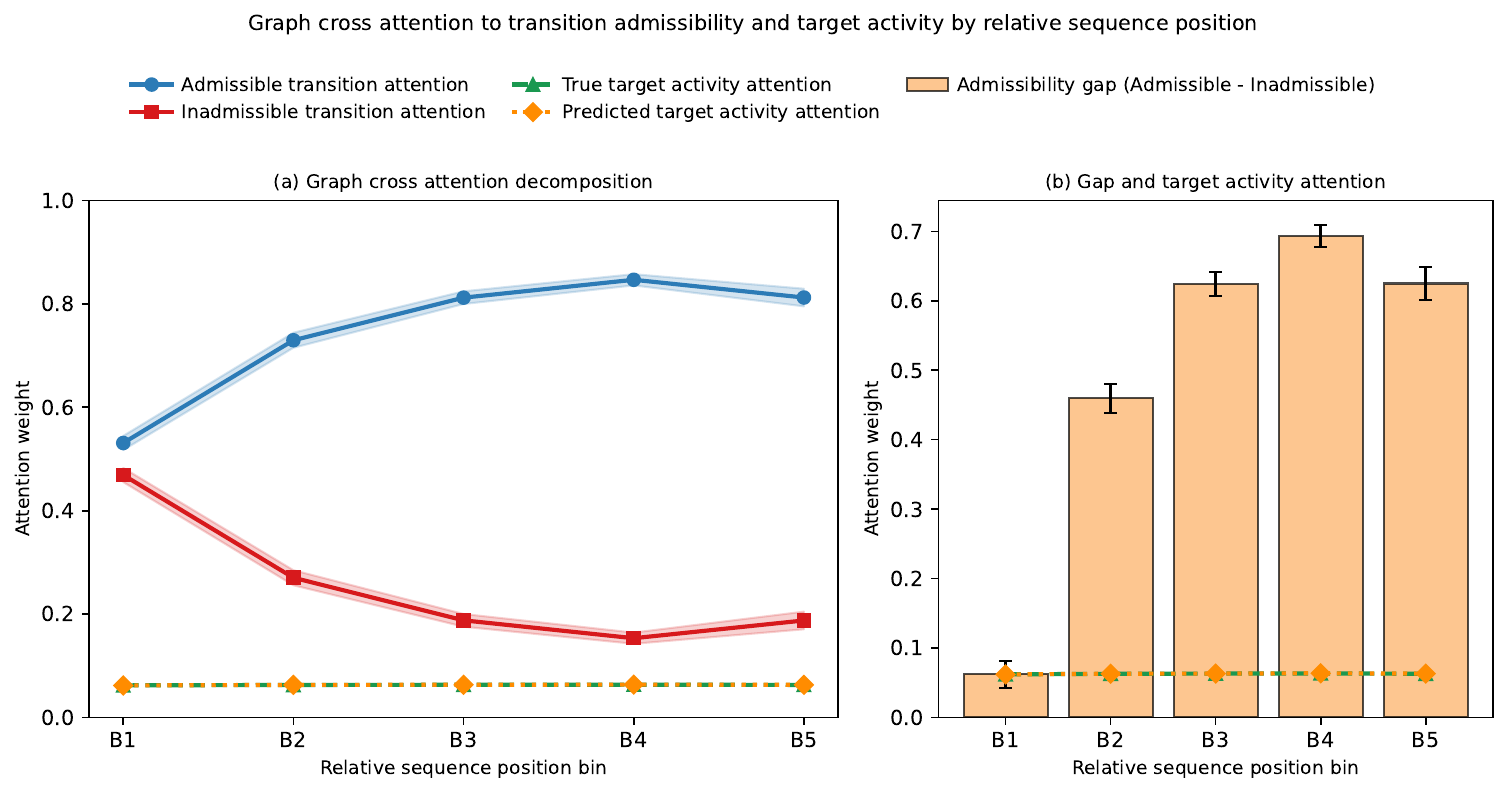}
    \\
    {\footnotesize\textbf{(b) Sepsis}}
  \end{subfigure}  
  
    \begin{subfigure}[b]{0.48\textwidth}
    \centering
    \includegraphics[width=\linewidth, trim={0 0 0 7mm}, clip]{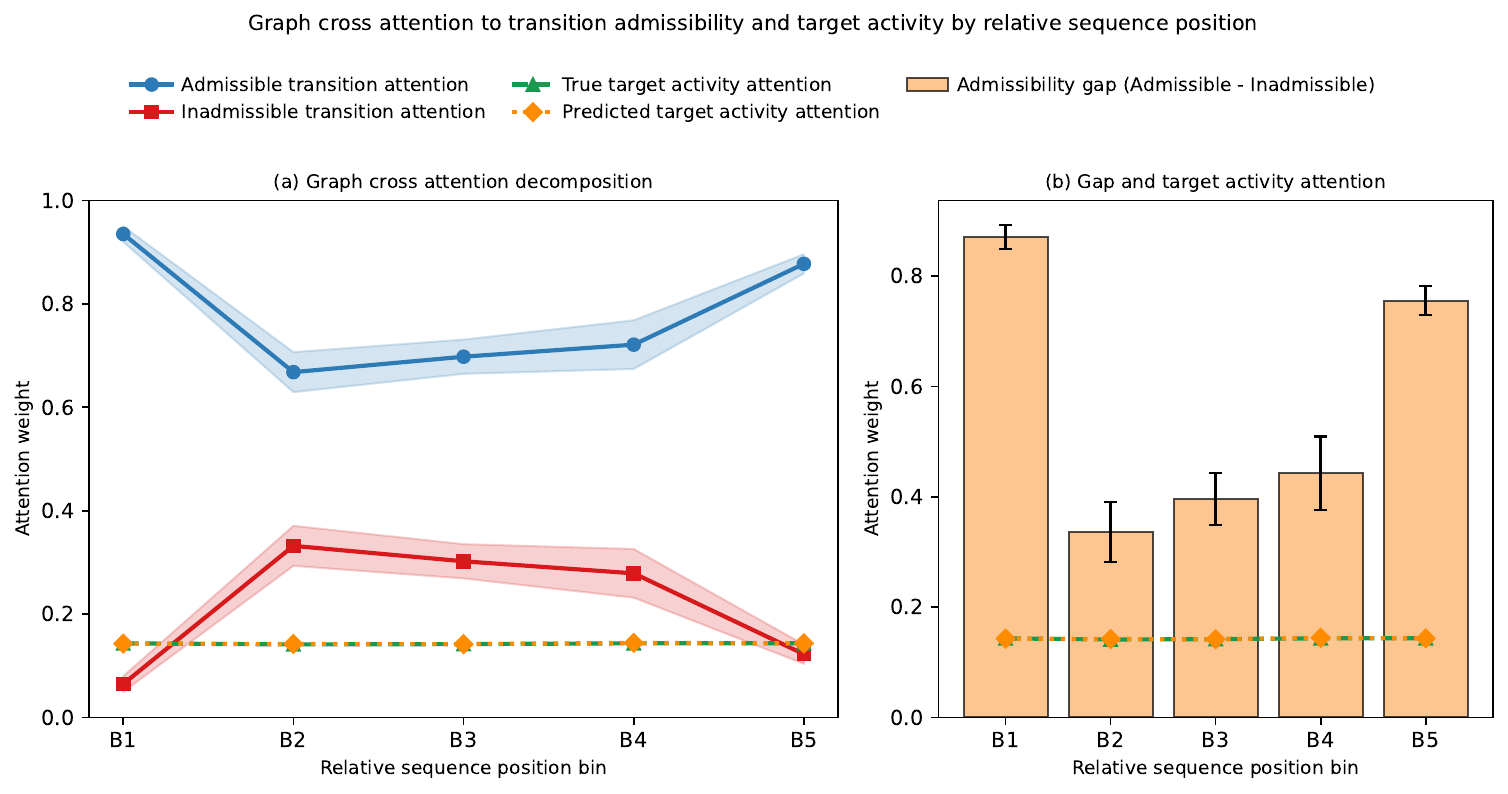}\\
    {\footnotesize\textbf{(c) BPI13C}}
  \end{subfigure}
  \hfill
  \begin{subfigure}[b]{0.48\textwidth}
    \centering
    \includegraphics[width=\linewidth, trim={0 0 0 7mm}, clip]{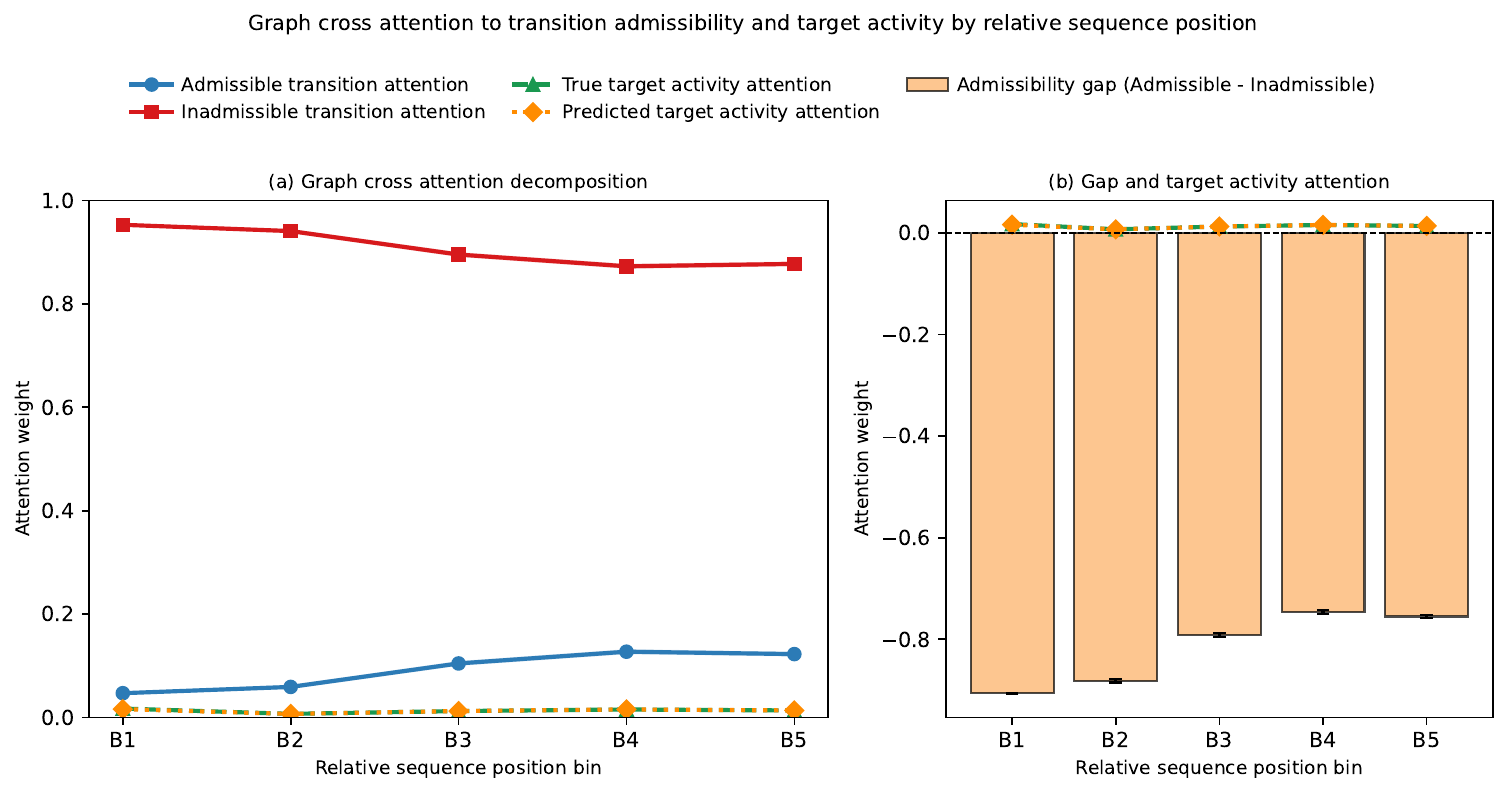}\\
    {\footnotesize\textbf{(d) BPI17}}
  \end{subfigure}    
  % --- Main caption for the entire figure ---
  \caption{Graph Grounded Cross Attention to Transition Admissibility and Target Activity for Helpdesk, Sepsis, BPI13C and BPI17}
  \label{appfig:inter_mass}
\end{figure}

\begin{figure}[pos=htbp]
  \centering
    \begin{subfigure}[b]{0.48\textwidth}
    \centering
    \includegraphics[width=\linewidth]{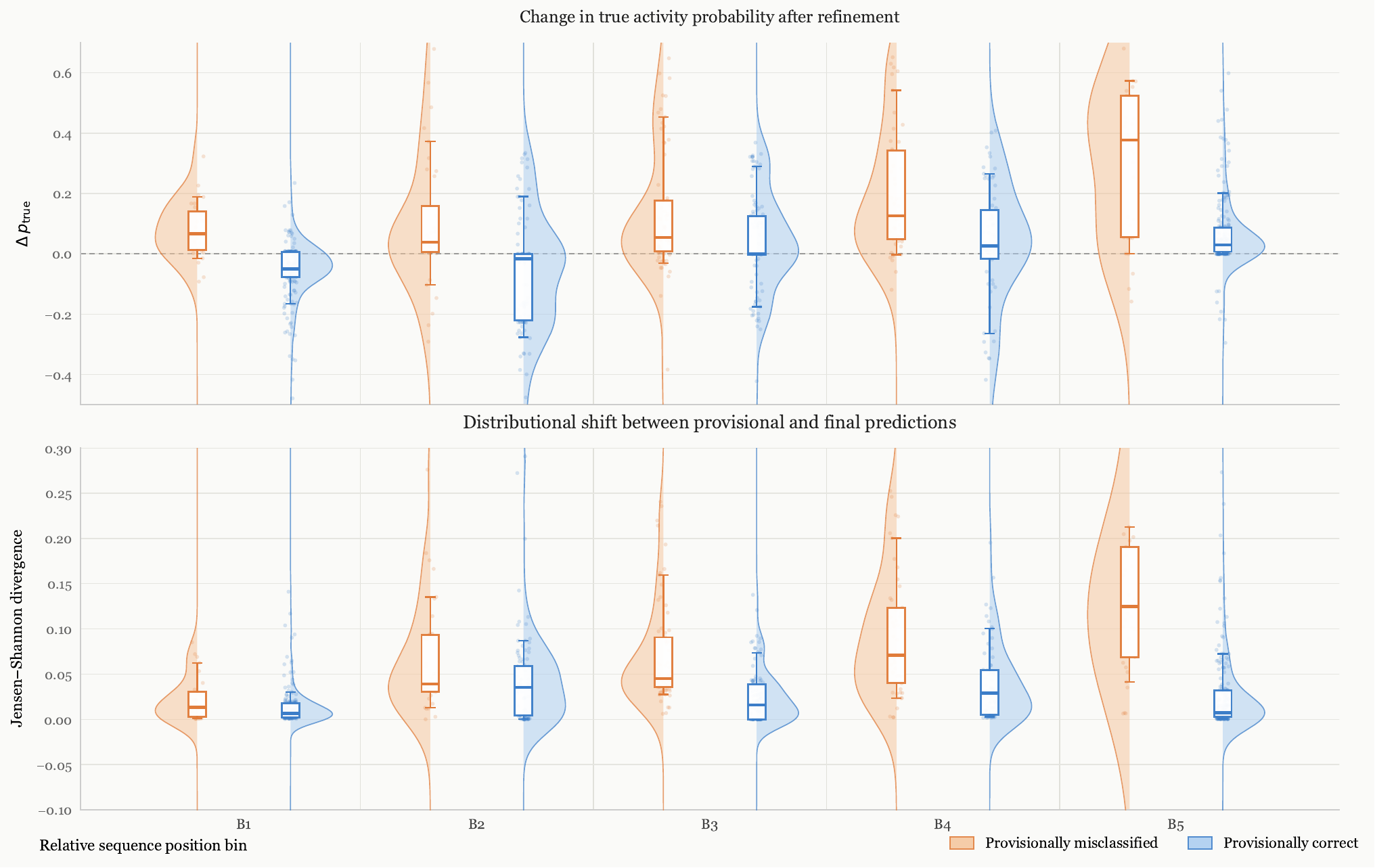}
    \\
    {\footnotesize\textbf{(a) BPI13I}}
  \end{subfigure}
  \hfill
  \begin{subfigure}[b]{0.48\textwidth}
    \centering
    \includegraphics[width=\linewidth]{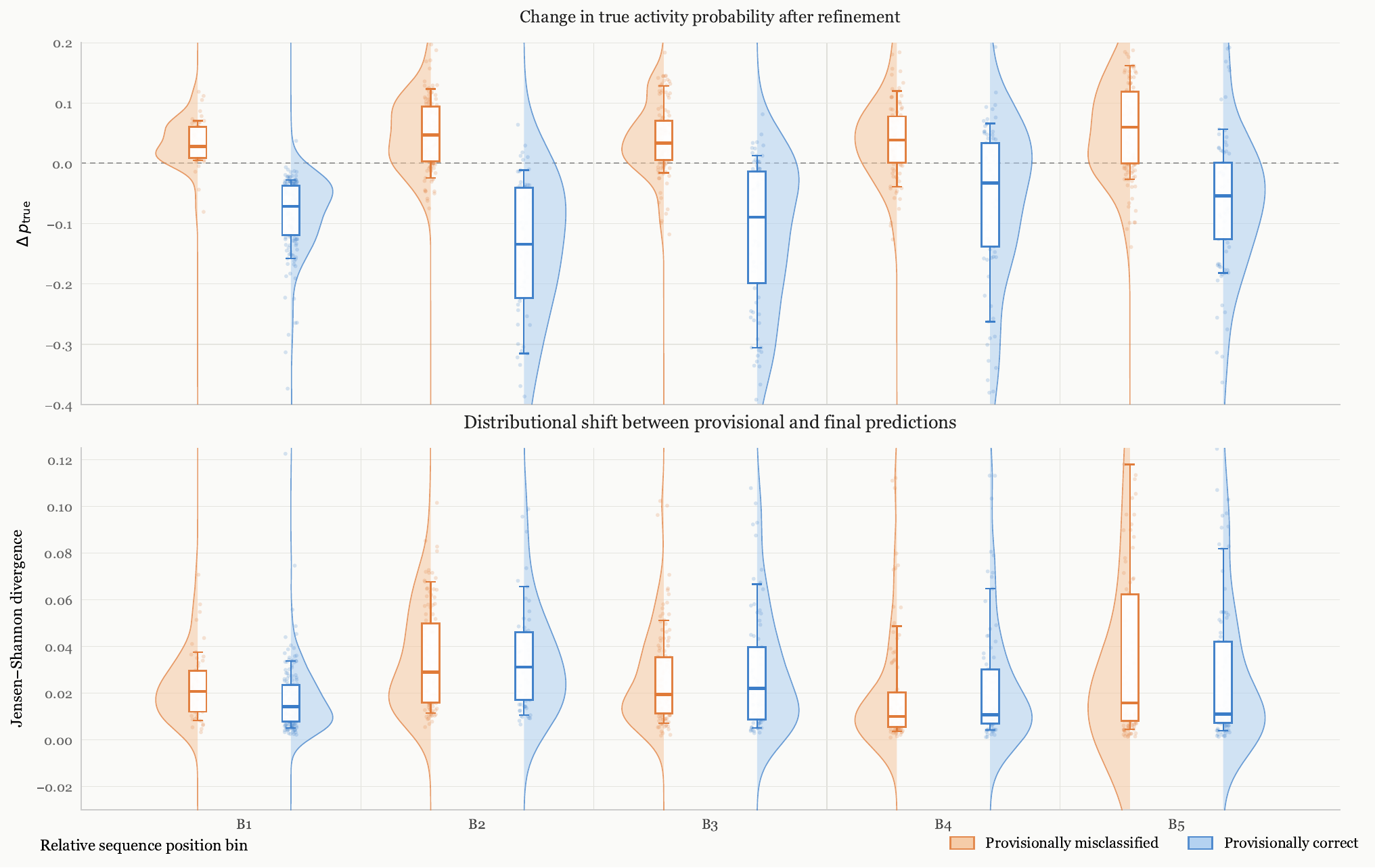}
    \\
    {\footnotesize\textbf{(b) Sepsis}}
  \end{subfigure}  
      \begin{subfigure}[b]{0.48\textwidth}
    \centering
    \includegraphics[width=\linewidth]{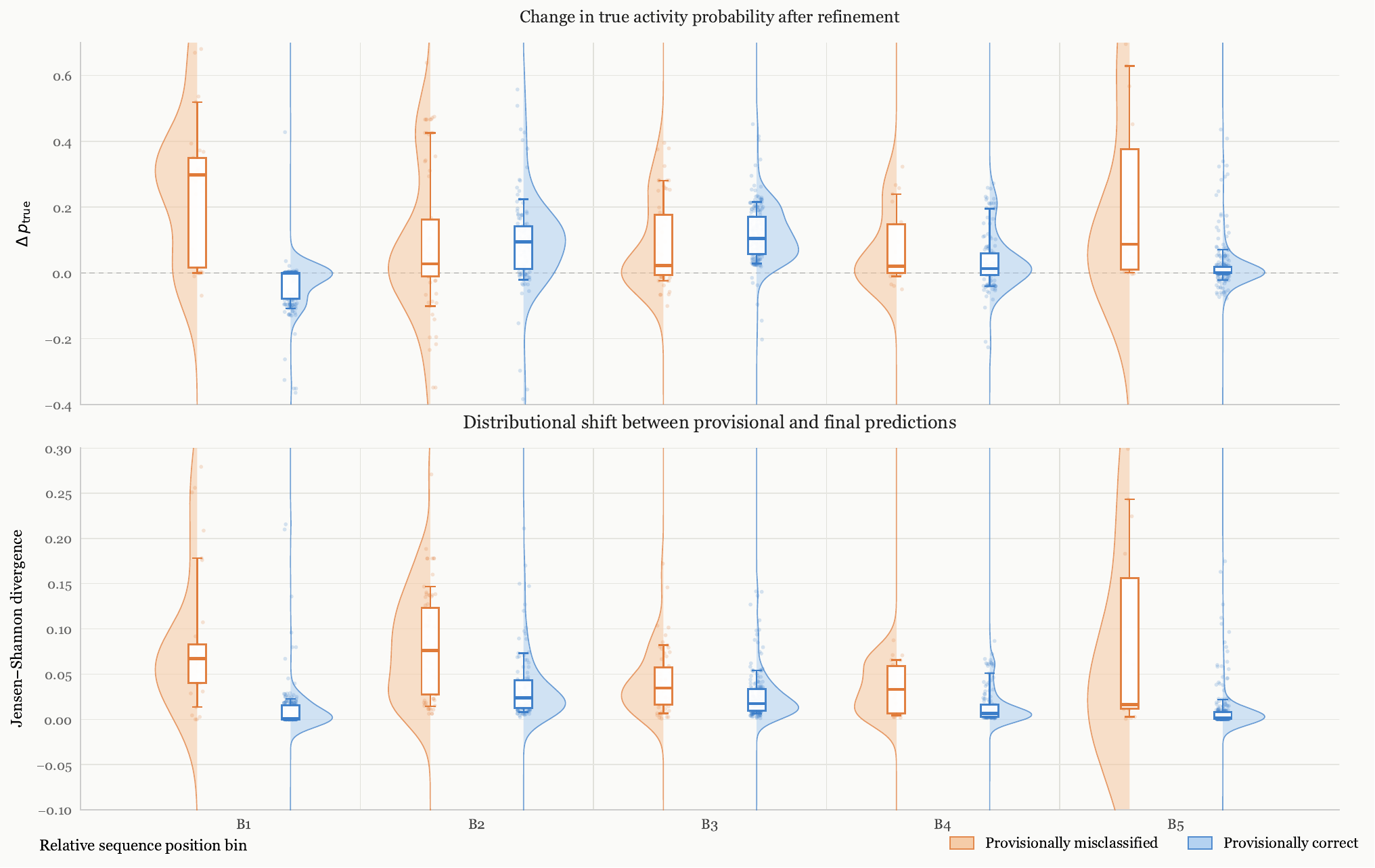}
    \\
    {\footnotesize\textbf{(c) BPI20}}
  \end{subfigure}
  \hfill
  \begin{subfigure}[b]{0.48\textwidth}
    \centering
    \includegraphics[width=\linewidth]{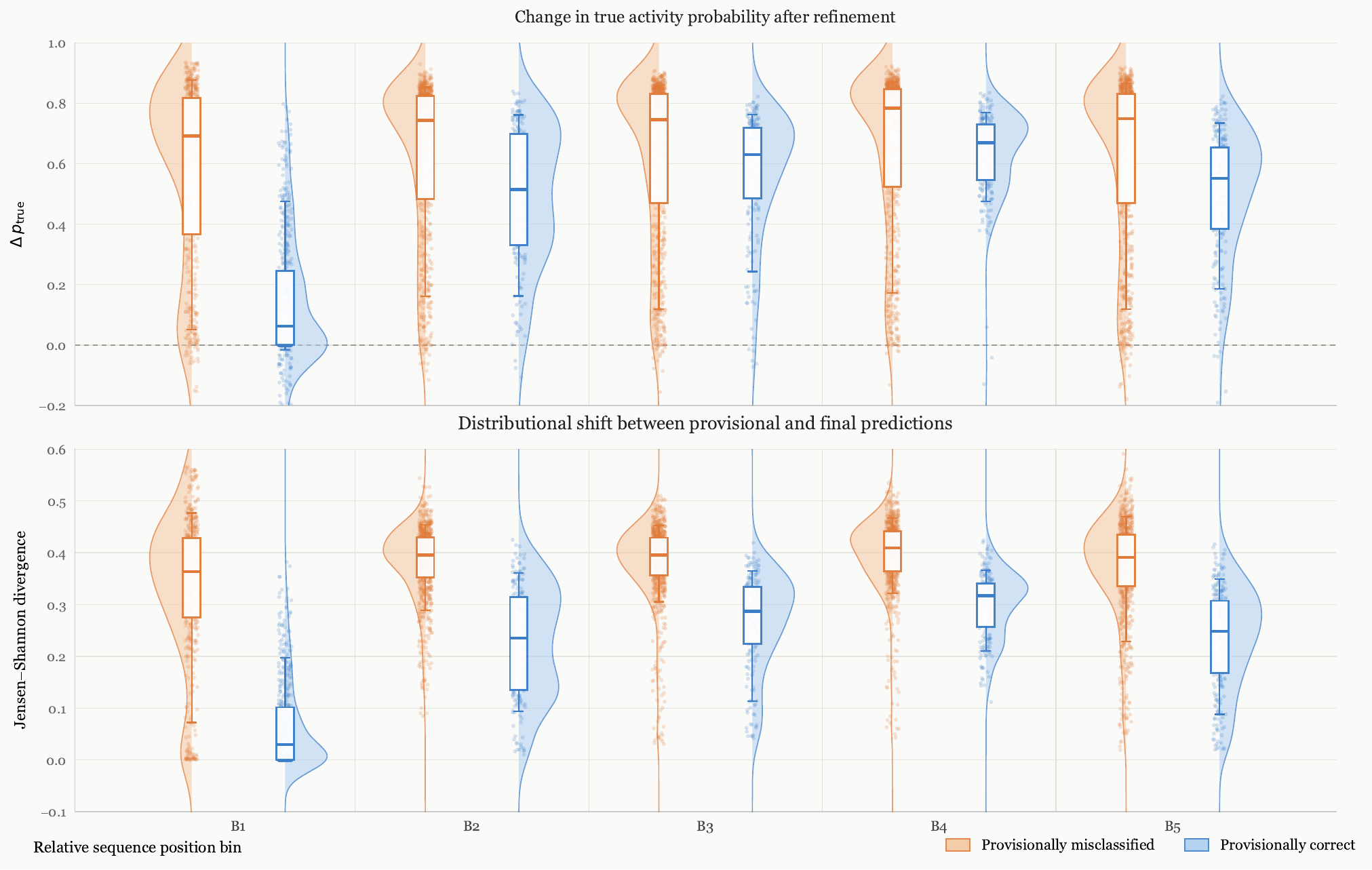}
    \\
    {\footnotesize\textbf{(d) BPI17}}
  \end{subfigure}  
  \caption{Refinement Based Activity Distribution Reshaping for BPI13I, Sepsis, BPI20 and BPI17}
  \label{appfig:refine}
\end{figure}

\begin{figure}[pos=htbp]
  \centering  
 \begin{subfigure}[b]{\textwidth}
    \centering
    \includegraphics[width=\linewidth, trim={20mm 20mm 20mm 20mm}, clip]{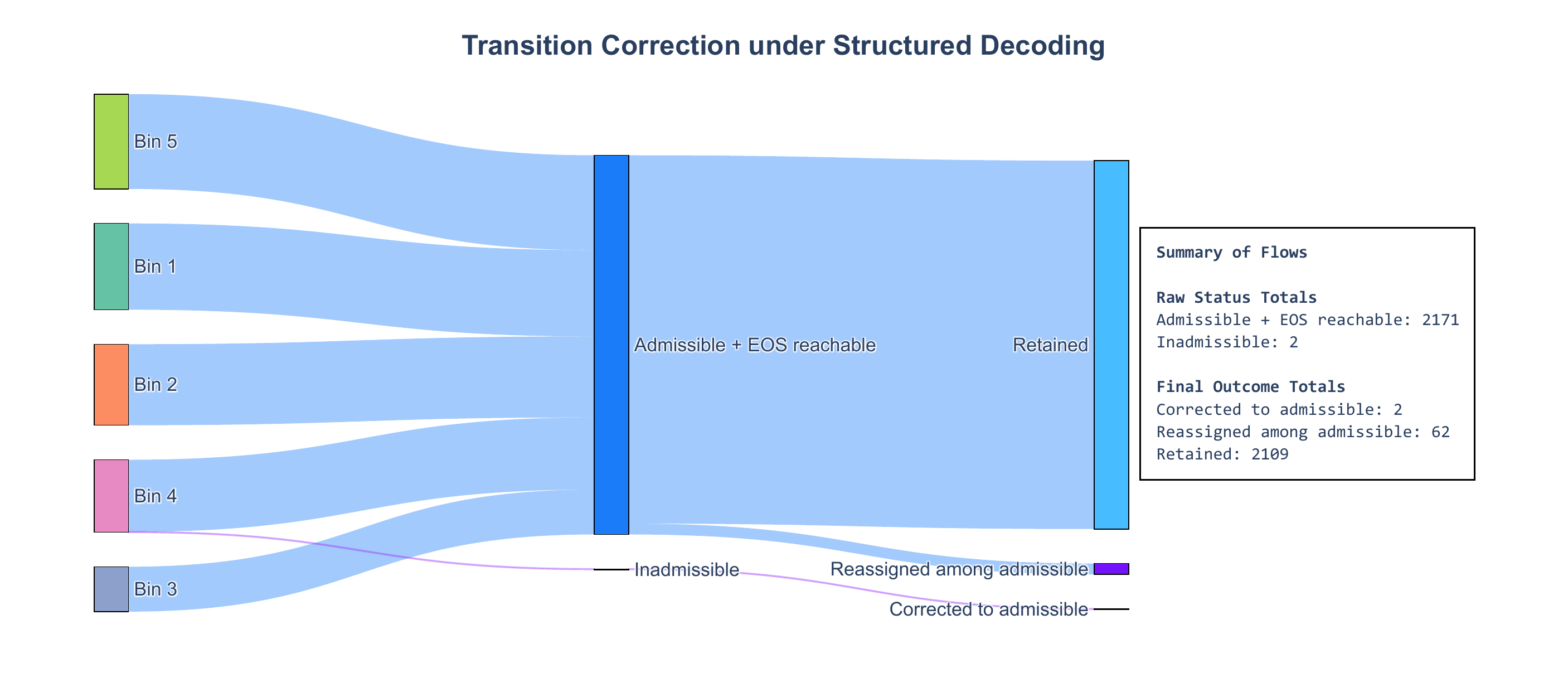}
    \\
    {\footnotesize\textbf{(a) Helpdesk}}
  \end{subfigure}
   \begin{subfigure}[b]{\textwidth}
    \centering
    \includegraphics[width=\linewidth, trim={20mm 20mm 20mm 20mm}, clip]{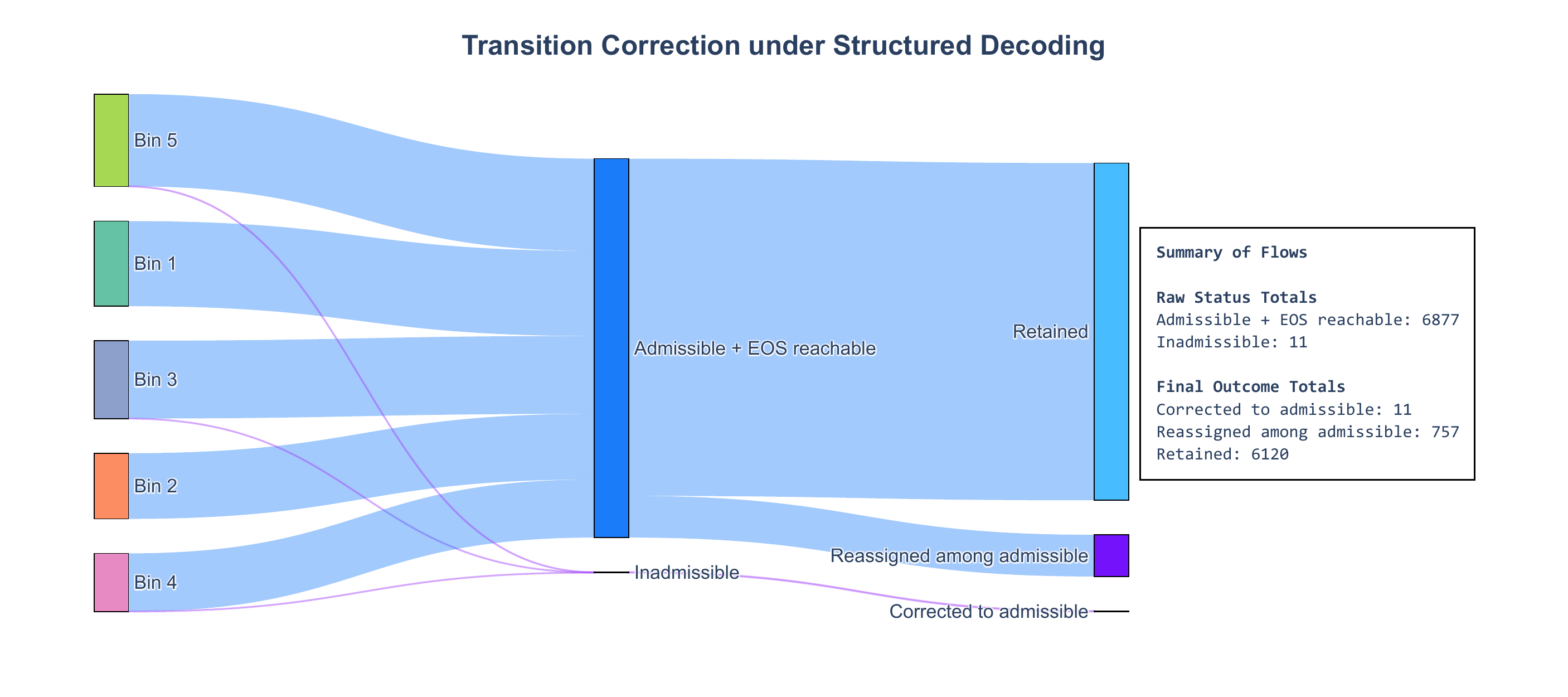}
    \\
    {\footnotesize\textbf{(b) BPI13I}}
  \end{subfigure}
   \begin{subfigure}[b]{\textwidth}
    \centering
    \includegraphics[width=\linewidth, trim={20mm 20mm 20mm 20mm}, clip]{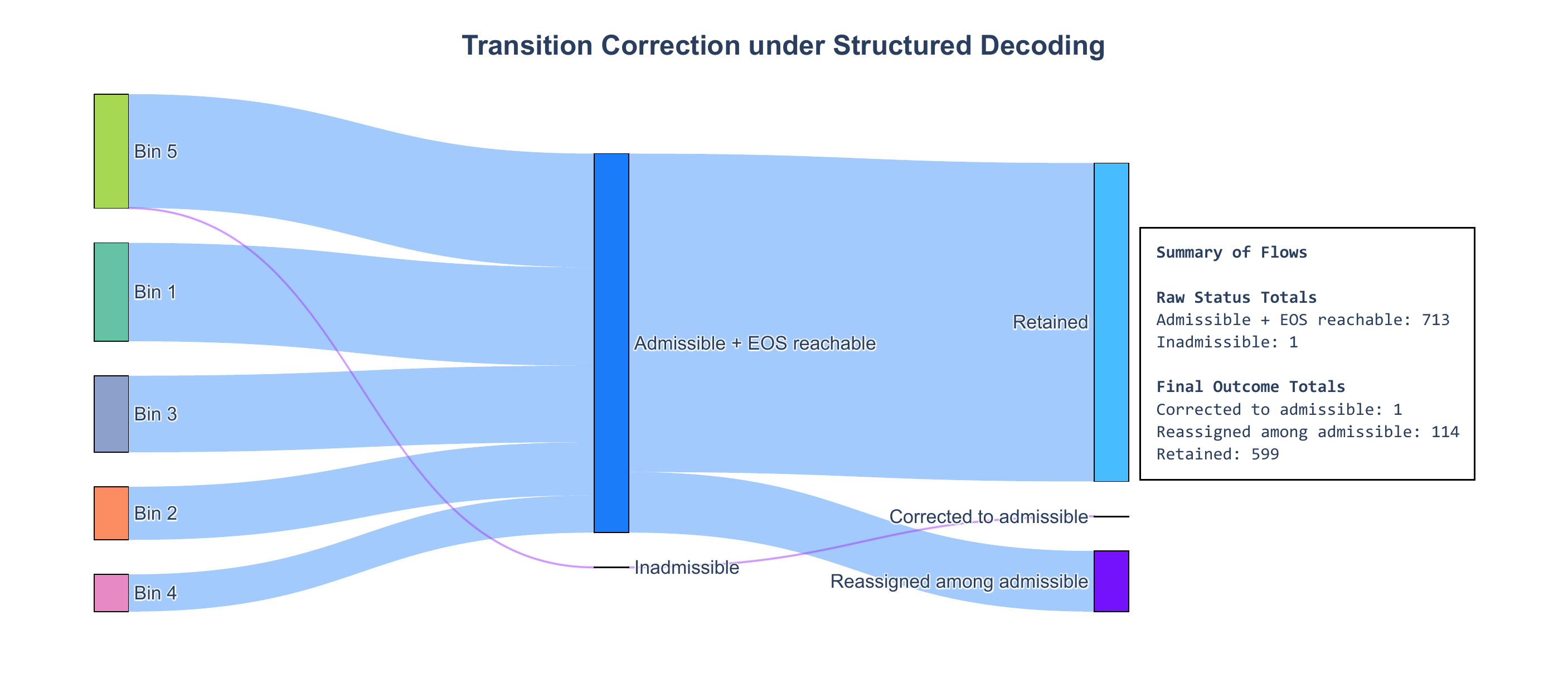}
    \\
    {\footnotesize\textbf{(c) BPI13C}}
  \end{subfigure}
  % --- Main caption for the entire figure ---
  \caption{Transition Correction under Structured Decoding for Helpdesk, BPI13I, and BPI13C}
  \label{appfig:correction_1}
\end{figure}

\begin{figure}[pos=htbp]
  \centering  
 \begin{subfigure}[b]{\textwidth}
    \centering
    \includegraphics[width=\linewidth, trim={20mm 20mm 20mm 20mm}, clip]{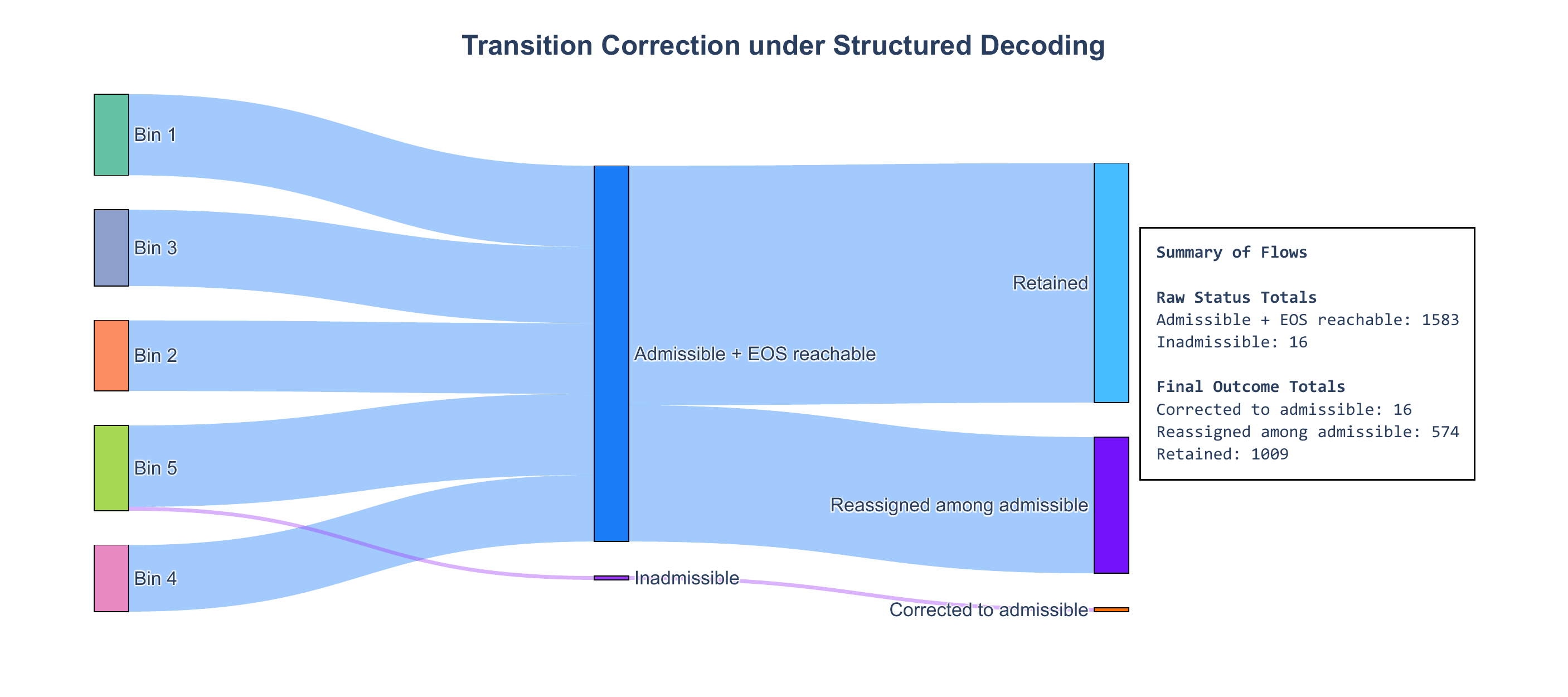}
    \\
    {\footnotesize\textbf{(d) Sepsis}}
  \end{subfigure}
   \begin{subfigure}[b]{\textwidth}
    \centering
    \includegraphics[width=\linewidth, trim={20mm 20mm 20mm 20mm}, clip]{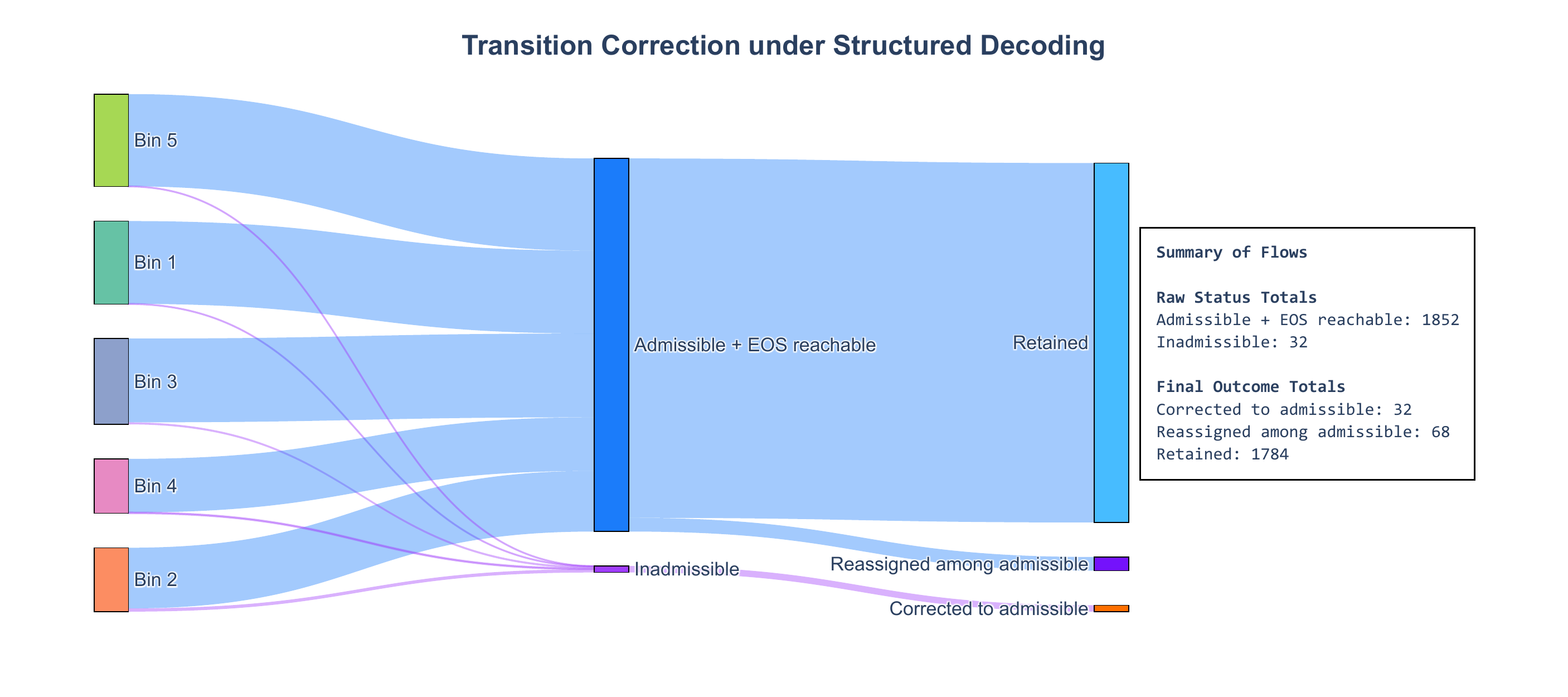}
    \\
    {\footnotesize\textbf{(e) BPI20}}
  \end{subfigure}
  % --- Main caption for the entire figure ---
  \caption{Transition Correction under Structured Decoding for Sepsis and BPI20}
  \label{appfig:correction_2}
\end{figure}
\end{document}